%% file: main.tex
\documentclass[10pt,twocolumn,letterpaper]{article}

\usepackage[pagenumbers]{wacv} 


\usepackage[most]{tcolorbox}
\definecolor{wacvblue}{rgb}{0.21,0.49,0.74}
\usepackage[pagebackref,breaklinks,colorlinks,allcolors=wacvblue]{hyperref}

\def\wacvPaperID{*****} 
\def\confName{WACV}
\def\confYear{2027}

\title{RoboPhys-3D: A Comprehensive Embodied World Model Evaluation \\ via 3D Reconstruction}

\author{Tianyi Wang\\
The University of Texas at Austin\\
{\tt\small bonny.wang@utexas.edu}
\and
Jiazhou Chen\\
The University of Texas at Austin\\
{\tt\small jiazhouchen@utexas.edu}
\and
Yiming Xu\\
The University of Texas at Austin\\
{\tt\small yiming.xu@utexas.edu}
\and
Xiangyu Li\\
The University of Texas at Austin\\
{\tt\small xiangyu$\_$li@utexas.edu}
\and
Tianyi Zeng\\
Purdue University\\
{\tt\small zeng366@purdue.edu}
\and
Chih-Hsien Chou\\
Futurewei Technologies, Inc.\\
{\tt\small chih.hsien.chou@futurewei.com}
\and
Ning Lu\\
Futurewei Technologies, Inc.\\
{\tt\small nlu@futurewei.com}
\and
Liang Peng\\
Futurewei Technologies, Inc.\\
{\tt\small lpeng@futurewei.com}
\and
Junfeng Jiao\\
The University of Texas at Austin\\
{\tt\small jjiao@austin.utexas.edu}
\and
Christian Claudel\textsuperscript{\dag}\\
The University of Texas at Austin\\
{\tt\small christian.claudel@utexas.edu}
}%

\begin{document}
\maketitle

\input{sec/0_abstract}    
\input{sec/1_introduction}
\input{sec/2_relatedwork}
\input{sec/3_dataset}
\input{sec/4_evaluation}
\input{sec/5_experiment}
\input{sec/6_conclusion}
{
    \small
    \bibliographystyle{ieeenat_fullname}
    \bibliography{main}
}

\newpage
\appendix
\onecolumn

\section{Supplementary Details about RoboPhys-3D Dataset}
\label{appendix:dataset}

\subsection{Dataset Statistics and Task Taxonomy}

The key statistics and the complete task taxonomy of the RoboPhys-3D dataset are summarized in Tables \ref{tab2} and \ref{tab3}, respectively.
RoboPhys-3D contains 50 manipulation tasks organized into four regimes according to the number of active manipulators (single- or dual-arm) and the presence of a designated target.
Episodes are partitioned per task into 70\%/15\%/15\% train/validation/test split, and each episode is treated as an indivisible split unit: all synchronized camera views, textual conditions, scene annotations, reconstruction outputs, and generated derivatives associated with an episode are assigned to the same partition.
This protocol prevents cross-view and derivative-level leakage between training and evaluation sets.

\begin{table}[ht]
\centering
\caption{Key statistics of the RoboPhys-3D dataset.}
\small
\setlength{\tabcolsep}{4pt}
\begin{tabular}{ll}
\toprule
\textbf{Component} & \textbf{RoboPhys-3D} \\
\midrule
Simulation Platform        & RoboTwin 2.0 \\
Manipulation Regimes       & 4 \\
Tasks                      & 50 \\
Episodes per Task          & 100 \\
Camera Views per Episode   & 5 \\
\midrule
Single-Arm w/o Target Task & 11 \\
Single-Arm w/ Target Task  & 21 \\
Dual-Arm w/o Target Task   & 5  \\
Dual-Arm w/ Target Task    & 13 \\
\midrule
Ground-Truth Episodes      & 5,000 \\
Ground-Truth Videos (5 Views) & 25,000 \\
\midrule
Text Conditions per Instance & 3 \\
3D Reconstruction Methods  & 4 \\
Reconstructed 3D Scenes (Main-Perspective) & 20,000 \\
\midrule
Evaluated World Models     & 4 \\
Generated Videos           & 120,000 \\
Videos Scored in Evaluation & 145,000 \\
\midrule
Split Unit                  & Episode-Level \\
Split Strategy              & Per-Task \\
Train/Validation/Test       & 70 / 15 / 15 Episodes per Task \\
Split Leakage Control       & All Views and Generated Data from One Episode Remain in the Same Split \\
\bottomrule
\end{tabular}
\label{tab2}
\end{table}

\begin{table}[ht]
\centering
\caption{Task taxonomy of the 50 RoboTwin 2.0 manipulation tasks in RoboPhys-3D.}
\small
\setlength{\tabcolsep}{4pt}
\begin{tabular}{@{}l p{0.42\textwidth} p{0.42\textwidth}@{}}
\toprule
& \textbf{Without Target} & \textbf{With Target} \\
\midrule
\textbf{Single-Arm}
& \textit{(11 Tasks)} \newline
   Adjust Bottle, Click Alarmclock, Click Bell, Press Stapler, Turn Switch, Shake Bottle, Shake Bottle Horizontally, Open Laptop, Open Microwave, Rotate QRcode, Move Playingcard Away
& \textit{(21 Tasks)} \newline
   Move Can Pot, Move Pillbottle Pad, Move Stapler Pad, Place A2B Left, Place A2B Right, Place Container Plate, Place Empty Cup, Place Fan, Place Mouse Pad, Place Object Scale, Place Object Stand, Place Phone Stand, Place Shoe, Stack Blocks Two, Stack Blocks Three, Stack Bowls Two, Stack Bowls Three, Blocks Ranking RGB, Blocks Ranking Size,
   Beat Block Hammer, Stamp Seal \\
\midrule
\textbf{Dual-Arm}
& \textit{(5 Tasks)} \newline
   Grab Roller, Lift Pot, Pick Dual Bottles, Pick Diverse Bottles,
   Handover Mic
& \textit{(13 Tasks)} \newline
   Put Object Cabinet, Handover Block, Hanging Mug, Place Bread Skillet,
   Place Dual Shoes, Place Burger Fries, Place Cans Plasticbox,
   Place Can Basket, Place Object Basket, Put Bottles Dustbin,
   Scan Object, Place Bread Basket,
   Dump Bin Bigbin \\
\bottomrule
\end{tabular}
\label{tab3}
\end{table}

\subsection{Model Configurations}

The configurations of the evaluated reconstruction methods, video world models, and IDMs are summarized in Tables \ref{tab4}--\ref{tab6}.

The reconstruction methods cover two complementary paradigms.
VGGT and VGGT-$\Omega$ perform pose-free feed-forward inference with fixed network capacity, whereas 4DGS and 4C4D optimize a scene-specific dynamic representation whose complexity is scene-dependent, using the camera calibration exported from the simulator.
In RoboPhys-3D, 4C4D produces substantially denser Gaussian representations than 4DGS.
For 4C4D, four synchronized camera streams are required for reconstruction and the resulting scene is evaluated from the designated main perspective.

The evaluated video world models span substantially different architectural scales, from the compact spatiotemporal U-Net of RoboDreamer to large Transformer-based models such as Wan 2.2 and Cosmos 3.
To ensure comparability, all generated videos are converted to the common evaluation resolution and fixed to 81 frames before scoring.
The reported inference times are measured under an identical hardware and software configuration and characterize the computational cost of our evaluation setup rather than intrinsic model efficiency.

The three IDMs further represent distinct video-to-action mechanisms.
MIDM directly regresses actions from masked visual evidence, DreamGen uses a SigLIP-2-conditioned DiT with an action flow-matching objective, and J-IDM estimates a dense embodiment Jacobian and recovers actions through a regularized inverse mapping. 

\begin{table}[ht]
\centering
\caption{Characteristics of the evaluated 3D/4D reconstruction methods.}
\small
\resizebox{\linewidth}{!}{%
\begin{tabular}{lccccc}
\toprule
\textbf{Method} & \textbf{Paradigm} & \textbf{Representation} &
\textbf{Pose Input} & \textbf{Temporal Modeling} & \textbf{Model / Scene Size}\\
\midrule
\textit{Pose-Free Feed-Forward Estimator} & & & & & \\
\midrule
VGGT & Feed-Forward & Point/Depth Maps & Not Required & None & 1.26 B Parameters \\
VGGT-$\Omega$ & Feed-Forward & Depth Maps + Cameras & Not Required & None & 1.14 B Parameters \\
\midrule
\textit{Dynamic 4D Representation} & & & & & \\
\midrule
4DGS & Per-Scene Optimization & 3D Gaussians + 4D Deformation Field & Required & Time-Conditioned Deformation & 0.18 M Gaussians/Scene \\
4C4D & Per-Scene Optimization & 4D Gaussians & Required & Explicit 4D Dynamics & 4.28 M Gaussians/Scene \\
\bottomrule
\end{tabular}}
\label{tab4}
\end{table}

\begin{table}[ht]
\centering
\caption{Characteristics of the evaluated video world models.}
\small
\resizebox{\linewidth}{!}{%
\begin{tabular}{lcccccc}
\toprule
\textbf{Method} & \textbf{Architecture} & \textbf{Conditioning} & \textbf{Evaluation Resolution} & \textbf{Frames} & \textbf{Inference (s/video)} & \textbf{Parameters} \\
\midrule
\textit{General World Model} & & & & & & \\
\midrule
Wan-2.2-I2V-A14B & MoE DiT & Video + Text & 1280$\times$720 & 81 & 299.0 & 15.17 B \\
CogVideoX1.5-5B & 3D DiT & Video + Text & 1280$\times$768 & 81 & 191.0 & 10.55 B \\
\midrule
\textit{Embodied World Model} & & & & & & \\
\midrule
Cosmos-3-Nano & MoT DiT & Video + Text & 1280$\times$720 & 81 & 33.4 & 16.40 B \\
RoboDreamer & Spatiotemporal U-Net & Video + Text & 1280$\times$768 & 81 & 32.0 & 340.50 M \\
\bottomrule
\end{tabular}}
\label{tab5}
\end{table}

\begin{table}[ht]
\centering
\caption{Detailed model parameters of inverse dynamic models.}
\small
\resizebox{\linewidth}{!}{%
\begin{tabular}{lcccccc}
\toprule
\textbf{Method} & \textbf{Vision Encoder} & \textbf{Feature Dimensions} & \textbf{Encoder Parameters} & \textbf{Action Module} & \textbf{Prediction Mechanism} & \textbf{Module Parameters} \\
\midrule
J-IDM                 & DINOv2 ViT-L & 1024 & 303.3 M & DPT & Jacobian Field + Pseudoinverse & 32.7 M \\
MIDM               & ResNet-50 & 2048 & 23.6 M & Mask Predictor + Regressor & Masked Action Regression & 0.03 M \\
DreamGen                 & SigLIP2-L & 1024 & 316.0 M & DiT & Action Flow Matching & 321.4 M \\
\bottomrule
\end{tabular}}
\label{tab6}
\end{table}

\subsection{Prompt for Egocentric Video Caption.} 

Below is the instruction prompt used to generate captions for egocentric video segments from Physion-Eval \cite{zhang2026physion}.
\begin{tcolorbox}[colback=gray!10, colframe=gray!20, left=2mm, right=2mm, top=1mm, bottom=1mm, boxrule=0pt]
\small
You are an expert in video understanding and captioning for egocentric videos.  
Given a short {egocentric (first-person) video segment} corresponding to a {verb action label} \textbf{\textit{\{the corresponding action label\}}}, generate a concise caption describing the action occurring in the segment.
The caption must follow these requirements:
\begin{enumerate}
\item The caption should describe the scene explicitly {from a first-person viewpoint}.
\item The description must remain {objective and neutral}; do not use first-person pronouns such as {I}, {we}, or {my}.
\item Ensure the caption is {consistent with the provided verb label} and focuses on the {primary action and manipulated object(s)}.
\item Describe only {visually observable events} in the video and avoid speculation about intentions or unseen actions.
\item The caption should be {a single concise sentence}.
\end{enumerate}
The output must follow exactly the format specified below:
\begin{verbatim}
{"action_label": "text",
"caption": "text"}
\end{verbatim}
Return only the JSON object and no additional text.
\end{tcolorbox}

\newpage

Below is the final view-aware prompt used to generate captions for egocentric video segments.
\begin{tcolorbox}[colback=gray!10, colframe=gray!20, left=2mm, right=2mm, top=1mm, bottom=1mm, boxrule=0pt]
\small
You are an expert in video understanding and captioning for egocentric videos of robotic manipulation. You are given a short egocentric (first-person) video segment of a dual-arm robot performing a tabletop manipulation, together with a verb action label: \textbf{\textit{\{the corresponding action label\}}}.

\textbf{PURPOSE}: The caption you produce is the text condition for an image-to-video
generation model. That model receives only the initial frame of this same video plus your caption, and must regenerate a video matching the original as closely as possible. The initial frame already supplies the scene appearance (background, lighting, object identities, colors, starting layout), so do not spend the caption on static appearance. Spend it on the motion and temporal dynamics that transform the initial frame into the full clip - that is what the model cannot get from the frame.

\textbf{ROBOT}: Every video is captured with an Aloha-AgileX dual-arm robot. The robot model is usually not visually identifiable from the scene - typically only the two grippers are visible at the left and right edges of the frame. Regardless of how the manipulators appear, Always treat and describe them as the left and right grippers of an Aloha-AgileX dual-arm robot. Do not infer any other robot type, and do not describe the manipulators as human hands or generic tools. State the robot as an ``Aloha-AgileX dual-arm robot" in the caption so the generation model renders the correct robot. Before writing the caption, reason through these steps internally (do not output the reasoning). 
Work step by step:

\textbf{STEP 1 - Parse the action label}. 
Decompose {the corresponding action label} into: verb/action; target object(s); state cues (``headup" = ends/stays upright; ``lying down"; color; container type); effector constraint (``with the correct arm" = the arm best positioned to reach the object; identify left or right from the video).

\textbf{STEP 2 - Observe scene and viewpoint}. 
Confirm a static first-person view (camera fixed on/over the workspace) and the two Aloha-AgileX grippers at the left and right frame edges. Treat the first frame as the generation seed: separate what is already established in it from what will change. Note whether the camera stays static (it almost always does).

\textbf{STEP 3 - Identify the manipulated object and its initial state}. 
Its type, and especially its starting pose/orientation (upright, lying on its side, cap direction) and location (left/center/right, near/far). Motion is described relative to this start.

\textbf{STEP 4 - Determine the acting effector}. 
Identify which gripper of the Aloha-AgileX dual-arm robot approaches and contacts the object (left vs. right); confirm it matches the ``correct arm" constraint. Note whether the other gripper stays stationary.

\textbf{STEP 5 - Trace the motion trajectory over time}. 
Track ordered phases: initial state → approach (direction the gripper enters from) → grasp (gripper closing/contact) → execute (the path: lift / translate / rotate / pour / hand over — with direction and approximate extent) → final state. Record the object's pose and location at start and end (e.g., starts lying on its right side, ends grasped and upright near center) and the manner of motion (e.g., a single smooth continuous motion).

\textbf{STEP 6 - Cross-check consistency}. 
Verify the description matches the verb label. Resolve ambiguous terms using only what is visible (``headup" → object ends/held upright). Ensure the described starting configuration matches the initial frame. Describe only visually observable events; do not speculate about intent, success, or off-screen actions.

\textbf{STEP 7 - Compose the caption}. 
Allocate it to temporal dynamics, not static appearance. Convey, in order: the robot (an Aloha-AgileX dual-arm robot) and its acting gripper (left/right), the manipulated object (named, with only the attributes needed to pick it out), the motion trajectory from the initial configuration to the final state (direction and extent), and the resulting final pose/location. Phrase the motion as a change beginning from the configuration shown in the initial frame. State that the camera remains static. If the other gripper stays still, say so.

Caption requirements:
\begin{enumerate}
\item First-person viewpoint, static camera (the camera does not move) unless the clip clearly shows otherwise.
\item Objective and neutral; do not use first-person pronouns (I, we, my, our).
\item Consistent with the verb label; focus on the primary action and manipulated object(s).
\item Only visually observable events; no speculation about intentions or unseen actions.
\item Temporally complete: convey the ordered phases (approach → grasp → execute → final state) and the direction/extent of motion - this is what the model must reproduce beyond the given first frame.
\item Identify the manipulator as the Aloha-AgileX dual-arm robot's left/right gripper (its type is not visually obvious, so naming it is required, not redundant). Name the manipulated object so the motion binds to the correct entities; do not redescribe static appearance already visible in the initial frame.
\item Concise but motion-dense: target \{MAX\_WORDS, e.g. 50-80\} words.
\end{enumerate}
The output must follow exactly the format specified below:
\begin{verbatim}
{"action_label": "text",
"caption": "text"}
\end{verbatim}
Return only the JSON object and no additional text.
\end{tcolorbox}

\section{Supplementary Details about RoboPhys-3D Evaluation}
\label{appendix:evaluation}

\begin{figure}[ht]
  \centering
  \includegraphics[width=\linewidth]{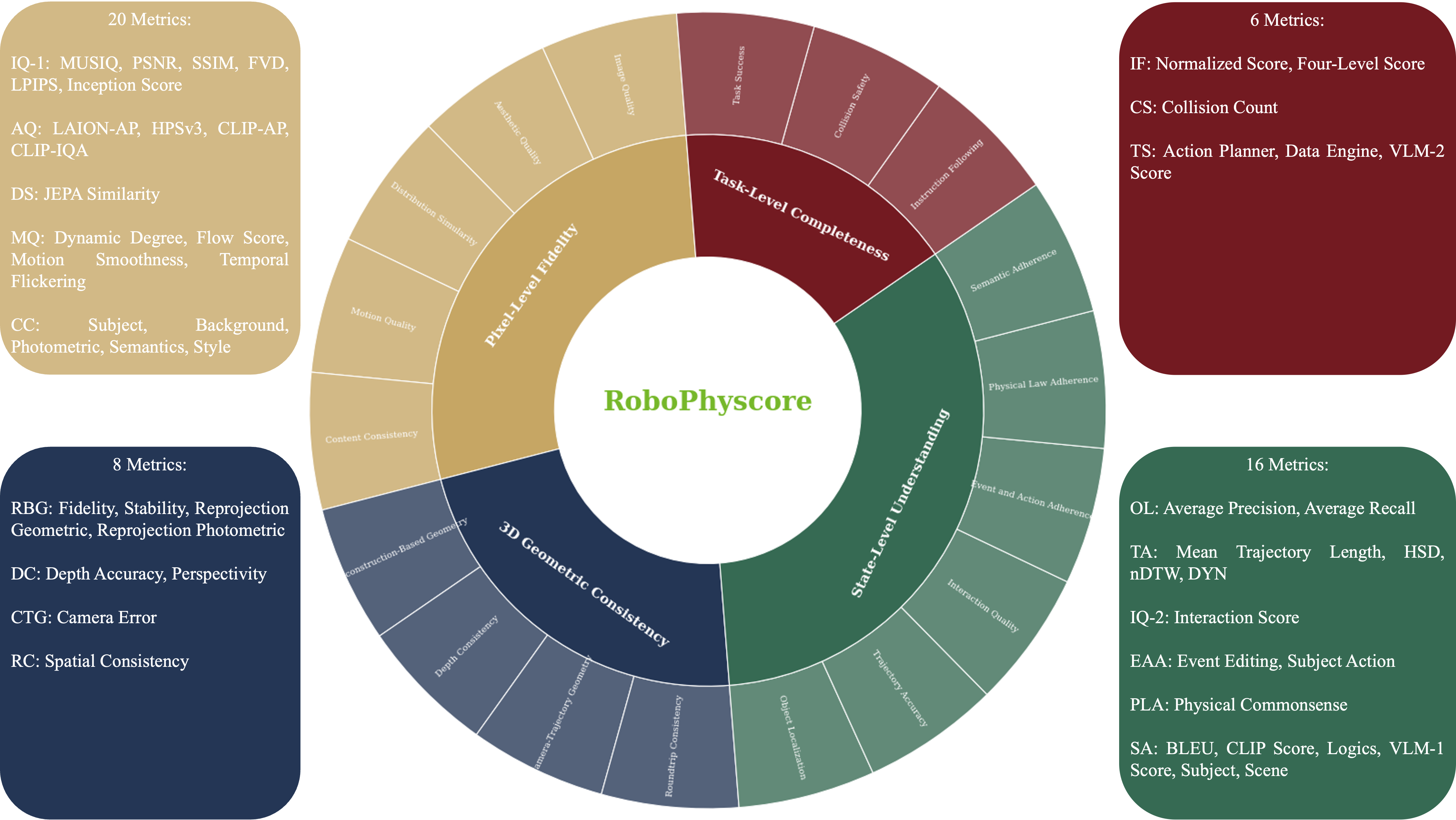}
  \caption{Overview of the RoboPhyscore.}
  \label{fig5}
\end{figure}

Figure \ref{fig5} shows the specific taxonomy of the RoboPhys-3D benchmark.

\subsection{Pixel-Level Fidelity}

Pixel-level fidelity evaluates the low-level perceptual quality of generated videos, considering five sub-dimensions: image quality, aesthetic quality, distribution similarity, motion quality, and content consistency.

\subsubsection{Image Quality}

Image quality assesses whether individual frames are perceptually reliable. 
We evaluate it using six metrics.

\textbf{MUSIQ} ([0,1]$\uparrow$) \cite{ke2021musiq} measures the clarity and sharpness of frames using the no-reference multi-scale image quality (MUSIQ) transformer, which detects distortions such as overexposure, sensor noise, and compression artifacts. 

\textbf{PSNR} ([0,+$\infty$)$\uparrow$) \cite{pratt2007digital} measures absolute pixel accuracy as the peak signal-to-noise ratio (PSNR) between each generated frame and its ground-truth counterpart. 

\textbf{SSIM} ([0,1]$\uparrow$) \cite{wang2004image} measures structural fidelity using the structural similarity (SSIM) index, which compares luminance, contrast, and local structure against the ground-truth frame. 
SSIM is complementary to PSNR, capturing structural rather than absolute error.

\textbf{FVD} ([0,+$\infty$)$\downarrow$) \cite{unterthiner2018towards} measures clip-level distributional realism as the Fréchet video distance (FVD) between the inflated 3D ConvNet (I3D) \cite{carreira2017quo} feature distributions of generated and ground-truth videos. 

\textbf{LPIPS} ([0,1]$\downarrow$) \cite{zhang2018unreasonable} measures learned perceptual image patch similarity (LPIPS) to the reference using deep features, complementing the pixel-level PSNR and SSIM.

\textbf{Inception Score} ([1,+$\infty$)$\uparrow$) \cite{salimans2016improved} measures the quality and diversity of generated frames using Inception features.
It is reference-free, computed from the internal statistics of the scored set alone, and is retained for comparability with earlier baselines.

\subsubsection{Aesthetic Quality}

Aesthetic quality evaluates the visual appeal of the generated video. 
We evaluate it using four learned human-preference predictors, none of which is reference-based.

\textbf{LAION-AP} ([0,1]$\uparrow$) \cite{schuhmann2022laion} evaluates the aesthetic appeal of frames using the LAION aesthetic predictor (LAION-AP), reflecting lighting and color composition. 

\textbf{HPSv3} ([0,1]$\uparrow$) \cite{ma2025hpsv3} measures wide-spectrum human preference using the human preference score v3 (HPSv3) reward model, percentile-normalized across the evaluated model pool, and captures overall preference rather than composition alone.

\textbf{CLIP-AP} ([0,10]$\uparrow$) \cite{schuhmann2022clip+} provides an alternative aesthetic estimate to the LAION predictor using an independent CLIP-based predictor, which focuses on layout composition, color harmony, realism, and artistic appeal.

\textbf{CLIP-IQA} ([0,1]$\uparrow$) \cite{wang2023exploring} measures perceptual quality using CLIP representations, assessing distortions such as overexposure, noise, and blur.

\subsubsection{Distribution Similarity}

Distribution similarity measures how closely the feature distribution of generated videos matches that of ground-truth videos beyond the Gaussian assumption underlying FVD, using one metric.

\textbf{JEPA Similarity} ((0,1]$\uparrow$) \cite{luo2024beyond} measures distributional realism from the maximum mean discrepancy (MMD) between V-JEPA feature distributions of generated and ground-truth videos. 

\subsubsection{Motion Quality}

Motion quality evaluates the intensity, realism, and temporal coherence of motion, ensuring that generated dynamics are meaningful and free of discontinuities. 
We evaluate it using four metrics.

\textbf{Dynamic Degree} ([0,1]$\uparrow$) \cite{huang2024vbench} quantifies motion intensity using the recurrent all-pairs field transforms (RAFT) optical-flow model \cite{teed2020raft}, focusing on the top 5\% of active pixels. 

\textbf{Flow Score} ([0,+$\infty$)$\uparrow$) \cite{liu2024evalcrafter} measures overall motion intensity by averaging RAFT optical-flow magnitudes \cite{teed2020raft} over time, distinct from Dynamic Degree, which targets localized active regions.

\textbf{Motion Smoothness} ([0,+$\infty$)$\uparrow$) \cite{duan2025worldscore} evaluates the temporal coherence of motion by predicting intermediate frames with a video frame-interpolation model \cite{zhang2024vfimamba} and comparing them to the generated frames. 
Motion magnitude is used as a weighting factor to avoid over-crediting static backgrounds and to avoid unfairly penalizing rapid motion.

\textbf{Temporal Flickering} ([0,1]$\uparrow$) \cite{huang2024vbench} measures high-frequency temporal noise that Motion Smoothness misses, computed from frame-by-frame appearance differences. 

\subsubsection{Content Consistency}

Content consistency measures the stability of objects and scenes across frames at both semantic and appearance levels. 
We evaluate it using five self-referential metrics.

\textbf{Subject Consistency} ([0,1]$\uparrow$) \cite{ying2026wbench} assesses object stability by isolating the subject with segment anything model (SAM) 2 masks \cite{ravi2025sam} and averaging DINOv2 adjacent-frame cosine similarity \cite{oquab2023dinov2} for local continuity with CLIP first-frame-anchored similarity \cite{radford2021learning} for global drift. 

\textbf{Background Consistency} ([0,1]$\uparrow$) \cite{huang2024vbench} evaluates scene stability as the mean CLIP cosine similarity \cite{radford2021learning} between consecutive frames.

\textbf{Photometric Consistency} ([0,+$\infty$)$\uparrow$) \cite{duan2025worldscore} measures pixel-level texture stability derived from the average end-point error (AEPE) of optical flow between consecutive frames. 

\textbf{Semantics Consistency} ([0,1]$\uparrow$) \cite{huang2024vbench} measures overall semantic stability using ViCLIP features \cite{wang2024internvid}. 

\textbf{Style Consistency} ([0,1]$\uparrow$) \cite{duan2025worldscore} measures the stability of rendering style using Gram-matrix statistics \cite{gatys2015neural}, ensuring that the visual style does not drift across frames.

\subsection{3D Geometric Consistency}

3D geometry consistency evaluates whether a generated video corresponds to a coherent 3D scene, and whether that scene agrees with the ground-truth geometry, along four sub-dimensions: reconstruction-based geometry, depth consistency, camera-trajectory geometry, and roundtrip consistency.

\subsubsection{Reconstruction-Based Geometry}

Reconstruction-based geometry evaluates scene-level geometric coherence by reconstructing the generated video into 3D and comparing it against ground-truth geometry. 
We evaluate it using four metrics.

\textbf{Reconstruction Fidelity} ([0,+$\infty$$\uparrow$) \cite{li2024sora} reconstructs the generated video using COLMAP SfM \cite{schonberger2016structure} followed by 3DGS \cite{kerbl20233d}, and reports the number of recovered Gaussian primitives, which falls when the input frames are too geometrically incoherent to instantiate structure.

\textbf{Sustained Stability} ([0,1]$\uparrow$) \cite{li2024sora} measures long-horizon geometric drift as the keep-ratio of correct SfM matches under increasing frame intervals. 

\textbf{Reprojection Geometric Consistency} ([0,1]$\uparrow$) \cite{ying2026wbench} estimates per-frame depth and camera pose jointly with depth anything 3 (DA3) \cite{lin2025depth}, back-projects pixels, and reprojects them across views. 
The score is then derived from the mean normalized reprojection displacement.

\textbf{Reprojection Photometric Consistency} ([0,+$\infty$)$\uparrow$) \cite{ying2026wbench} uses the same DA3 \cite{lin2025depth} reprojection pipeline to warp appearance across views and measures the PSNR between reprojected frame pairs, capturing texture flicker that geometric displacement alone misses.

\subsubsection{Depth Consistency}

Depth consistency evaluates whether the generated video preserves real-world spatial geometry and perspective plausibility. 
We evaluate it using two metrics.

\textbf{Depth Accuracy} \cite{shang2026worldarena} ([0,+$\infty$)$\downarrow$) compares monocular depth estimates of the generated and ground-truth videos, applying a median-based scaling strategy to address scale ambiguity. 

\textbf{Perspectivity} ([0,1]$\uparrow$) \cite{shang2026worldarena} assesses the 3D plausibility of the video using a Qwen3-VL judge \cite{bai2025qwen3}, focusing on scale variation with depth, lighting consistency, and occlusion relationships during camera motion. 

\subsubsection{Camera-Trajectory Geometry}

Camera trajectory geometry evaluates whether the camera motion implied by the generated video is geometrically accurate and faithful to the requested control, using one metric.

\textbf{Camera Control} ([0,1]$\uparrow$) \cite{duan2025worldscore} measures the accuracy of the recovered camera trajectory through camera-pose estimation using DROID-SLAM \cite{teed2021droid}.
For fixed-view episodes of RoboPhys-3D, this measures spurious camera drift hallucinated by the video generator.

\subsubsection{Roundtrip Consistency}

Roundtrip consistency evaluates out-of-sight persistence, a memory property that frame-to-frame and instantaneous-geometry metrics cannot capture, using one metric.

\textbf{Spatial Consistency} ([0,1]$\uparrow$) \cite{ying2026wbench} uses MegaSaM-estimated poses \cite{li2025megasam} to locate the return frame whose viewpoint best matches the initial frame, and scores DreamSim perceptual similarity \cite{fu2023dreamsim} to the first frame to measure whether the revisited scene is preserved. 

\subsection{State-Level Understanding}

State-level understanding evaluates whether the structured state of the scene matches the ground truth, spanning six sub-dimensions: object localization, trajectory accuracy, interaction quality, event and action adherence, physical law adherence, and semantic alignment. 

\subsubsection{Object Localization}

Object localization measures whether objects appear at the correct image locations over time, scored against the privileged per-frame segmentation masks exported from the simulator. 
We evaluate it using two detection metrics.

\textbf{Average Precision} ([0,1]$\uparrow$) \cite{lin2014microsoft} indicates the localization precision of the manipulated objects across frames.

\textbf{Average Recall} ([0,1]$\uparrow$) \cite{lin2014microsoft} measures localization coverage: the fraction of relevant objects correctly detected across frames.

\subsubsection{Trajectory Accuracy}

Trajectory accuracy quantifies how well the predicted object and end-effector trajectories align with the ground truth in both shape and dynamics. 
We evaluate it using four metrics.

\textbf{Mean Trajectory Length} ([0,+$\infty$)$\downarrow$) \cite{wang2025roboeval} measures the average number of decision steps across all episodes.

\textbf{Symmetric Hausdorff Distance (HSD)} ([0,1]$\uparrow$) \cite{serra1998hausdorff} measures spatial alignment as the worst-case deviation between the generated and ground-truth trajectories. 

\textbf{Normalized Dynamic Time Warping (nDTW)} ([0,1]$\uparrow$) \cite{muller2007dynamic} captures spatial-temporal alignment under monotone time warping, scoring the correct sequence and timing of motion phases.

\textbf{Dynamic Consistency (DYN)} ([0,1]$\uparrow$) \cite{yue2025ewmbench} evaluates motion dynamics such as velocity and acceleration using the Wasserstein distance \cite{villani2009wasserstein} with motion normalization.

\subsubsection{Interaction Quality}

Interaction quality evaluates whether interactions between the robot and objects are physically plausible rather than merely visually plausible, using one metric.

\textbf{Interaction Score} ([0,1]$\uparrow$) \cite{shang2026worldarena} uses Qwen3-VL \cite{bai2025qwen3} to assess contact behavior, force transmission, the absence of interpenetration, and grasp correctness. 

\subsubsection{Event and Action Adherence}

Event and action adherence evaluates whether instructed events and robot actions are actually realized in the generated video. 
We evaluate it using two structured VLM protocols adapted from interactive world model evaluation.

\textbf{Event Editing Adherence} ([0,1]$\uparrow$) \cite{ying2026wbench} applies five binary checks derived from the action specification: change detection, event occurrence, completion, detail accuracy, and anomaly absence. 

\textbf{Subject Action Adherence} ([0,1]$\uparrow$) \cite{ying2026wbench} applies the same five-check protocol tuned to whether the subject performs the instructed action, with the final check assessing unnatural motion such as physically implausible poses or interactions. 

\subsubsection{Physical Law Adherence}

Physical law adherence evaluates whether observable physical laws are respected, measured with a single score.

\textbf{Physical Commonsense} ([1,5]$\uparrow$) \cite{bansal2025videophy} judges whether the generated video follows intuitive real-world physics, covering Newton's first law, conservation of mass, solid and fluid mechanics, impenetrability, and gravitation.

\subsubsection{Semantic Adherence}

Semantic adherence evaluates whether the generated video faithfully reflects the textual condition and the specified world setting. 
We evaluate it using six metrics.

\textbf{BLEU} ([0,1]$\uparrow$) \cite{yue2025ewmbench} generates a global caption of the video and compares it with the raw task instruction using the bilingual evaluation understudy (BLEU) score \cite{papineni2002bleu} to evaluate overall alignment between the task goal and the generated video's content.

\textbf{CLIP Score} ([0,1]$\uparrow$) \cite{yue2025ewmbench} produces a detailed, step-by-step description of the task's key steps and compares it with the ground-truth step descriptions using the CLIP Score \cite{schuhmann2022clip+}.

\textbf{Logics} ([0,1]$\uparrow$) \cite{yue2025ewmbench} evaluates generated videos for commonsense violations, explicitly penalizing errors such as hallucinated object manipulations or illogical spatial relationships, so that coherent task execution is prioritized over superficial plausibility. 

\textbf{VLM-1 Score} ([0,1]$\uparrow$) \cite{shang2026worldarena} uses Qwen3-VL \cite{bai2025qwen3} and CLIP similarity \cite{radford2021learning} to assess whether the generated video truly understands and executes the given textual instruction.

\textbf{Subject Adherence} ([0,1]$\uparrow$) \cite{ying2026wbench} scores whether the subject's visual attributes match the described appearance, and whether its movement style matches declared motion priors.

\textbf{Scene Adherence} ([0,0.5]$\uparrow$) \cite{ying2026wbench} scores whether initially visible elements remain consistent throughout, and whether described but offscreen elements eventually appear.

\subsection{Task-Level Completeness}

Task-level completeness evaluates whether a generated video, once decoded into executable actions by an IDM, actually completes the manipulation task. 
We evaluate it via three sub-dimensions: instruction following, collision safety, and task success.

\subsubsection{Instruction Following}

Instruction following measures whether the depicted behavior achieves the instructed final task state. 
We evaluate it using two VLM-based scores.

\textbf{Normalized Score} ([0,1]$\uparrow$) \cite{shang2026worldarena} assesses accuracy in following instructions with respect to action type, target object, and task state, measured by Qwen3-VL \cite{bai2025qwen3} and normalized to [0,1].
 
\textbf{Four-Level Score} ([0,3]$\uparrow$) \cite{li2026worldmodelbench} defines four levels of instruction-following performance and assigns a score according to the level 0-3.

\subsubsection{Collision Safety}

Collision safety measures whether the executed behavior collides with unintended objects or with the robot itself, using one metric.

\textbf{Collision Count} ([0,+$\infty$)$\downarrow$) \cite{wang2025roboeval} reports the number of unintended robot-environment and robot–self collisions during execution in simulation.

\subsubsection{Task Success}

Task success measures whether the executed behavior accomplishes the intended task.
We evaluate it using three metrics.

\textbf{Action Planner} ([0,1]$\uparrow$) replays the decoded actions using DreamGen's IDM \cite{jang2025dreamgen} in the RoboTwin 2.0 simulator \cite{chen2025robotwin} and reports task-level success against the task's ground-truth success criterion.

\textbf{Data Engine} ([0,1]$\uparrow$) trains a baseline $\pi_{0.5}$ \cite{intelligence2025pi} policy with varying amounts of synthetic data (actions recovered by DreamGen's IDM \cite{jang2025dreamgen} and generated videos).
The performance gain of the policy in the RoboTwin 2.0 simulator \cite{chen2025robotwin} reflects the world model’s capability to enhance policy learning.

\textbf{VLM-2 Score} ([0,1]$\uparrow$) \cite{shang2026worldarena} determines from the generated video alone whether the embodied task was executed successfully.

\subsection{Score Normalization}

The 50 metrics of RoboPhys-3D originate from heterogeneous evaluation  protocols and differ in scale and direction.
Each raw metric value $x_m$ is mapped to a normalized score $s_m \in [0, 1]$ with higher values indicating better performance.
Of the 50 metrics, 33 metrics are natively defined on $[0, 1]$ with higher-is-better orientation and require no transformation; 11 metrics are normalized by the clipped affine map using the bounds in Table \ref{tab7}; two metrics use the absolute anchors; and four metrics admit dedicated transforms.
For metrics with explicit lower and upper bounds $a_m$ and $b_m$, respectively, we use:
\begin{equation}
s_m =
\begin{cases}
\mathrm{clip}\!\left(
\dfrac{x_m-a_m}{b_m-a_m},0,1
\right),
& \text{higher is better}, \\
\mathrm{clip}\!\left(
\dfrac{b_m-x_m}{b_m-a_m},0,1
\right),
& \text{lower is better},
\end{cases}
\label{eq:norm}
\end{equation}
We distinguish three normalization strategies according to the provenance of the bounds.

\subsubsection{Upstream Benchmark Bounds}

Flow score (MQ), motion smoothness (MQ), and photometric consistency (CC) adopt the empirical bounds reported by WorldArena \cite{shang2026worldarena}.
Depth accuracy (DC) adopts the same source bounds but reverses the direction because its raw quantity is an error. 
Style consistency (CC) follows the Gram-matrix distance threshold of WorldScore \cite{duan2025worldscore} and is likewise treated as lower-is-better. 

\subsubsection{RoboPhys-3D Empirical Bounds}

For metrics whose original benchmarks do not provide usable normalization bounds, we follow the percentile strategy of WorldArena \cite{shang2026worldarena}: the 1st and 99th percentiles of the raw values over the RoboPhys-3D evaluation pool serve as $a_m$ and $b_m$, respectively. 
This strategy is applied to inception score (IQ-1), CLIP-AP (AQ), reconstruction fidelity (RBG), reprojection photometric consistency (RBG), mean trajectory length (TA), and collision count (CS).

\subsubsection{Custom Normalization}

Several metrics admit a natural scale or require a dedicated transform.
For PSNR (IQ-1) and FVD (IQ-1), we adopt the absolute anchors of WoW-World-Eval \cite{fan2026wow}, $U_{\mathrm{PSNR}}=50$ and $U_{\mathrm{FVD}}=2000$, with clipped linear mappings in place of its nonlinear human-preference calibration:
\begin{equation}
s_{\mathrm{PSNR}}
=
\min\!\left(\frac{x_{\mathrm{PSNR}}}{50},1\right), 
s_{\mathrm{FVD}}
=
1-\min\!\left(\frac{x_{\mathrm{FVD}}}{2000},1\right).
\end{equation}
LPIPS (IQ-1) is already bounded on $[0,1]$ but is lower-is-better, and is direction-aligned as:
\begin{equation}
s_{\mathrm{LPIPS}} = 1-x_{\mathrm{LPIPS}}.
\end{equation}
Physical commonsense (PLA) is defined on a 1--5 Likert scale and is normalized as:
\begin{equation}
s_{\mathrm{PLA}}
=
\frac{x_{\mathrm{PLA}}-1}{4}.
\end{equation}
Four-level score (IF) is defined on a 0--3 scale and is mapped by:
\begin{equation}
s_{\mathrm{IF}}
=
\frac{x_{\mathrm{IF}}}{3}.
\end{equation}
Finally, scene adherence (SA) has an effective maximum of $0.5$ in our setting:
\begin{equation}
s_{\mathrm{Scene}}
=
\mathrm{clip}\!\left(
\frac{x_{\mathrm{Scene}}}{0.5},0,1
\right).
\end{equation}
After these transformations, every metric is expressed on a common $[0,1]$ higher-is-better scale.
The corresponding numerical bounds are reported in Table \ref{tab7}.

\begin{table}[ht]
\centering
\caption{Bounds used for score normalization.}
\small
\begin{tabular}{lcccl}
\toprule
\textbf{Metric} &
\textbf{Direction} &
\textbf{Upper} &
\textbf{Lower} &
\textbf{Bound Source} \\
\midrule
Flow Score & $\uparrow$ & 8.9414 & 0.0531 & WorldArena \\
Motion Smoothness & $\uparrow$ & 2.6413 & 0.0000 & WorldArena \\
Photometric Consistency & $\uparrow$ & 6.7899 & 0.1257 & WorldArena \\
Depth Accuracy & $\downarrow$ & 4.3711 & 0.2228 & WorldArena \\
Style Consistency & $\downarrow$ & 0.0070 & 0.0000 & WorldScore \\
\midrule
PSNR & $\uparrow$ & 50.0000 & 0.0000 & WoW-World-Eval\\ 
FVD & $\downarrow$ & 2000.0000 & 0.0000 & WoW-World-Eval \\
\midrule
Inception Score & $\uparrow$ & 3.5279 & 2.1041 & RoboPhys-3D (Ours) \\
CLIP-AP & $\uparrow$ & 4.9646 & 3.8572 & RoboPhys-3D (Ours) \\
Reconstruction Fidelity & $\uparrow$ & 168.5000 & 22.0000 & RoboPhys-3D (Ours) \\
Reprojection Photometric Consistency & $\uparrow$ & 21.7803 & 7.7000 & RoboPhys-3D (Ours) \\
Mean Trajectory Length & $\downarrow$ & 3.8800 & 0.1349 & RoboPhys-3D (Ours) \\
Collision Count & $\downarrow$ & 68.0000 & 0.0000 & RoboPhys-3D (Ours) \\
\bottomrule
\end{tabular}
\label{tab7}
\end{table}

\subsection{Metrics Classification}

Tables \ref{tab:withreference} and \ref{tab:withoutreference} classify the 50 metrics by the signal that grounds their judgment. 
A metric is reference-based when it consults information external to the evaluated video, and the reference source yields three groups: video-grounded metrics (13) compare against the reconstruction-matched ground-truth video; VLM-grounded metrics (12) rely on a Qwen3-VL judgment conditioned on the instruction or rubric; and simulator-grounded metrics (5) require privileged state, calibration, or action replay in RoboTwin 2.0 simulator. 
The remaining 20 metrics are reference-free and computable from the generated video alone. 
This classification determines where each metric can be deployed: reference-free and VLM-grounded metrics transfer to videos without ground truth, video-grounded metrics require the paired reference, and simulator-grounded metrics are available only in simulation. 
Finally, the eight metrics selected into \textit{RoboPhyscore} comprise four video-grounded, three VLM-grounded, and one reference-free metric.

\begin{table}[ht]
  \centering
  \caption{Metrics evaluated with an external reference.}
  \label{tab:withreference}
  \small
  \begin{tabular}{lc}
  \toprule
  \textbf{Metric} & \textbf{Grounding} \\
  \midrule
  PSNR (IQ-1) & Video-Grounded \\
  SSIM (IQ-1) & Video-Grounded \\
  FVD (IQ-1) & Video-Grounded \\
  LPIPS (IQ-1) & Video-Grounded \\
  JEPA Similarity (DS) & Video-Grounded \\
  Depth Accuracy (DC) & Video-Grounded \\
  Average Precision (OL) & Video-Grounded \\
  Average Recall (OL) & Video-Grounded \\
  HSD (TA) & Video-Grounded \\
  nDTW (TA) & Video-Grounded \\
  DYN (TA) & Video-Grounded \\
  BLEU (SA) & Video-Grounded \\
  CLIP Score (SA) & Video-Grounded \\
  \midrule
  Perspectivity (DC) & VLM-Grounded \\
  Interaction Score (IQ-2) & VLM-Grounded \\
  Event Editing Adherence (EAA) & VLM-Grounded \\
  Subject Action Adherence (EAA) & VLM-Grounded \\
  Physical Commonsense (PLA) & VLM-Grounded \\
  Logics (SA) & VLM-Grounded \\
  VLM-1 Score (SA) & VLM-Grounded \\
  Subject Adherence (SA) & VLM-Grounded \\
  Scene Adherence (SA) & VLM-Grounded \\
  Normalized Score (IF) & VLM-Grounded \\
  Four-Level Score (IF) & VLM-Grounded \\
  VLM-2 Score (TS) & VLM-Grounded \\
  \midrule
  Camera Control (CTG) & Simulator-Grounded \\
  Mean Trajectory Length (TA) & Simulator-Grounded \\
  Collision Count (CS) & Simulator-Grounded \\
  Action Planner (TS) & Simulator-Grounded \\
  Data Engine (TS) & Simulator-Grounded \\
  \bottomrule
  \end{tabular}
\end{table}

\begin{table}[ht]
  \centering
  \caption{Metrics evaluated without any external reference.}
  \label{tab:withoutreference}
  \small
  \begin{tabular}{l}
  \toprule
  \textbf{Metric} \\
  \midrule
  MUSIQ (IQ-1) \\
  Inception Score (IQ-1) \\
  LAION AP (AQ) \\
  HPSv3 (AQ) \\
  CLIP-AP (AQ) \\
  CLIP-IQA (AQ) \\
  Dynamic Degree (MQ) \\
  Flow Score (MQ) \\
  Motion Smoothness (MQ) \\
  Temporal Flickering (MQ) \\
  Subject Consistency (CC) \\
  Background Consistency (CC) \\
  Photometric Consistency (CC) \\
  Semantics Consistency (CC) \\
  Style Consistency (CC) \\
  Reconstruction Fidelity (RBG) \\
  Sustained Stability (RBG) \\
  Reprojection Geometric Consistency (RBG) \\
  Reprojection Photometric Consistency (RBG) \\
  Spatial Consistency (RC) \\
  \bottomrule
  \end{tabular}
\end{table}

\subsection{Analysis of Average Full Score}

Figures \ref{fig8} and \ref{fig9} provide a fine-grained view of the 18 sub-dimensions underlying AFS for different video world models and reconstruction methods, respectively. 
For visualization, each sub-dimension score is divided by its theoretical maximum, such that all radar axes lie in $[0,1]$ and are directly comparable despite their different native ranges.

\subsubsection{Video World Models}

\begin{figure}[ht]
  \centering
  \includegraphics[width=\linewidth]{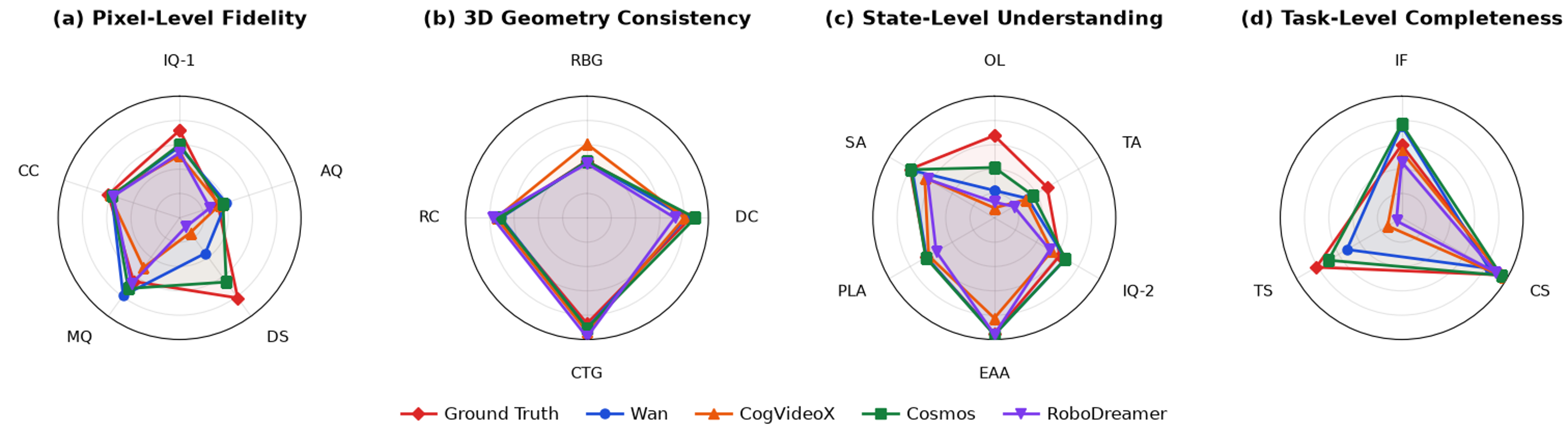}
  \caption{Detailed results of the RoboPhys-3D benchmark for four representative video world models.}
  \label{fig8}
\end{figure}

Figure \ref{fig8} shows that Cosmos achieves the strongest and most balanced performance among the evaluated world models. 
Its AFS reaches $2.7797$ ($97.9\%$ of Ground Truth), followed by Wan ($2.5714$), CogVideoX ($2.2517$), and RoboDreamer ($2.1246$).
Across the four levels, Cosmos retains $93.7\%$ of Ground Truth in pixel-level fidelity and $92.0\%$ in state-level understanding, while slightly exceeding it in 3D geometry consistency ($102.2\%$) and task-level completeness ($102.7\%$).
The detailed results also reveal that strong performance at one level does not imply accurate embodied prediction at another level. 
CogVideoX achieves the highest 3D geometry consistency among all models, but its state- and task-level scores drop to only $72.4\%$ and $69.9\%$ of Ground Truth, respectively.
RoboDreamer exhibits a similar discrepancy: its geometry-level score is approximately equal to Ground Truth, while its task-level completeness retains only $59.7\%$. 
In contrast, Wan maintains relatively strong perceptual and semantic behavior but exhibits pronounced degradation in object localization ($33.2\%$) and trajectory accuracy ($63.3\%$). 
These patterns confirm that internally coherent geometry, perceptual quality, and semantic plausibility capture complementary rather than interchangeable aspects of EWM capability.

\subsubsection{Reconstruction Methods}

\begin{figure}[ht]
  \centering
  \includegraphics[width=\linewidth]{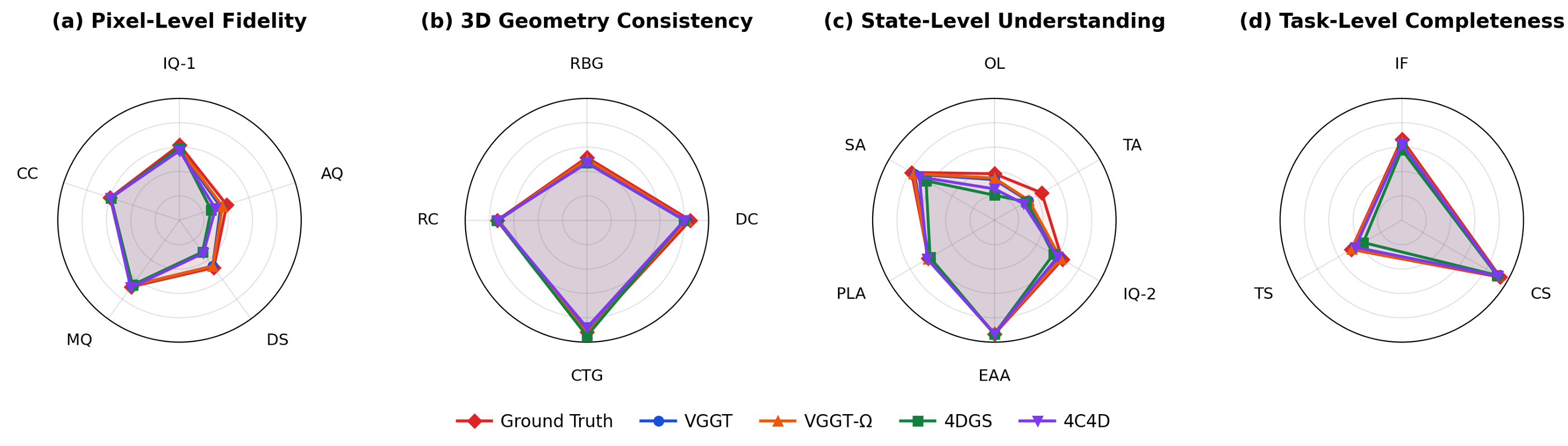}
  \caption{Detailed results of the RoboPhys-3D benchmark for four representative reconstruction methods.}
  \label{fig9}
\end{figure}

Figure \ref{fig9} shows that the two feed-forward reconstruction methods achieve the best overall benchmark performance. 
VGGT-$\Omega$ and VGGT achieve AFS values of $2.5598$ and $2.5292$, retaining $97.0\%$ and $95.8\%$ of the reference ($2.6400$), respectively.
4C4D retains $92.8\%$ ($2.4500$), whereas 4DGS exhibits the largest overall degradation at $2.3865$ ($90.4\%$).
Reconstruction sensitivity is strongly dimension-dependent: 3D geometry consistency remains comparatively stable across substrates, whereas state-level understanding and task-level completeness exhibit larger degradation. 
Thus, a reconstruction can remain internally consistent in camera and roundtrip geometry while losing information required to recover the correct physical state.
The results demonstrate that errors introduced by generation and reconstruction manifest differently across perceptual, geometric, state-grounded, and execution-oriented dimensions.

\subsection{Analysis of RoboPhyscore}

Figure \ref{fig6} further decomposes RoboPhyscore into its eight task-aligned source metrics, which are already normalized to $[0,1]$.

\subsubsection{Video World Models}

Figure \ref{fig6}(a) shows that Cosmos provides the most balanced task-aligned profile among the generated models, achieving a RoboPhyscore of $0.6330$. 
However, larger deficits remain in the more reference-grounded dimensions, such as average recall $0.4773$ ($67.3\%$) and nDTW $0.4297$ ($72.4\%$). 
Strong semantic and physical plausibility does not guarantee accurate state recovery or motion evolution.
Wan exhibits a sharper imbalance: its semantic and distributional components remain comparatively strong, while its average recall and nDTW retain only $40.9\%$, and $52.5\%$ of Ground Truth. 
CogVideoX and RoboDreamer obtain substantially lower RoboPhyscores. 
Notably, RoboDreamer maintains a relatively high normalized FVD score despite its low overall score, illustrating that no single perceptual or distributional metric is sufficient to characterize embodied prediction quality.

\subsubsection{Reconstruction Methods}

\begin{figure}[ht]
  \centering
  \includegraphics[width=\linewidth]{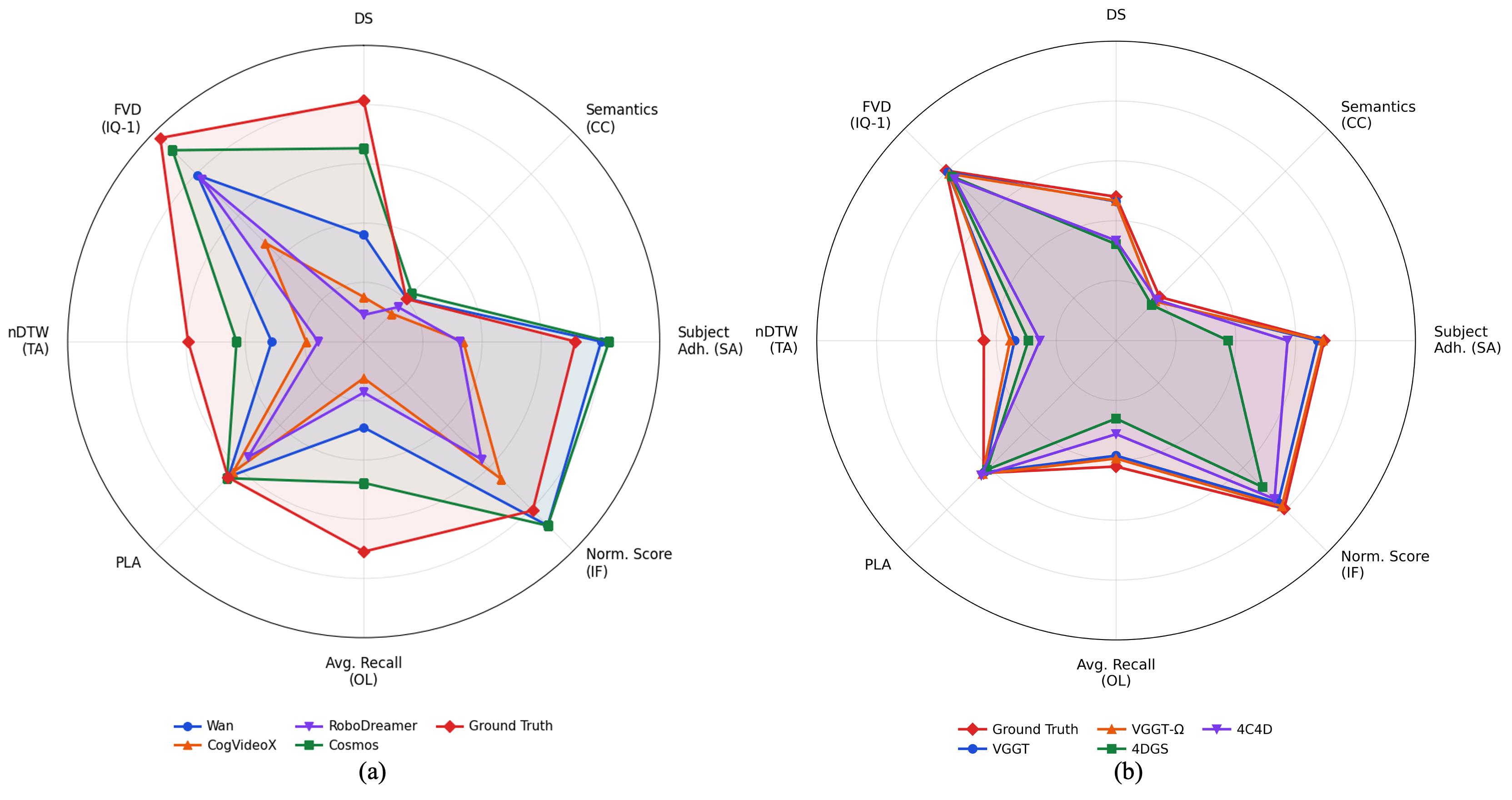}
  \caption{Detailed results of the proposed RoboPhyscore. (a) RoboPhyscore for representative video world models. (b) RoboPhyscore for representative reconstruction methods.}
  \label{fig6}
\end{figure}

Figure \ref{fig6}(b) reveals a clear separation between the feed-forward and optimization-based reconstruction substrates. 
VGGT-$\Omega$ and VGGT retain $96.1\%$ and $94.7\%$ of the raw-substrate RoboPhyscore, whereas 4C4D and 4DGS retain $85.4\%$ and $78.2\%$.
The largest reconstruction-induced losses occur in metrics that depend on accurate spatial state and temporal correspondence. 
Relative to the raw substrate, 4DGS retains only $53.9\%$ of subject adherence, $61.9\%$ of average recall, and $66.3\%$ of nDTW. 
In contrast, physical commonsense and FVD remain at $97.5\%$ and $96.6\%$ of their reference values. 
4C4D better preserves subject adherence and average recall than 4DGS, but its nDTW decreases to $0.2549$, only $57.7\%$ of the Ground Truth.
These results explain the greater reconstruction sensitivity of RoboPhyscore compared with AFS. 
Its selected metrics place substantial weight on whether object identity, localization, trajectory evolution, and instruction-conditioned behavior survive the reconstruction pipeline, while still retaining complementary semantic and physical-plausibility signals. 

\subsection{Full Metric-Level Correlation Analysis}

\begin{figure*}[ht]
  \centering
  \includegraphics[width=\linewidth]{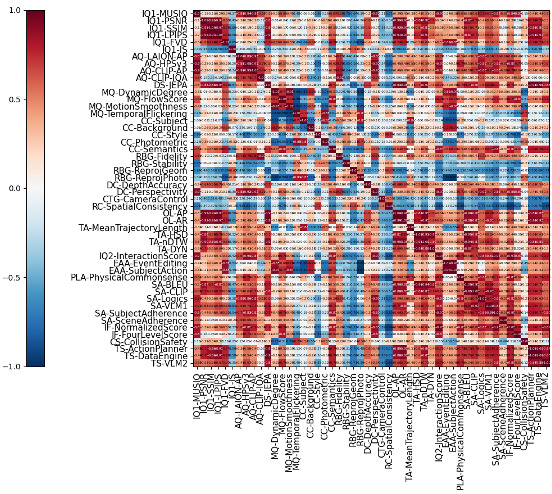}
  \caption{Pearson correlation among 50 metrics.}
  \label{fig7}
\end{figure*}

Figure \ref{fig7} presents the Pearson correlation matrix over all
50 metrics in RoboPhys-3D, computed across the model–reconstruction configurations, and complements the sub-dimension-level analysis.
The matrix reveals a clear block structure: metrics measuring closely related perceptual, geometric, state, or task-level properties are often strongly correlated, whereas metrics belonging to different evaluation dimensions exhibit substantially more heterogeneous relationships.

Strong positive correlations first appear within several metric families.
Among image-quality metrics, PSNR is highly correlated with SSIM ($r=0.913$) and LPIPS ($r=0.940$), while SSIM and LPIPS exhibit an even stronger relationship ($r=0.965$). 
Average precision and average recall (OL) are similarly almost redundant at the configuration level ($r=0.992$). 
Strong within-family agreement is also observed among trajectory metrics: HSD and trajectory dynamics correlate at $r=0.967$, while HSD and nDTW reach $r=0.912$. 
These patterns indicate that metrics derived from closely related visual or state representations often respond similarly to changes in generation and reconstruction quality.

Task-oriented metrics form another prominent positive cluster.
Action planner and data engine are nearly perfectly correlated ($r=0.999$), indicating that rollouts that support successful IDM-decoded execution also tend to provide more useful synthetic data for downstream policy learning. 
Their correlations with VLM-2 are lower ($r=0.869$ and $r=0.876$), showing that visual task-completion judgment captures related but not identical information. 
Interaction score is strongly correlated with subject adherence ($r=0.968$) and normalized score ($r=0.968$), while BLEU and CLIP-based semantic scores correlate at $r=0.947$.

In contrast, reconstruction- and reprojection-oriented metrics correlate negatively with state- and task-level measurements.
Roundtrip spatial consistency is strongly negatively correlated with interaction score ($r=-0.862$), physical commonsense ($r=-0.890$), and normalized score ($r=-0.936$). 
Reprojection photometric consistency exhibits a similarly strong inverse relationship with event editing ($r=-0.952$) and subject action adherence ($r=-0.974$). 
These results show that strong internal geometric or reprojection regularity does not necessarily imply accurate interaction semantics, physically meaningful state evolution, or successful downstream execution.

Overall, the correlation matrix reveals both redundancy and complementarity within RoboPhys-3D. 
High correlations within metric families confirm that related measurements capture common aspects of quality, while substantially weaker or even negative cross-family correlations demonstrate that perceptual fidelity, geometric consistency, state accuracy, semantic adherence, and task success are not interchangeable. 
No single metric family is sufficient to characterize EWM capability, and combining complementary measurements is necessary to distinguish visually plausible rollouts from physically and behaviorally correct ones.

\subsection{Threshold Sensitivity of RoboPhyscore}

Table \ref{tab:quantitative10} examines the sensitivity of \textit{RoboPhyscore} to the correlation threshold used for task-aligned metric selection.
As the threshold increases, the selected metric set becomes progressively more compact: 19 metrics are retained at $0.5$, 12 at $0.6$, eight at the default threshold of $0.7$, only one at $0.8$, and none at $0.9$.

At threshold $0.5$, \textit{RoboPhyscore} contains LPIPS, FVD, PSNR, and SSIM (IQ-1); HPSv3 (AQ); JEPA Similarity (DS); Flow Score (MQ); Semantics Consistency (CC); Average Precision and Average Recall (OL); nDTW (TA); Interaction Score (IQ-2); Physical Commonsense (PLA); BLEU, CLIP Score, Logics, VLM-1 Score, and Subject Adherence (SA); and Normalized Score (IF).
Increasing the threshold to $0.6$ reduces the set to 12 metrics: LPIPS and FVD (IQ-1), JEPA Similarity (DS), Semantics Consistency (CC), Average Precision and Average Recall (OL), nDTW (TA), Interaction Score (IQ-2), Physical Commonsense (PLA), VLM-1 Score and Subject Adherence (SA), and Normalized Score (IF).
At the default threshold of $0.7$, eight metrics remain: FVD (IQ-1), JEPA Similarity (DS), Semantics Consistency (CC), Average Recall (OL), nDTW (TA), Physical Commonsense (PLA), Subject Adherence (SA), and Normalized Score (IF).
At $0.8$, only JEPA Similarity remains, while no metric satisfies the selection criterion at $0.9$, making \textit{RoboPhyscore} undefined.

The alignment results reveal a trade-off between linear calibration and rank consistency.
For human evaluation, the $0.7$ threshold achieves the highest Pearson correlation ($r=0.9761$), the highest linear-fit $R^2=0.9528$, and the lowest MAE ($0.0317$), while $0.8$ yields the highest Spearman correlation ($\rho=0.9692$). 
A similar pattern is observed for VLM-2 Score: threshold $0.7$ provides the strongest Pearson correlation ($0.8696$), $R^2$ ($0.7563$), and lowest MAE ($0.1611$), whereas the single-metric $0.8$ setting produces the highest rank correlation ($\rho=0.9505$).

For the two execution-grounded task-success signals, the most aggressive valid threshold performs best numerically. 
At $0.8$, JEPA Similarity alone reaches $r=0.9665$, $\rho=0.9717$, $R^2=0.9341$, and MAE $=0.0575$ for Action Planner, and $r=0.9616$, $\rho=0.9675$, $R^2=0.9246$, and MAE $=0.0373$ for Data Engine. 
However, this setting reduces RoboPhyscore to a single distribution-similarity metric and therefore sacrifices the multi-dimensional coverage and interpretability intended by the unified score. 
Conversely, lower thresholds retain substantially more metrics but provide weaker human and VLM-2 linear alignment.

Finally, We adopt $0.7$ as the default threshold. 
It retains eight complementary metrics spanning perceptual, state, semantic, physical, and task-level properties while achieving the best human-alignment and VLM-2 results and maintaining strong correlations with both execution-grounded task-success measures. 
The sensitivity analysis thus indicates that the selected threshold provides a favorable balance between task alignment and metric diversity rather than simply maximizing a single correlation statistic.

\begin{table}[t]
\centering
\caption{Parameter sensitivity of RoboPhyscore under different threshold settings.}
\small
\begin{tabular}{lccccc}
\toprule
\textbf{Threshold} & \textbf{Pearson $r$ $\uparrow$} & \textbf{Spearman $\rho$ $\uparrow$} & \textbf{$R^2$ $\uparrow$} & \textbf{MAE $\downarrow$} & \textbf{\# of Metrics} \\
\midrule
\multicolumn{6}{l}{\textit{Human Alignment}} \\
\midrule
0.5 & 0.9592 & 0.9200 & 0.9201 & 0.0419 & 19 \\
0.6 & 0.9649 & 0.9023 & 0.9310 & 0.0404 & 12 \\
0.7 & \textbf{0.9761} & 0.8962 & \textbf{0.9528} & \textbf{0.0317} & 8 \\
0.8 & 0.9454 & \textbf{0.9692} & 0.8938 & 0.0499 & 1 \\
0.9 & N/A & N/A & N/A & N/A & 0 \\
\midrule
\multicolumn{6}{l}{\textit{Action Planner (TS)}} \\
\midrule
0.5 & 0.9478 & 0.8748 & 0.8984 & 0.0785 & 19 \\
0.6 & 0.9572 & 0.8656 & 0.9162 & 0.0726 & 12 \\
0.7 & 0.9551 & 0.8586 & 0.9122 & 0.0749 & 8 \\
0.8 & \textbf{0.9665} & \textbf{0.9717} & \textbf{0.9341} & \textbf{0.0575} & 1 \\
0.9 & N/A & N/A & N/A & N/A & 0 \\
\midrule
\multicolumn{6}{l}{\textit{Data Engine (TS)}} \\
\midrule
0.5 & 0.9378 & 0.8632 & 0.8794 & 0.0480 & 19 \\
0.6 & 0.9483 & 0.8548 & 0.8993 & 0.0439 & 12 \\
0.7 & 0.9461 & 0.8475 & 0.8950 & 0.0469 & 8 \\
0.8 & \textbf{0.9616} & \textbf{0.9675} & \textbf{0.9246} & \textbf{0.0373} & 1 \\
0.9 & N/A & N/A & N/A & N/A & 0 \\
\midrule
\multicolumn{6}{l}{\textit{VLM-2 (TS)}} \\
\midrule
0.5 & 0.8350 & 0.8316 & 0.6972 & 0.1811 & 19 \\
0.6 & 0.8430 & 0.8224 & 0.7107 & 0.1772 & 12 \\
0.7 & \textbf{0.8696} & 0.8132 & \textbf{0.7563} & \textbf{0.1611} & 8 \\
0.8 & 0.8290 & \textbf{0.9505} & 0.6872 & 0.1936 & 1 \\
0.9 & N/A & N/A & N/A & N/A & 0 \\
\bottomrule
\end{tabular}
\label{tab:quantitative10}
\end{table}

\section{Supplementary Details about Qualitative Results}
\label{appendix:qualitative}

Figures \ref{fig10}--\ref{fig12} extend the qualitative comparison along the three remaining experimental axes: the reconstruction method, the textual conditioning, and the IDM.

\subsection{Effect of the Reconstruction Method}

\begin{figure*}[ht]
  \centering
  \includegraphics[width=\linewidth]{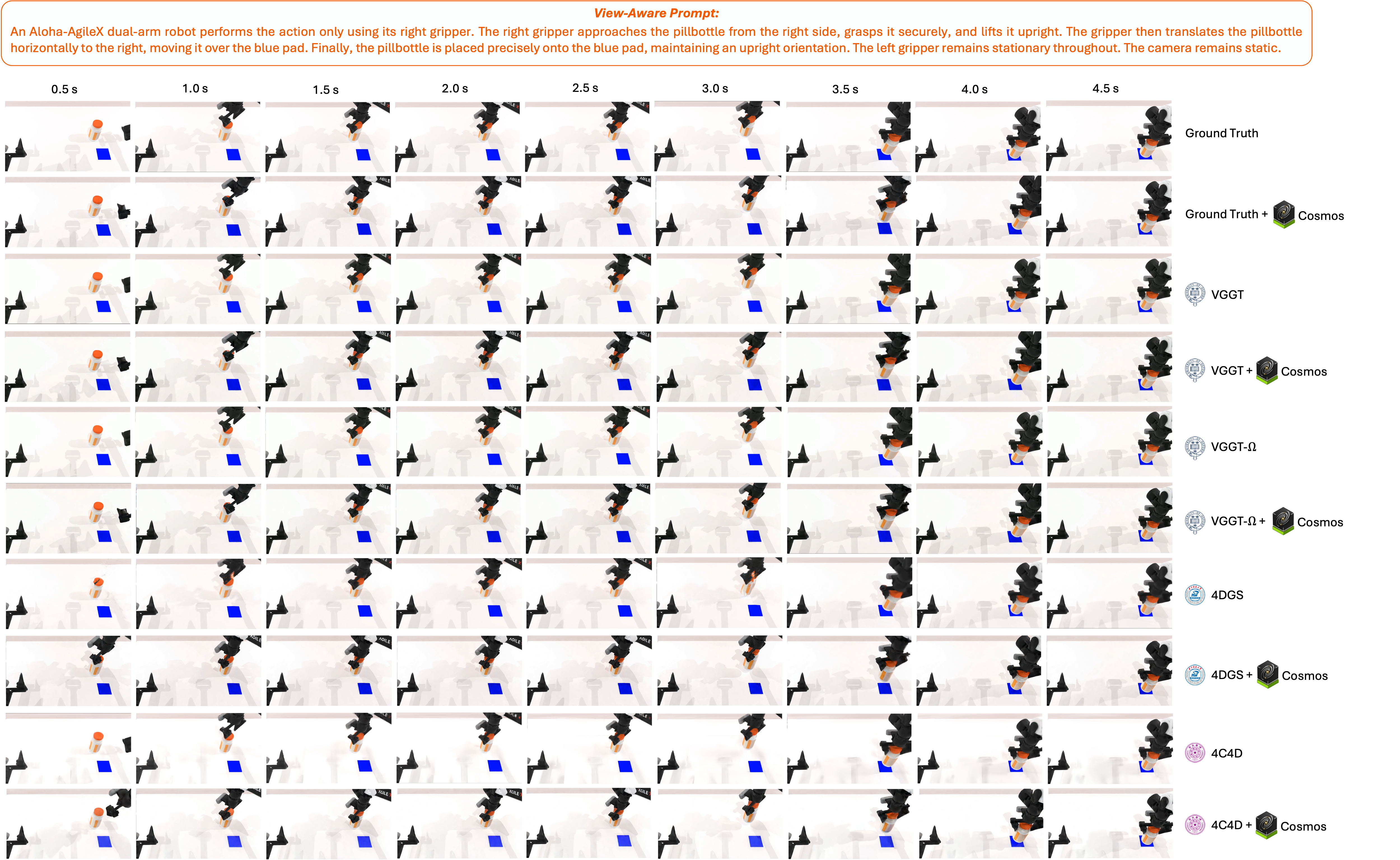}
  \caption{Qualitative examples of \textit{Move Pillbottle Pad} task (Cosmos + different reconstruction methods + view-aware prompt).}
  \label{fig10}
\end{figure*}

Figure \ref{fig10} shows the \textit{Move Pillbottle Pad} task under the view-aware prompt, with rows arranged in pairs: each reconstruction method is followed by the Cosmos 3 rollout conditioned on it. 
The substrates themselves differ in appearance fidelity: the feed-forward reconstructions remain close to the raw video, whereas 4DGS and 4C4D exhibit reduced contrast, faded background structure, and a visibly weakened rendering of the target pillbottle pad. 
In addition, the rollouts inherit the appearance of their conditioning substrate: Cosmos 3 preserves the instructed right-gripper trajectory and reaches the target region under every substrate. 
The example illustrates that motion intent survives reconstruction largely intact, while the appearance degradations of the dynamic representations erode exactly the reference-grounded quantities, such as object localization and subject adherence, that separate the feed-forward substrates from 4C4D and 4DGS. 
The degraded rendering also indicates that part of the loss is reconstruction-induced rather than generation-induced, since it afflicts the substrate before any generation occurs, which is the confound the reconstruction-matched protocol is designed to separate.

\subsection{Effect of Prompt Specificity}

\begin{figure*}[ht]
  \centering
  \includegraphics[width=\linewidth]{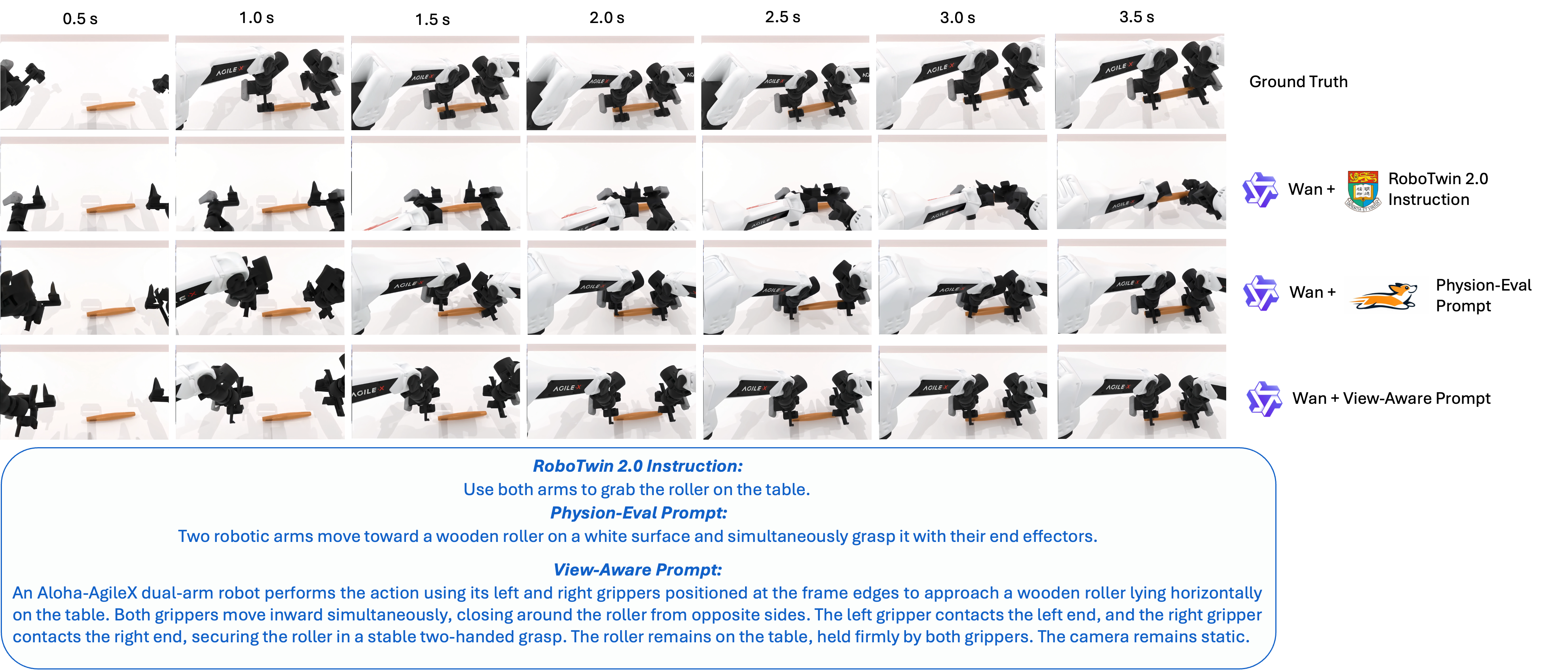}
  \caption{Qualitative examples of \textit{Grab Roller} task (Wan + ground truth + different prompts).}
  \label{fig11}
\end{figure*}

Figure \ref{fig11} shows the \textit{Grab Roller} task for Wan conditioned on the ground-truth substrate under the three annotation levels. 
Under the native RoboTwin 2.0 instruction, the embodiment morphs away from the Aloha-AgileX arms, transient streak artifacts appear near the roller, and the sequence ends with the roller displaced rightward without a stable bimanual grasp. 
The Physion-Eval caption restores a coherent two-arm approach, but the roller is tilted and partially lifted mid-sequence, and the terminal grasp configuration deviates from the demonstration. 
The view-aware caption yields the rollout closest to ground truth: both grippers enter from the frame edges as described, converge symmetrically, and secure the horizontal roller in a stable two-handed grasp while the embodiment remains intact. 
The progression matches the ablation study, in which Wan improves from 0.4962 to 0.5474 to 0.5782 across the three conditions, and indicates that the gain arises from the explicit spatial anchors of the view-aware caption, whose clauses on gripper laterality, approach direction, and the roller remaining on the table correspond directly to the failure modes visible under the sparser conditions.

\subsection{Effect of the Inverse Dynamics Model}

\begin{figure*}[ht]
  \centering
  \includegraphics[width=\linewidth]{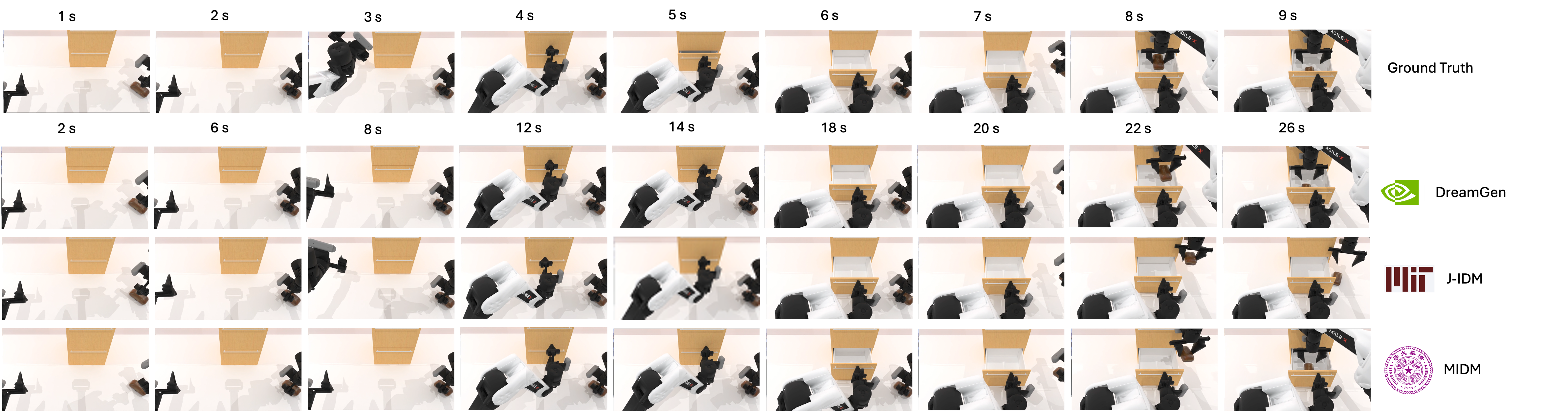}
  \caption{Qualitative examples of \textit{Put Object Cabinet} task (ground truth + different inverse dynamic models).}
  \label{fig12}
\end{figure*}

Figure \ref{fig12} shows the \textit{Put Object Cabinet} task, comparing the ground-truth demonstration with simulator replays of the action sequences that the three IDMs decode from that same demonstration. 
All three IDMs recover the phase structure of the demonstration, comprising the approach, the opening of the cabinet, the transport of the object, and the insertion, so the differences concentrate in end-effector precision during the contact-rich segments. 
The DreamGen replay tracks the demonstration most closely and terminates with the object placed in the opened compartment; the J-IDM replay accumulates the largest late-episode deviations, with the arm departing from the demonstrated pose above the cabinet and an ambiguous final placement; and the MIDM replay lies between the two. 
This ordering is consistent with quantitative results and supports the adoption of DreamGen for all execution-grounded evaluations. 

\section{Supplementary Details about Human Study Protocol}
\label{appendix:human}

\textbf{Participants and ethics.}
We recruited 70 participants through internal volunteer sign-up. 
All participants had normal or corrected-to-normal vision and had not been exposed to the evaluated samples or the design of RoboPhyscore. 
The study was conducted in accordance with institutional guidelines and involved only the evaluation of pre-generated robot manipulation videos without collecting sensitive personal information. 
Participants were not informed of the corresponding world model, reconstruction method, or any automatic evaluation score associated with any sample.

\textbf{Interface and rating procedure.}
Participants interacted with a web-based interface that displayed each video together with the corresponding task instruction at the same spatial resolution, playback rate, and visualization format for all methods. 
The identities of the video world model and reconstruction method were hidden, and the presentation order of all samples was randomized independently for each participant.
For each sample, participants provided an overall human evaluation score on a 10-point scale reflecting the quality of the generated embodied interaction. 
Participants were asked to make a holistic judgment of whether the generated sequence represented a coherent, physically plausible, and successfully executed robot interaction. 
This design provides an external human criterion that is independent of the metric composition and aggregation procedure used by the automatic scores.

\textbf{Sample selection and task assignment.}
Human evaluation covered samples across all evaluated video world models, reconstruction methods, and four task categories. 
Samples were selected using stratified sampling to balance the coverage of different world models, reconstruction methods, task categories, and RoboPhyscore ranges. 
In particular, the evaluation set included samples spanning low, medium, and high RoboPhyscore values to avoid artificially inflating correlation by evaluating only clearly successful or clearly failed generations.
Each sample was independently evaluated by at least three participants. 
Task assignments were randomized across participants while maintaining approximately balanced annotation coverage. 
Ground-truth samples were presented in the same manner as generated samples and were not identified as ground truth to the participants.

\textbf{Instructions.}
Before the formal evaluation, participants completed a brief written tutorial and several practice examples. 
They were instructed to assess the complete sequence rather than individual frames and to consider the overall coherence, physical plausibility, and successful execution of the intended robot interaction.
Participants were not instructed to explicitly evaluate the individual metrics, sub-dimensions, or hierarchical levels used to construct RoboPhyscore. 
No information regarding the automatic evaluation results, model identities, reconstruction methods, or expected rankings was provided. 
This blind evaluation protocol was designed to minimize anchoring effects and ensure that the resulting human scores served as an independent reference for validating RoboPhyscore.

\textbf{Data collection and aggregation.}
For each evaluated sample and participant, we recorded the anonymized sample identifier, human evaluation score, evaluation completion time, and task assignment information. 
For each sample, the final human reference score was computed by averaging the ratings from all independent annotators:
\begin{equation}
H_i = \frac{1}{K_i}\sum_{k=1}^{K_i} h_{ik},
\end{equation}
where $h_{ik}$ denotes the score assigned by annotator $k$ to sample $i$, and $K_i$ is the number of valid annotations for that sample.

The resulting human scores were subsequently compared with RoboPhyscore and other unified benchmark scores using Pearson correlation, Spearman rank correlation, coefficient of determination ($R^2$), and mean absolute error (MAE). 
Pearson correlation measures linear agreement with human judgments, while Spearman correlation evaluates consistency in sample ranking. 
$R^2$ quantifies the proportion of variation in human scores explained by each unified score, and MAE measures absolute agreement after different unified scores are linearly rescaled to a common range according to their theoretical score ranges.
All correlation analyses were performed at the sample level rather than only on model-level averages, thereby avoiding artificially high correlations caused by the small number of evaluated model families. 
Together, these measurements quantify both the ranking consistency and absolute agreement between the composite RoboPhyscore and independent human judgments.

\subsection{Comparison of Different Unified Scores}

We further compare AFS and RoboPhyscore with the unified scores of WorldModelBench, EWMBench, and WorldArena. 
Table \ref{tab:quantitative7} reports the five scores accordingly.
Because these unified scores have different native ranges, absolute-error comparison requires a common scale. 
For comparison with human evaluation, each score $S$ is linearly mapped to $[0,10]$ according to its theoretical range $[S_{\min},S_{\max}]$:
\begin{equation}
S^{(10)}
=
10\frac{S-S_{\min}}{S_{\max}-S_{\min}}.
\end{equation}
Accordingly, AFS, RoboPhyscore, WorldModelBench, EWMBench, and WorldArena are rescaled from $[0,4]$, $[0,1]$, $[0,10]$, $[0,8]$, and $[0,100]$, respectively. 
For task success comparison, the same transformation maps all unified scores to $[0,1]$, matching the native ranges of Action Planner, Data Engine, and VLM-2. 
Pearson $r$ and Spearman $\rho$ are invariant to positive linear rescaling and are computed directly from the configuration-level scores. 
We additionally report the linear-fit $R^2$ from ordinary least-squares regression with an intercept, together with MAE on the corresponding common scale.

\begin{table}[t]
\centering
\caption{Different unified scores for each video world model and each reconstruction method.}
\small
\begin{tabular}{lccccc}
\toprule
\textbf{Method} & \textbf{Ground Truth} & \textbf{VGGT} & \textbf{VGGT-$\Omega$} & \textbf{4DGS} & \textbf{4C4D} \\
\midrule
\multicolumn{6}{l}{\textit{Average Full Score (/4)}} \\
\midrule
Ground Truth & 3.1563 & 2.9475 & 2.9777 & 2.4591 & 2.6492 \\
Wan          & 2.7320 & 2.5793 & 2.6126 & 2.4454 & 2.4878 \\
CogVideoX    & 2.2774 & 2.2081 & 2.2573 & 2.2666 & 2.2491 \\
Cosmos       & 2.9024 & 2.7977 & 2.8234 & 2.6321 & 2.7431 \\
RoboDreamer  & 2.1318 & 2.1134 & 2.1280 & 2.1292 & 2.1206 \\
\midrule
\multicolumn{6}{l}{\textit{RoboPhyscore (/1)}} \\
\midrule
Ground Truth & 0.8361 & 0.7444 & 0.7592 & 0.4660 & 0.6088 \\
Wan          & 0.5782 & 0.5559 & 0.5609 & 0.4816 & 0.5029 \\
CogVideoX    & 0.3473 & 0.3269 & 0.3366 & 0.3387 & 0.3322 \\
Cosmos       & 0.6800 & 0.6649 & 0.6705 & 0.5530 & 0.5965 \\
RoboDreamer  & 0.3503 & 0.3519 & 0.3549 & 0.3453 & 0.3434 \\
\midrule
\multicolumn{6}{l}{\textit{WorldModelBench (/10)}} \\
\midrule
Ground Truth & 7.9925 & 7.5050 & 7.5800 & 4.9000 & 6.0475 \\
Wan          & 7.7575 & 7.3775 & 7.3500 & 5.9050 & 7.0850 \\
CogVideoX    & 6.3950 & 5.7475 & 5.9225 & 5.0250 & 5.4250 \\
Cosmos       & 7.0200 & 6.4850 & 6.5800 & 6.3725 & 6.9575 \\
RoboDreamer  & 5.0550 & 4.9100 & 4.9575 & 4.8150 & 4.8600 \\
\midrule
\multicolumn{6}{l}{\textit{EWMBench (/8)}} \\
\midrule
Ground Truth & 7.7643 & 4.7560 & 4.9052 & 3.5773 & 4.0176 \\
Wan          & 4.1620 & 3.9178 & 3.9437 & 3.5784 & 3.5262 \\
CogVideoX    & 4.4773 & 3.4594 & 3.5689 & 3.4862 & 3.3462 \\
Cosmos       & 3.7839 & 4.3867 & 4.3438 & 3.8269 & 3.8553 \\
RoboDreamer  & 3.2608 & 3.3619 & 3.2975 & 3.2134 & 3.2261 \\
\midrule
\multicolumn{6}{l}{\textit{WorldArena (/100)}} \\
\midrule
Ground Truth & 79.6741 & 77.6056 & 78.1523 & 57.7767 & 67.6977 \\
Wan          & 71.4137 & 69.9265 & 70.1120 & 68.8051 & 69.9834 \\
CogVideoX    & 54.3010 & 51.2716 & 50.9231 & 53.1802 & 52.8569 \\
Cosmos       & 74.0892 & 72.9803 & 73.2502 & 69.3486 & 71.7103 \\
RoboDreamer  & 58.7084 & 58.0006 & 58.0139 & 57.8532 & 57.9862 \\
\bottomrule
\end{tabular}
\label{tab:quantitative7}
\end{table}

\subsubsection{Human Alignment}

As shown in Table \ref{tab:quantitative8}, both proposed scores exhibit stronger agreement with human evaluation than the three existing unified scores. 
RoboPhyscore achieves the numerically highest Pearson correlation ($r=0.9761$) and $R^2=0.9528$. 
AFS achieves nearly identical linear agreement ($r=0.9746$, $R^2=0.9499$) and the highest rank correlation ($\rho=0.9715$) and the lowest MAE ($0.0284$). 
In comparison, the strongest external baseline, WorldArena, reaches $r=0.9263$, while WorldModelBench and EWMBench obtain $r=0.8090$ and $0.6417$, respectively. 
EWMBench's comparatively high rank correlation ($\rho=0.8669$) alongside its low Pearson value indicates a monotone but nonlinear relation to human judgment. 
The two proposed scores are complementary: RoboPhyscore is the marginally stronger linear predictor of human judgment, while AFS preserves the human ordering of the configurations most faithfully and is the best calibrated under the common scale, consistent with its comprehensive coverage of all evaluation dimensions.

\begin{table}[t]
\centering
\caption{Agreement between different unified scores and human evaluation.}
\small
\begin{tabular}{lcccc}
\toprule
\textbf{Unified Score} & \textbf{Pearson $r$ $\uparrow$} & \textbf{Spearman $\rho$ $\uparrow$} & \textbf{$R^2$ $\uparrow$} & \textbf{MAE $\downarrow$} \\
\midrule
Average Full Score & 0.9746 & \textbf{0.9715} & 0.9499 & \textbf{0.0284} \\
RoboPhyscore & \textbf{0.9761} & 0.8962 & \textbf{0.9528} & 0.0317 \\
WorldModelBench & 0.8090 & 0.8200 & 0.6545 & 0.0868 \\
EWMBench & 0.6417 & 0.8669 & 0.4117 & 0.1282 \\
WorldArena & 0.9263 & 0.8046 & 0.8580 & 0.0497 \\
\bottomrule
\end{tabular}
\label{tab:quantitative8}
\end{table}

\subsubsection{Task Alignment}

Table \ref{tab:quantitative9} further examines agreement with the three task success measurements used to characterize downstream embodied utility.
For Action Planner, RoboPhyscore achieves the highest Pearson correlation ($r=0.9551$), and $R^2=0.9122$, while AFS attains the highest Spearman correlation ($\rho=0.9525$) and lowest MAE ($0.0732$).
The same pattern largely holds for Data Engine: RoboPhyscore provides the strongest linear association ($r=0.9461$, $R^2=0.8950$), and AFS preserves the ranking ($\rho=0.9386$) and MAE ($0.0462$). 
For VLM-2, AFS performs best on all four measures.
The external scores trail on every measure and every target, with WorldArena again the strongest among them. 
Overall, RoboPhyscore is the sharper linear predictor of the execution-grounded outcomes, whereas AFS provides stronger rank-order consistency and calibration throughout and is uniformly best against the VLM-based judgment.
The margins between the two proposed scores are small, and the division of labor matches their intended roles, with AFS summarizing the complete taxonomy and RoboPhyscore emphasizing task-aligned constituents.

\begin{table}[t]
\centering
\caption{Agreement between different unified scores and task success evaluation.}
\small
\begin{tabular}{lcccc}
\toprule
\textbf{Unified Score} & \textbf{Pearson $r$ $\uparrow$} & \textbf{Spearman $\rho$ $\uparrow$} & \textbf{$R^2$ $\uparrow$} & \textbf{MAE $\downarrow$} \\
\midrule
\multicolumn{5}{l}{\textit{Action Planner}} \\
\midrule
Average Full Score & 0.9503 & \textbf{0.9525} & 0.9031 & \textbf{0.0732} \\
RoboPhyscore & \textbf{0.9551} & 0.8586 & \textbf{0.9122} & 0.0749 \\
WorldModelBench & 0.7217 & 0.7359 & 0.5209 & 0.1992 \\
EWMBench & 0.6505 & 0.8175 & 0.4231 & 0.2279 \\
WorldArena & 0.8552 & 0.7475 & 0.7313 & 0.1500 \\
\midrule
\multicolumn{5}{l}{\textit{Data Engine}} \\
\midrule
Average Full Score & 0.9411 & \textbf{0.9386} & 0.8858 & \textbf{0.0462} \\
RoboPhyscore & \textbf{0.9461} & 0.8475 & \textbf{0.8950} & 0.0469 \\
WorldModelBench & 0.7033 & 0.7059 & 0.4946 & 0.1237 \\
EWMBench & 0.6386 & 0.8048 & 0.4079 & 0.1398 \\
WorldArena & 0.8435 & 0.7236 & 0.7114 & 0.0905 \\
\midrule
\multicolumn{5}{l}{\textit{VLM-2}} \\
\midrule
Average Full Score & \textbf{0.8861} & \textbf{0.9098} & \textbf{0.7852} & \textbf{0.1501} \\
RoboPhyscore & 0.8696 & 0.8132 & 0.7563 & 0.1611 \\
WorldModelBench & 0.7611 & 0.6754 & 0.5792 & 0.1947 \\
EWMBench & 0.4771 & 0.7570 & 0.2277 & 0.3319 \\
WorldArena & 0.8484 & 0.6904 & 0.7198 & 0.1586 \\
\bottomrule
\end{tabular}
\label{tab:quantitative9}
\end{table}

\section{Supplementary Details about Quantitative Results}
\label{appendix:experiment}

\subsection{Pixel-Level Fidelity}

\subsubsection{Image Quality}

\begin{table}[t]
\centering
\caption{Image quality score comparison of different world models.}
\small
\begin{tabular}{lccccccc}
\toprule
\textbf{Method} & \textbf{MUSIQ} & \textbf{PSNR} & \textbf{SSIM} & \textbf{LPIPS} & \textbf{FVD} & \textbf{Inception Score} & \textbf{IQ-1} \\
\midrule
\textit{RoboTwin} & & & & & & & \textit{4.3156/6} \\
\midrule
GT                 & 0.6135 & 1.0000 & 1.0000 & 1.0000 & 1.0000 & 0.2240 & 4.8375 \\
VGGT               & 0.5579 & 0.6912 & 0.9711 & 0.9262 & 0.9973 & 0.1922 & 4.3359 \\
VGGT-$\Omega$      & 0.5765 & 0.7188 & 0.9758 & 0.9501 & 0.9979 & 0.1853 & 4.4044 \\
4DGS               & 0.3907 & 0.4892 & 0.9052 & 0.8241 & 0.9446 & 0.5588 & 4.1126 \\
4C4D               & 0.4122 & 0.5576 & 0.9560 & 0.8413 & 0.9198 & 0.2007 & 3.8876 \\
\midrule
\textit{Wan} & & & & & & & \textit{3.5108/6} \\
\midrule
GT                 & 0.6596 & 0.2366 & 0.7814 & 0.6782 & 0.8118 & 0.5235 & 3.6911 \\
VGGT               & 0.6128 & 0.2285 & 0.7481 & 0.6469 & 0.8002 & 0.3705 & 3.4070 \\
VGGT-$\Omega$      & 0.6242 & 0.2316 & 0.7504 & 0.6538 & 0.7991 & 0.3887 & 3.4478 \\
4DGS               & 0.5386 & 0.2126 & 0.7381 & 0.6282 & 0.7898 & 0.6751 & 3.5824 \\
4C4D               & 0.5556 & 0.2077 & 0.7760 & 0.6181 & 0.7683 & 0.4998 & 3.4255 \\
\midrule
\textit{CogVideoX} & & & & & & & \textit{3.0803/6} \\
\midrule
GT                 & 0.5669 & 0.2179 & 0.7519 & 0.6262 & 0.5013 & 0.3976 & 3.0618 \\
VGGT               & 0.4858 & 0.2134 & 0.7311 & 0.5979 & 0.4717 & 0.3669 & 2.8668 \\
VGGT-$\Omega$      & 0.4889 & 0.2150 & 0.7291 & 0.6006 & 0.4412 & 0.5218 & 2.9966 \\
4DGS               & 0.4448 & 0.2152 & 0.7248 & 0.6131 & 0.4633 & 0.8328 & 3.2940 \\
4C4D               & 0.4544 & 0.2130 & 0.8040 & 0.6168 & 0.4775 & 0.6168 & 3.1825 \\
\midrule
\textit{Cosmos} & & & & & & & \textit{3.6128/6} \\
\midrule
GT                 & 0.6243 & 0.3490 & 0.8590 & 0.8133 & 0.9332 & 0.1459 & 3.7247 \\
VGGT               & 0.5781 & 0.3374 & 0.8709 & 0.7856 & 0.9274 & 0.1230 & 3.6224 \\
VGGT-$\Omega$      & 0.5946 & 0.3383 & 0.8715 & 0.7896 & 0.9275 & 0.1253 & 3.6468 \\
4DGS               & 0.5585 & 0.3131 & 0.8315 & 0.7426 & 0.9046 & 0.2324 & 3.5827 \\
4C4D               & 0.5916 & 0.3159 & 0.8557 & 0.7388 & 0.8747 & 0.1106 & 3.4873 \\
\midrule
\textit{RoboDreamer} & & & & & & & \textit{3.2018/6} \\
\midrule
GT                 & 0.4865 & 0.2208 & 0.7793 & 0.6396 & 0.7683 & 0.3323 & 3.2268 \\
VGGT               & 0.4631 & 0.2192 & 0.7834 & 0.6338 & 0.7771 & 0.3246 & 3.2012 \\
VGGT-$\Omega$      & 0.4699 & 0.2196 & 0.7833 & 0.6356 & 0.7739 & 0.3436 & 3.2259 \\
4DGS               & 0.4354 & 0.2142 & 0.7804 & 0.6277 & 0.7741 & 0.3775 & 3.2093 \\
4C4D               & 0.4462 & 0.2165 & 0.7776 & 0.6235 & 0.7709 & 0.3112 & 3.1459 \\
\bottomrule
\end{tabular}
\label{tab:comparison1}
\end{table}

\begin{table}[t]
\centering
\caption{Image quality score comparison of different prompts.}
\small
\begin{tabular}{lccccccc}
\toprule
\textbf{Method} & \textbf{MUSIQ} & \textbf{PSNR} & \textbf{SSIM} & \textbf{LPIPS} & \textbf{FVD} & \textbf{Inception Score} & \textbf{IQ-1} \\
\midrule
\textit{Wan} & & & & & & & \textit{3.5599/6} \\
\midrule
GT+Instruction        & 0.6345 & 0.2264 & 0.7645 & 0.6596 & 0.7741 & 0.3907 & 3.4498 \\
GT+Previous Prompt    & 0.6523 & 0.2326 & 0.7768 & 0.6688 & 0.8009 & 0.4075 & 3.5389 \\
GT+Current Prompt     & 0.6596 & 0.2366 & 0.7814 & 0.6782 & 0.8118 & 0.5235 & 3.6911 \\
\midrule
\textit{Cosmos} & & & & & & & \textit{3.7059/6} \\
\midrule
GT+Instruction        & 0.6237 & 0.3436 & 0.8526 & 0.8054 & 0.9296 & 0.1412 & 3.6961 \\
GT+Previous Prompt    & 0.6237 & 0.3462 & 0.8562 & 0.8094 & 0.9296 & 0.1317 & 3.6968 \\
GT+Current Prompt     & 0.6243 & 0.3490 & 0.8590 & 0.8133 & 0.9332 & 0.1459 & 3.7247 \\
\bottomrule
\end{tabular}
\label{tab:comparison2}
\end{table}

\paragraph{Impact of Reconstruction Quality.}
The RoboTwin reference experiments reveal a clear reconstruction-quality hierarchy: $\text{GT} > \text{VGGT-}\Omega > \text{VGGT} > \text{4DGS} > \text{4C4D}$, with IQ-1 scores of $4.8375$, $4.4044$, $4.3359$, $4.1126$, and $3.8876$, respectively. 
Relative to GT, VGGT-$\Omega$ and VGGT retain $91.0\%$ and $89.6\%$ of IQ-1, compared with $85.0\%$ for 4DGS and $80.4\%$ for 4C4D, demonstrating the advantage of the feed-forward reconstruction methods as conditioning substrates. 
The degradation is strongly metric dependent: even when PSNR drops substantially for 4DGS and 4C4D ($0.4892$ and $0.5576$), SSIM remains comparatively high ($0.9052$ and $0.9560$), indicating that local structural information can be preserved despite considerable photometric mismatch. 
FVD further distinguishes the two 4D representations ($0.9446$ vs.\ $0.9198$), suggesting that reconstruction artifacts also affect clip-level distributional fidelity.
Moreover, generation substantially reduces sensitivity to the reconstruction substrate. 
The IQ-1 range across substrates decreases from $0.9499$ for RoboTwin to $0.2841$, $0.4272$, $0.2374$, and $0.0809$ for Wan, CogVideoX, Cosmos, and RoboDreamer, respectively. 
Thus, although better reconstruction generally provides a stronger input signal, the learned generative prior can partially suppress differences introduced by reconstruction.

\paragraph{Comparison of Different World Models.}
At the aggregate level, Cosmos achieves the highest IQ-1 among the evaluated world models at $3.6128/6$, followed by Wan ($3.5108$), RoboDreamer ($3.2018$), and CogVideoX ($3.0803$). 
These correspond to $83.7\%$, $81.4\%$, $74.2\%$, and $71.4\%$ of the RoboTwin reference score ($4.3156$), respectively. 
Cosmos' advantage is particularly consistent across reference-sensitive metrics, with mean PSNR, SSIM, LPIPS, and FVD scores of $0.3307$, $0.8577$, $0.7740$, and $0.9135$, outperforming the other generated models on these dimensions. 
Wan achieves the highest mean MUSIQ ($0.5982$), whereas CogVideoX obtains the highest mean IS ($0.5472$), further illustrating that perceptual quality or diversity alone does not guarantee faithful prediction of the reference future.
The interaction between reconstruction and generation is also model dependent. 
Cosmos preserves the same substrate ordering observed on RoboTwin, whereas Wan, CogVideoX, and RoboDreamer do not. 
Most notably, CogVideoX obtains its highest IQ-1 with 4DGS ($3.2940$) rather than GT ($3.0618$), driven largely by its unusually high IS. 
This result suggests that sufficiently strong generative priors can dominate the visual conditioning signal, making reconstruction quality and rollout quality related but not interchangeable quantities.

\paragraph{Prompt Specification.}
Prompt specification produces a consistent but model-dependent improvement. 
For Wan, IQ-1 increases monotonically from $3.4498$ with the instruction-only prompt to $3.5389$ with the previous prompt and $3.6911$ with the proposed current prompt. 
The resulting $0.2413$ improvement over the instruction baseline corresponds to a $7.0\%$ relative gain, with improvements observed across all six constituent metrics. 
Cosmos exhibits the same aggregate ordering, increasing from $3.6961$ to $3.6968$ and $3.7247$, but with a substantially smaller total gain of $0.0286$ ($0.8\%$). 
These results indicate that explicit, view-aware conditioning is particularly important for Wan, whereas Cosmos is comparatively robust to prompt formulation. 
Overall, visual reconstruction remains a major determinant of pixel-level fidelity, while richer language conditioning provides an additional improvement in generation quality without, by itself, establishing physical correctness.

\subsubsection{Aesthetic Quality}

\begin{table}[t]
\centering
\caption{Aesthetic quality score comparison of different world models.}
\small
\begin{tabular}{lccccc}
\toprule
\textbf{Method} & \textbf{LAION-AP} & \textbf{HPSv3} & \textbf{CLIP-AP} & \textbf{CLIP-IQA} & \textbf{AQ} \\
\midrule
\textit{RoboTwin} & & & & & \textit{1.3323/4} \\
\midrule
GT                 & 0.4192 & 0.2258 & 0.7231 & 0.4050 & 1.7731 \\
VGGT               & 0.4234 & 0.1655 & 0.5940 & 0.3260 & 1.5089 \\
VGGT-$\Omega$      & 0.4250 & 0.1898 & 0.6220 & 0.3480 & 1.5848 \\
4DGS               & 0.3351 & 0.0068 & 0.1985 & 0.2630 & 0.8034 \\
4C4D               & 0.3607 & 0.0257 & 0.3150 & 0.2897 & 0.9911 \\
\midrule
\textit{Wan} & & & & & \textit{1.6047/4} \\
\midrule
GT                 & 0.4183 & 0.2580 & 0.7231 & 0.4724 & 1.8718 \\
VGGT               & 0.4283 & 0.1945 & 0.6447 & 0.3876 & 1.6551 \\
VGGT-$\Omega$      & 0.4285 & 0.2069 & 0.6635 & 0.4117 & 1.7106 \\
4DGS               & 0.3988 & 0.1397 & 0.4162 & 0.3576 & 1.3123 \\
4C4D               & 0.4005 & 0.1590 & 0.5413 & 0.3731 & 1.4739 \\
\midrule
\textit{CogVideoX} & & & & & \textit{1.3855/4} \\
\midrule
GT                 & 0.4097 & 0.1559 & 0.6020 & 0.4731 & 1.6407 \\
VGGT               & 0.4078 & 0.0913 & 0.4719 & 0.4127 & 1.3837 \\
VGGT-$\Omega$      & 0.4120 & 0.1067 & 0.5023 & 0.4423 & 1.4633 \\
4DGS               & 0.3621 & 0.0909 & 0.3314 & 0.3623 & 1.1467 \\
4C4D               & 0.3790 & 0.0799 & 0.4557 & 0.3783 & 1.2929 \\
\midrule
\textit{Cosmos} & & & & & \textit{1.5019/4} \\
\midrule
GT                 & 0.4008 & 0.2214 & 0.6543 & 0.4297 & 1.7062 \\
VGGT               & 0.4020 & 0.1477 & 0.5786 & 0.3688 & 1.4971 \\
VGGT-$\Omega$      & 0.4040 & 0.1653 & 0.6086 & 0.3914 & 1.5693 \\
4DGS               & 0.3741 & 0.1327 & 0.4355 & 0.3425 & 1.2848 \\
4C4D               & 0.3767 & 0.1771 & 0.5335 & 0.3649 & 1.4522 \\
\midrule
\textit{RoboDreamer} & & & & & \textit{1.0521/4} \\
\midrule
GT                 & 0.3736 & 0.0241 & 0.4188 & 0.3476 & 1.1641 \\
VGGT               & 0.3693 & 0.0131 & 0.3661 & 0.3303 & 1.0788 \\
VGGT-$\Omega$      & 0.3695 & 0.0154 & 0.3730 & 0.3347 & 1.0926 \\
4DGS               & 0.3584 & 0.0016 & 0.2723 & 0.2985 & 0.9308 \\
4C4D               & 0.3569 & 0.0036 & 0.3200 & 0.3139 & 0.9944 \\
\bottomrule
\end{tabular}
\label{tab:comparison3}
\end{table}

\begin{table}[t]
\centering
\caption{Aesthetic quality score comparison of different prompts.}
\small
\begin{tabular}{lccccc}
\toprule
\textbf{Method} & \textbf{LAION-AP} & \textbf{HPSv3} & \textbf{CLIP-AP} & \textbf{CLIP-IQA} & \textbf{AQ} \\
\midrule
\textit{Wan} & & & & & \textit{1.8578/4} \\
\midrule
GT+Instruction      & 0.4138 & 0.2336 & 0.7104 & 0.4724 & 1.8302 \\
GT+Previous Prompt  & 0.4171 & 0.2479 & 0.7130 & 0.4768 & 1.8548 \\
GT+Current Prompt   & 0.4183 & 0.2580 & 0.7231 & 0.4889 & 1.8883 \\
\midrule
\textit{Cosmos} & & & & & \textit{1.6861/4} \\
\midrule
GT+Instruction      & 0.3989 & 0.2147 & 0.6502 & 0.4024 & 1.6662 \\
GT+Previous Prompt  & 0.3993 & 0.2147 & 0.6442 & 0.4277 & 1.6859 \\
GT+Current Prompt   & 0.4008 & 0.2214 & 0.6543 & 0.4297 & 1.7062 \\
\bottomrule
\end{tabular}
\label{tab:comparison4}
\end{table}

\paragraph{Impact of Reconstruction Quality.}
Aesthetic quality exhibits a clear and consistent dependence on reconstruction fidelity. 
On RoboTwin, the aggregate AQ follows $\text{GT} > \text{VGGT-}\Omega > \text{VGGT} > \text{4C4D} > \text{4DGS}$, with scores of $1.7731$, $1.5848$, $1.5089$, $0.9911$, and $0.8034$, respectively. 
Relative to GT, VGGT-$\Omega$ and VGGT retain $89.4\%$ and $85.1\%$ of the reference AQ, whereas 4C4D and 4DGS retain only $55.9\%$ and $45.3\%$. 
This degradation is substantially stronger than that observed for conventional image-quality metrics, indicating that aesthetic predictors are particularly sensitive to reconstruction artifacts that affect visual realism and perceptual preference.
The effect is highly metric dependent. 
For 4DGS, LAION-AP retains $79.9\%$ of the GT score and CLIP-IQA retains $64.9\%$, whereas CLIP-AP and HPSv3 decrease to only $27.5\%$ and $3.0\%$, respectively. 
HPSv3 therefore provides the strongest separation between high- and low-quality reconstruction substrates in this experiment. 
Notably, LAION-AP alone assigns slightly higher scores to VGGT and VGGT-$\Omega$ than to GT, while the remaining metrics strongly favor GT. 
This discrepancy further motivates aggregating complementary aesthetic predictors rather than relying on any individual metric.

\paragraph{Comparison of Different World Models.}
The world models exhibit markedly different aesthetic characteristics. 
Averaged across reconstruction substrates, Wan achieves the highest AQ at $1.6047/4$, followed by Cosmos ($1.5019$), CogVideoX ($1.3855$), RoboTwin ($1.3323$), and RoboDreamer ($1.0521$). 
Relative to the RoboTwin reference, Wan, Cosmos, and CogVideoX achieve AQ increases of approximately $20.4\%$, $12.7\%$, and $4.0\%$, respectively, whereas RoboDreamer is $21.0\%$ lower. 
These results illustrate an important distinction between aesthetic appeal and reference fidelity: a generated rollout can receive a higher aesthetic score than the corresponding real or reconstructed observation despite being less faithful to the reference future. 
Consequently, AQ should be interpreted as perceptual preference rather than evidence of more accurate world modeling.
Reconstruction quality nevertheless propagates consistently through generation. 
For all four world models, the aggregate ordering remains $\text{GT} > \text{VGGT-}\Omega > \text{VGGT} > \text{4C4D} > \text{4DGS}$.
At the same time, generation substantially compresses the effect of the reconstruction substrate: the AQ range decreases from $0.9697$ for RoboTwin to $0.5595$, $0.4940$, $0.4214$, and $0.2333$ for Wan, CogVideoX, Cosmos, and RoboDreamer, respectively. 
This contraction is particularly pronounced for the lower-quality 4D substrates. 
For example, Cosmos improves AQ over RoboTwin by $0.4814$ on 4DGS and $0.4611$ on 4C4D, while producing slightly lower scores on the three stronger substrates. 
This behavior suggests that the generative prior can partially regularize reconstruction artifacts and reduce sensitivity to degraded visual conditioning, although the magnitude of this effect is strongly model dependent.

\paragraph{Prompt Specification.}
Prompt specificity consistently improves aggregate aesthetic quality for both evaluated world models. 
For Wan, AQ increases monotonically from $1.8302$ with the instruction-only prompt to $1.8548$ with the previous prompt and $1.8883$ with the proposed current prompt, corresponding to a $3.2\%$ improvement over the instruction baseline. 
All four constituent metrics improve under the current prompt, including HPSv3 ($0.2336 !\to! 0.2580$), CLIP-AP ($0.7104 !\to! 0.7231$), and CLIP-IQA ($0.4724 !\to! 0.4889$). 
Cosmos exhibits the same aggregate trend, increasing from $1.6662$ to $1.6859$ and $1.7062$, yielding a $2.4\%$ improvement. 
Although its intermediate prompt does not improve every individual component, the proposed current prompt achieves the highest final score across all four metrics.
Overall, the prompt-induced gains ($+0.0581$ for Wan and $+0.0400$ for Cosmos) are considerably smaller than the corresponding variation caused by reconstruction substrates ($0.5595$ and $0.4214$). 
This suggests that visual conditioning remains the dominant factor governing aesthetic quality, while more informative prompting provides a consistent secondary benefit. 
Importantly, these improvements reflect enhanced perceptual plausibility and preference rather than physical or task-level correctness, which are evaluated separately in RoboPhys-3D.

\subsubsection{Distribution Similarity}

\begin{table}[t]
\centering
\caption{Distribution similarity score comparison of different world models.}
\small
\begin{tabular}{lc}
\toprule
\textbf{Method} & \textbf{JEPA Similarity (DS)} \\
\midrule
\textit{RoboTwin} & \textit{0.8142/1} \\
\midrule
GT                 & 1.0000 \\
VGGT               & 0.9683 \\
VGGT-$\Omega$      & 0.9664 \\
4DGS               & 0.5328 \\
4C4D               & 0.6036 \\
\midrule
\textit{Wan} & \textit{0.3602/1} \\
\midrule
GT                 & 0.4086 \\
VGGT               & 0.3896 \\
VGGT-$\Omega$      & 0.3903 \\
4DGS               & 0.3111 \\
4C4D               & 0.3012 \\
\midrule
\textit{CogVideoX} & \textit{0.1505/1} \\
\midrule
GT                 & 0.1426 \\
VGGT               & 0.1371 \\
VGGT-$\Omega$      & 0.1434 \\
4DGS               & 0.1957 \\
4C4D               & 0.1335 \\
\midrule
\textit{Cosmos} & \textit{0.6525/1} \\
\midrule
GT                 & 0.7585 \\
VGGT               & 0.7372 \\
VGGT-$\Omega$      & 0.7407 \\
4DGS               & 0.4770 \\
4C4D               & 0.5490 \\
\midrule
\textit{RoboDreamer} & \textit{0.0898/1} \\
\midrule
GT                 & 0.0915 \\
VGGT               & 0.0908 \\
VGGT-$\Omega$      & 0.0913 \\
4DGS               & 0.0925 \\
4C4D               & 0.0827 \\
\bottomrule
\end{tabular}
\label{tab:comparison5}
\end{table}

\begin{table}[t]
\centering
\caption{Distribution similarity score comparison of different prompts.}
\small
\begin{tabular}{lc}
\toprule
\textbf{Method} & \textbf{JEPA Similarity (DS)} \\
\midrule
\textit{Wan} & \textit{0.3746/1} \\
\midrule
GT+Instruction      & 0.3120 \\
GT+Previous Prompt  & 0.4033 \\
GT+Current Prompt   & 0.4086 \\
\midrule
\textit{Cosmos} & \textit{0.7478/1} \\
\midrule
GT+Instruction      & 0.7405 \\
GT+Previous Prompt  & 0.7444 \\
GT+Current Prompt   & 0.7585 \\
\bottomrule
\end{tabular}
\label{tab:comparison6}
\end{table}

\paragraph{Impact of Reconstruction Quality.}
Distribution similarity is highly sensitive to the reconstruction substrate. 
On RoboTwin, GT achieves the maximum DS of $1.0000$, while VGGT and VGGT-$\Omega$ retain $96.8\%$ and $96.6\%$ of this score, respectively. 
In contrast, 4C4D and 4DGS retain only $60.4\%$ and $53.3\%$. 
The separation between the weakest feed-forward reconstruction (VGGT-$\Omega$, $0.9664$) and the strongest 4D reconstruction (4C4D, $0.6036$) accounts for approximately $77.7\%$ of the total substrate range, revealing a pronounced distinction between the two reconstruction families. 
Compared with conventional pixel-level measures, this result suggests that artifacts introduced by 4D reconstruction affect not only appearance but also the higher-level spatiotemporal representations captured by the JEPA encoder.
The same-substrate evaluation is particularly important in this setting because reconstruction error can otherwise be conflated with generation error. 
For Wan, for example, the reduction from the RoboTwin substrate to the generated rollout is large on GT, VGGT, and VGGT-$\Omega$ ($0.5914$, $0.5787$, and $0.5761$, respectively), indicating that generation dominates the discrepancy on high-quality substrates. 
For 4DGS and 4C4D, however, a substantial portion of the total distance from the reference is already introduced by reconstruction itself. 
These results demonstrate the importance of evaluating world-model rollouts against matched reconstruction controls rather than attributing the entire representation-space discrepancy to generation.

\paragraph{Comparison of Different World Models.}
The evaluated world models exhibit substantial differences in their ability to preserve the reference spatiotemporal distribution. 
Cosmos achieves the highest mean DS of $0.6525$, retaining $80.1\%$ of the RoboTwin reference score ($0.8142$). 
Wan follows at $0.3602$ ($44.2\%$), while CogVideoX and RoboDreamer decrease substantially to $0.1505$ ($18.5\%$) and $0.0898$ ($11.0\%$), respectively. 
The large margin between Cosmos and the remaining models indicates that Cosmos generates rollouts whose learned spatiotemporal representations remain considerably closer to those of the reference episodes.
The effect of reconstruction quality is also strongly model dependent. 
For Cosmos, DS decreases from $0.7585$ on GT to $0.4770$--$0.5490$ on the 4D substrates, broadly preserving the distinction between high- and low-quality reconstruction families. 
Wan exhibits a similar separation, although differences within each reconstruction family are small. 
In contrast, CogVideoX obtains its highest DS with 4DGS ($0.1957$), while RoboDreamer varies by only $0.0098$ across all five substrates. 
These non-monotonic behaviors at low absolute similarity indicate that improved reconstruction quality does not necessarily translate into higher distribution similarity once the generative model substantially departs from the conditioning observation. 
Thus, substrate quality and world-model capability should be evaluated jointly rather than assuming a universally monotonic relationship between them.
Generation also generally compresses the variation induced by reconstruction. 
The substrate range of $0.4672$ for RoboTwin decreases to $0.2815$ for Cosmos and $0.1074$ for Wan, corresponding to reductions of approximately $39.7\%$ and $77.0\%$, respectively. 
The contraction is even stronger for CogVideoX and RoboDreamer, although their very low absolute DS values suggest that this reduced sensitivity primarily reflects dominance of the generative prior rather than successful preservation of the conditioning signal.

\paragraph{Prompt Specification.}
Prompt specification improves distribution similarity for both Wan and Cosmos, but with markedly different magnitudes. 
For Wan, DS increases monotonically from $0.3120$ with the instruction-only prompt to $0.4033$ with the previous prompt and $0.4086$ with the proposed current prompt. 
The overall improvement of $0.0966$ corresponds to a $31.0\%$ relative increase over the instruction baseline. 
Notably, the transition from the instruction to the previous prompt accounts for $94.5\%$ of the total improvement, indicating that most of the benefit arises from replacing the generic instruction with a richer description, while the proposed view-aware refinement provides a smaller additional gain.
Cosmos exhibits the same monotonic ordering but is substantially less sensitive to prompt formulation, increasing from $0.7405$ to $0.7444$ and $0.7585$. 
The total gain is $0.0180$, or approximately $2.4\%$ relative to the instruction baseline. 
Together, these results suggest that explicit language conditioning is particularly important for Wan, whereas Cosmos maintains high distribution similarity even under less specific prompts. 
Because DS measures agreement with the reference episode in a learned spatiotemporal representation space, these improvements provide stronger evidence of reference-consistent generation than no-reference perceptual metrics such as aesthetic quality. 
Nevertheless, JEPA similarity does not directly verify geometric or physical correctness; these properties are therefore assessed separately by the geometry-, state-, and task-level dimensions of RoboPhys-3D.

\subsubsection{Motion Quality}

\begin{table}[t]
\centering
\caption{Motion quality score comparison of different world models.}
\small
\begin{tabular}{lccccc}
\toprule
\textbf{Method} & \textbf{Dynamic Degree} & \textbf{Flow Score} & \textbf{Motion Smoothness} & \textbf{Temporal Flickering} & \textbf{MQ} \\
\midrule
\textit{RoboTwin} & & & & & \textit{2.5848/4} \\
\midrule
GT                 & 0.5786 & 0.3042 & 0.7651 & 0.9834 & 2.6313 \\
VGGT               & 0.5743 & 0.2966 & 0.7627 & 0.9834 & 2.6170 \\
VGGT-$\Omega$      & 0.5746 & 0.2945 & 0.7637 & 0.9834 & 2.6162 \\
4DGS               & 0.5175 & 0.2119 & 0.6436 & 0.9855 & 2.3585 \\
4C4D               & 0.6196 & 0.4206 & 0.6796 & 0.9814 & 2.7012 \\
\midrule
\textit{Wan} & & & & & \textit{3.1620/4} \\
\midrule
GT                 & 0.7516 & 0.6043 & 0.8792 & 0.9702 & 3.2053 \\
VGGT               & 0.7065 & 0.5245 & 0.8795 & 0.9689 & 3.0794 \\
VGGT-$\Omega$      & 0.7041 & 0.5065 & 0.8739 & 0.9696 & 3.0541 \\
4DGS               & 0.7629 & 0.5504 & 0.9123 & 0.9653 & 3.1909 \\
4C4D               & 0.7829 & 0.6116 & 0.9208 & 0.9650 & 3.2803 \\
\midrule
\textit{CogVideoX} & & & & & \textit{2.0245/4} \\
\midrule
GT                 & 0.3665 & 0.0883 & 0.6181 & 0.9887 & 2.0616 \\
VGGT               & 0.3322 & 0.0835 & 0.6062 & 0.9895 & 2.0114 \\
VGGT-$\Omega$      & 0.3232 & 0.0773 & 0.5916 & 0.9901 & 1.9822 \\
4DGS               & 0.3309 & 0.0830 & 0.6241 & 0.9894 & 2.0274 \\
4C4D               & 0.3587 & 0.0799 & 0.6123 & 0.9892 & 2.0401 \\
\midrule
\textit{Cosmos} & & & & & \textit{2.8713/4} \\
\midrule
GT                 & 0.5992 & 0.5060 & 0.7927 & 0.9728 & 2.8707 \\
VGGT               & 0.5990 & 0.5205 & 0.7960 & 0.9717 & 2.8872 \\
VGGT-$\Omega$      & 0.6003 & 0.5145 & 0.7951 & 0.9717 & 2.8816 \\
4DGS               & 0.5670 & 0.4798 & 0.7997 & 0.9722 & 2.8187 \\
4C4D               & 0.5909 & 0.5184 & 0.8175 & 0.9713 & 2.8981 \\
\midrule
\textit{RoboDreamer} & & & & & \textit{2.6996/4} \\
\midrule
GT                 & 0.6786 & 0.2184 & 0.8318 & 0.9760 & 2.7048 \\
VGGT               & 0.6578 & 0.2092 & 0.8267 & 0.9763 & 2.6700 \\
VGGT-$\Omega$      & 0.6575 & 0.2106 & 0.8261 & 0.9763 & 2.6705 \\
4DGS               & 0.6836 & 0.2336 & 0.8423 & 0.9753 & 2.7348 \\
4C4D               & 0.6808 & 0.2272 & 0.8338 & 0.9760 & 2.7178 \\
\bottomrule
\end{tabular}
\label{tab:comparison7}
\end{table}

\begin{table}[t]
\centering
\caption{Motion quality score comparison of different prompts.}
\small
\begin{tabular}{lccccc}
\toprule
\textbf{Method} & \textbf{Dynamic Degree} & \textbf{Flow Score} & \textbf{Motion Smoothness} & \textbf{Temporal Flickering} & \textbf{MQ} \\
\midrule
\textit{Wan} & & & & & \textit{3.1164/4} \\
\midrule
GT+Instruction      & 0.7093 & 0.5808 & 0.8421 & 0.9675 & 3.0997 \\
GT+Previous Prompt  & 0.6986 & 0.5203 & 0.8567 & 0.9686 & 3.0442 \\
GT+Current Prompt   & 0.7516 & 0.6043 & 0.8792 & 0.9702 & 3.2053 \\
\midrule
\textit{Cosmos} & & & & & \textit{2.8610/4} \\
\midrule
GT+Instruction      & 0.5930 & 0.4964 & 0.7912 & 0.9725 & 2.8531 \\
GT+Previous Prompt  & 0.5977 & 0.4976 & 0.7911 & 0.9727 & 2.8591 \\
GT+Current Prompt   & 0.5992 & 0.5060 & 0.7927 & 0.9728 & 2.8707 \\
\bottomrule
\end{tabular}
\label{tab:comparison8}
\end{table}

\paragraph{Impact of Reconstruction Quality.}
Motion quality is comparatively robust to reconstruction for the feed-forward representations, but exhibits distinct behavior for the two 4D reconstruction methods. 
Relative to the GT MQ of $2.6313$, VGGT and VGGT-$\Omega$ retain $99.5\%$ and $99.4\%$, respectively, whereas 4DGS decreases to $2.3585$ ($89.6\%$). 
In contrast, 4C4D reaches $2.7012$, exceeding GT by $2.7\%$. 
This increase should not be interpreted as superior motion fidelity: 4C4D produces substantially higher Dynamic Degree ($0.6196$ vs.\ $0.5786$) and Flow Score ($0.4206$ vs.\ $0.3042$), while exhibiting lower Motion Smoothness ($0.6796$ vs.\ $0.7651$). 
Thus, the two 4D representations exhibit qualitatively different temporal distortions: 4DGS suppresses motion magnitude and smoothness, whereas 4C4D amplifies apparent motion without preserving the same level of smoothness.
Temporal Flickering is nearly saturated across reconstruction substrates ($0.9814$--$0.9855$), making it substantially less discriminative than Dynamic Degree, Flow Score, or Motion Smoothness in this setting. 
More generally, the non-monotonic reconstruction ordering highlights an important property of MQ: it evaluates desirable motion characteristics rather than direct correspondence to the reference trajectory. 
Consequently, reconstruction artifacts can either decrease or artificially increase individual motion scores, motivating their joint interpretation with reference-based and state-level metrics.

\paragraph{Comparison of Different World Models.}
The world models exhibit substantial differences in motion characteristics. 
Wan achieves the highest mean MQ of $3.1620$, followed by Cosmos ($2.8713$), RoboDreamer ($2.6996$), RoboTwin ($2.5848$), and CogVideoX ($2.0245$). Relative to RoboTwin, these correspond to changes of approximately $+22.3\%$, $+11.1\%$, $+4.4\%$, and $-21.7\%$ for Wan, Cosmos, RoboDreamer, and CogVideoX, respectively. 
Wan's high score is supported by the highest average Dynamic Degree ($0.7416$), Flow Score ($0.5595$), and Motion Smoothness ($0.8931$), indicating that it tends to produce more pronounced and smoother motion than the other evaluated models. 
Cosmos achieves a more moderate profile, whereas RoboDreamer combines relatively high Dynamic Degree ($0.6717$) with substantially lower Flow Score ($0.2198$).
CogVideoX presents the opposite behavior. 
It obtains the lowest Dynamic Degree ($0.3423$), Flow Score ($0.0824$), and Motion Smoothness ($0.6105$), but the highest Temporal Flickering score ($0.9894$). 
This result illustrates why temporal stability should not be evaluated independently: videos containing limited motion can trivially appear temporally stable while failing to represent the dynamics of the underlying interaction. 
Accordingly, the higher MQ values of Wan, Cosmos, and RoboDreamer should be interpreted as stronger motion characteristics rather than direct evidence of more accurate physical dynamics.
Reconstruction sensitivity is also considerably compressed after generation. 
The MQ range across substrates is $0.3427$ for RoboTwin, compared with $0.2262$ for Wan and only $0.0794$, $0.0794$, and $0.0648$ for CogVideoX, Cosmos, and RoboDreamer, respectively. 
This reduced variation suggests that model-specific generative dynamics increasingly dominate the temporal characteristics of the rollout, thereby attenuating differences inherited from the reconstruction substrate.

\paragraph{Prompt Specification.}
Prompt specification affects motion quality differently across the two evaluated models. 
For Wan, the proposed current prompt achieves the highest MQ of $3.2053$, compared with $3.0997$ for the instruction-only prompt and $3.0442$ for the previous prompt. 
The improvement over the instruction baseline is $0.1056$ ($3.4\%$), with gains across Dynamic Degree ($0.7093 !\to! 0.7516$), Flow Score ($0.5808 !\to! 0.6043$), Motion Smoothness ($0.8421 !\to! 0.8792$), and Temporal Flickering ($0.9675 !\to! 0.9702$). 
However, the progression is not monotonic: the previous prompt reduces both Dynamic Degree and Flow Score and consequently lowers aggregate MQ before the current prompt produces a substantial recovery. 
This indicates that richer textual conditioning is beneficial only when the additional specification effectively constrains the intended dynamics.
Cosmos exhibits a more stable monotonic trend, increasing from $2.8531$ to $2.8591$ and $2.8707$ as prompt specificity increases. 
The total improvement is modest ($+0.0176$, or $0.6\%$), suggesting substantially lower sensitivity to prompt formulation than Wan. 
Overall, the proposed current prompt yields the strongest motion characteristics for both models, with a considerably larger effect on Wan. 
These gains nevertheless indicate improved motion magnitude, smoothness, and temporal stability rather than correctness of the predicted action or physical evolution; the latter requires complementary state-, geometry-, and task-level evaluation.

\subsubsection{Content Consistency}

\begin{table}[t]
\centering
\caption{Content consistency score comparison of different world models.}
\small
\begin{tabular}{lcccccc}
\toprule
\textbf{Method} & \textbf{Subject} & \textbf{Background} & \textbf{Style} & \textbf{Photometric} & \textbf{Semantics}  & \textbf{CC} \\
\midrule
\textit{RoboTwin} & & & & & & \textit{3.0863/5} \\
\midrule
GT                 & 0.8167 & 0.8725 & 0.9734 & 0.2226 & 0.2113 & 3.0965 \\
VGGT               & 0.8155 & 0.8558 & 0.9843 & 0.2392 & 0.2173 & 3.1121 \\
VGGT-$\Omega$      & 0.8152 & 0.8563 & 0.9827 & 0.2397 & 0.2173 & 3.1112 \\
4DGS               & 0.8199 & 0.8954 & 0.9872 & 0.1823 & 0.1508 & 3.0356 \\
4C4D               & 0.7965 & 0.8726 & 0.9862 & 0.2008 & 0.2200 & 3.0761 \\
\midrule
\textit{Wan} & & & & & & \textit{2.9396/5} \\
\midrule
GT                 & 0.7896 & 0.8749 & 0.9628 & 0.1012 & 0.2497 & 2.9782 \\
VGGT               & 0.7886 & 0.8657 & 0.9753 & 0.1094 & 0.1934 & 2.9324 \\
VGGT-$\Omega$      & 0.7907 & 0.8684 & 0.9746 & 0.1132 & 0.1910 & 2.9379 \\
4DGS               & 0.7883 & 0.8718 & 0.9647 & 0.0980 & 0.1862 & 2.9090 \\
4C4D               & 0.7922 & 0.8667 & 0.9602 & 0.1142 & 0.2070 & 2.9403 \\
\midrule
\textit{CogVideoX} & & & & & & \textit{3.0361/5} \\
\midrule
GT                 & 0.8935 & 0.8013 & 0.9763 & 0.2365 & 0.1370 & 3.0446 \\
VGGT               & 0.8914 & 0.7675 & 0.9805 & 0.2302 & 0.1280 & 2.9976 \\
VGGT-$\Omega$      & 0.8869 & 0.7439 & 0.9792 & 0.2290 & 0.1263 & 2.9653 \\
4DGS               & 0.8707 & 0.8305 & 0.9725 & 0.2879 & 0.1340 & 3.0956 \\
4C4D               & 0.8975 & 0.8067 & 0.9788 & 0.2534 & 0.1412 & 3.0776 \\
\midrule
\textit{Cosmos} & & & & & & \textit{2.9250/5} \\
\midrule
GT                 & 0.7965 & 0.8378 & 0.9723 & 0.0947 & 0.2661 & 2.9674 \\
VGGT               & 0.7913 & 0.8303 & 0.9782 & 0.0939 & 0.2237 & 2.9174 \\
VGGT-$\Omega$      & 0.7906 & 0.8309 & 0.9735 & 0.0937 & 0.2234 & 2.9121 \\
4DGS               & 0.7868 & 0.8452 & 0.9729 & 0.1031 & 0.2101 & 2.9181 \\
4C4D               & 0.7902 & 0.8326 & 0.9651 & 0.0966 & 0.2256 & 2.9101 \\
\midrule
\textit{RoboDreamer} & & & & & & \textit{2.8726/5} \\
\midrule
GT                 & 0.7569 & 0.8713 & 0.9909 & 0.0959 & 0.1643 & 2.8793 \\
VGGT               & 0.7451 & 0.8652 & 0.9930 & 0.0997 & 0.1664 & 2.8694 \\
VGGT-$\Omega$      & 0.7535 & 0.8656 & 0.9933 & 0.0998 & 0.1656 & 2.8778 \\
4DGS               & 0.7394 & 0.8764 & 0.9932 & 0.0921 & 0.1584 & 2.8595 \\
4C4D               & 0.7530 & 0.8711 & 0.9876 & 0.0977 & 0.1677 & 2.8771 \\
\bottomrule
\end{tabular}
\label{tab:comparison9}
\end{table}

\begin{table}[t]
\centering
\caption{Content consistency score comparison of different prompts.}
\small
\begin{tabular}{lcccccc}
\toprule
\textbf{Method} & \textbf{Subject} & \textbf{Background} & \textbf{Style} & \textbf{Photometric} & \textbf{Semantics}  & \textbf{CC} \\
\midrule
\textit{Wan} & & & & & & \textit{2.9504/5} \\
\midrule
GT+Instruction      & 0.8124 & 0.8880 & 0.9743 & 0.0912 & 0.1762 & 2.9421 \\
GT+Previous Prompt  & 0.7959 & 0.8816 & 0.9698 & 0.0993 & 0.1843 & 2.9309 \\
GT+Current Prompt   & 0.7896 & 0.8749 & 0.9628 & 0.1012 & 0.2497 & 2.9782 \\
\midrule
\textit{Cosmos} & & & & & & \textit{2.9173/5} \\
\midrule
GT+Instruction      & 0.7937 & 0.8331 & 0.9724 & 0.0940 & 0.1713 & 2.8645 \\
GT+Previous Prompt  & 0.7987 & 0.8361 & 0.9726 & 0.0932 & 0.2193 & 2.9199 \\
GT+Current Prompt   & 0.7965 & 0.8378 & 0.9723 & 0.0947 & 0.2661 & 2.9674 \\
\bottomrule
\end{tabular}
\label{tab:comparison10}
\end{table}

\paragraph{Impact of Reconstruction Quality.}
Content consistency is substantially more robust to reconstruction than the preceding appearance-oriented dimensions. 
On RoboTwin, GT achieves a CC of $3.0965$, while VGGT and VGGT-$\Omega$ obtain slightly higher scores of $3.1121$ and $3.1112$, respectively. 
The two 4D representations remain close to the reference, with 4C4D retaining $99.3\%$ of the GT score ($3.0761$) and 4DGS retaining $98.0\%$ ($3.0356$). 
Thus, unlike Image Quality or Distribution Similarity, CC does not decrease monotonically with reconstruction fidelity. 
This behavior is expected because the constituent metrics primarily measure consistency across the video rather than exact correspondence to the reference; moderate reconstruction smoothing can therefore preserve, or even slightly increase, some consistency scores without improving reconstruction accuracy.
The component-wise results further reveal that the aggregate robustness masks different sensitivities. 
For 4DGS, Subject, Background, and Style remain comparable to or slightly above GT, whereas Photometric and Semantics decrease from $0.2226$ to $0.1823$ and from $0.2113$ to $0.1508$, respectively. 
The resulting CC reduction is therefore concentrated in these more discriminative components rather than reflecting a uniform deterioration of content. 
Similarly, 4C4D exhibits reduced Subject and Photometric consistency while maintaining strong Background, Style, and Semantics scores. 
These results indicate that CC complements fidelity-oriented metrics by characterizing whether visual content remains internally coherent, while not serving as a direct proxy for reconstruction accuracy.

\paragraph{Comparison of Different World Models.}
The differences among world models are relatively modest at the aggregate level but substantially more informative at the component level. 
RoboTwin achieves the highest mean CC of $3.0863$, followed closely by CogVideoX at $3.0361$, which retains $98.4\%$ of the reference score. 
Wan, Cosmos, and RoboDreamer achieve $2.9396$, $2.9250$, and $2.8726$, corresponding to $95.2\%$, $94.8\%$, and $93.1\%$ of RoboTwin, respectively. 
Compared with the much larger model gaps observed for Image Quality and Distribution Similarity, these results suggest that maintaining internally consistent subjects, backgrounds, and styles is considerably easier for current world models than reproducing the precise visual distribution of the ground-truth future.
However, similar aggregate CC values can arise from markedly different consistency profiles. 
CogVideoX achieves the highest average Subject ($0.8880$) and Photometric ($0.2474$) consistency among the evaluated models, but substantially lower Background ($0.7900$) and Semantics ($0.1333$) scores. 
RoboDreamer exhibits the highest Style consistency ($0.9916$) yet the lowest overall CC, primarily because of weaker Subject and Semantics consistency. 
In contrast, Cosmos obtains the highest mean Semantics score among the generated models ($0.2298$), despite its lower aggregate CC. 
These component-level differences demonstrate that a single consistency score cannot fully characterize model behavior and motivate the multi-metric design of this sub-dimension.
The influence of reconstruction on generated content is likewise model dependent. 
The CC range across substrates is $0.0765$ for RoboTwin and remains comparable for Wan ($0.0692$), decreases for Cosmos ($0.0573$) and RoboDreamer ($0.0198$), but increases to $0.1303$ for CogVideoX. 
Moreover, the substrate ranking is not consistently preserved after generation; for example, CogVideoX obtains its highest CC with 4DGS rather than GT. 
This non-monotonicity further indicates that content-consistency metrics capture properties of the generated sequence itself and can be strongly influenced by the model's generative prior, rather than solely by the fidelity of its conditioning substrate.

\paragraph{Prompt Specification.}
Prompt specification primarily influences semantic consistency, while its effect on low-level visual consistency is substantially weaker. 
For Wan, CC exhibits a non-monotonic progression from $2.9421$ with the instruction-only prompt to $2.9309$ with the previous prompt and $2.9782$ with the proposed current prompt. 
Although Subject, Background, and Style decrease from $0.8124$, $0.8880$, and $0.9743$ to $0.7896$, $0.8749$, and $0.9628$, respectively, Semantics increases substantially from $0.1762$ to $0.2497$ ($+41.7\%$), together with a smaller improvement in Photometric consistency. 
Consequently, the current prompt achieves the highest aggregate CC despite modest reductions in several appearance-oriented components.
The effect is even clearer for Cosmos, whose CC increases monotonically from $2.8645$ to $2.9199$ and $2.9674$, corresponding to a $3.6\%$ improvement over the instruction baseline. 
This gain is dominated by Semantics, which rises from $0.1713$ to $0.2193$ and ultimately $0.2661$, while Subject, Background, Style, and Photometric consistency remain nearly unchanged. 
Taken together, the two models indicate that richer prompt specification primarily improves preservation of semantically relevant content rather than simply enforcing greater visual invariance. 
This distinction is desirable for embodied world modeling, where successful prediction may require substantial visual change as an interaction unfolds; accordingly, high content consistency should be interpreted jointly with state- and task-level metrics rather than as evidence that the scene should remain visually static.

\subsection{3D Geometry Consistency }

\subsubsection{Reconstruction-Based Geometry}

\begin{table}[t]
\centering
\caption{Reconstruction-based geometry score comparison of different world models.}
\small
\begin{tabular}{lccccc}
\toprule
\textbf{Method} & \textbf{Fidelity} & \textbf{Stability} & \textbf{Reprojection Geometric} & \textbf{Reprojection Photometric} & \textbf{RBG} \\
\midrule
\textit{RoboTwin} & & & & & \textit{1.8689/4} \\
\midrule
GT                 & 0.5386 & 0.3960 & 0.9339 & 0.1974 & 2.0659 \\
VGGT               & 0.2993 & 0.4873 & 0.9059 & 0.2017 & 1.8942 \\
VGGT-$\Omega$      & 0.3043 & 0.4881 & 0.9081 & 0.2010 & 1.9015 \\
4DGS               & 0.0891 & 0.4912 & 0.9550 & 0.2244 & 1.7597 \\
4C4D               & 0.0969 & 0.4714 & 0.9676 & 0.1875 & 1.7234 \\
\midrule
\textit{Wan} & & & & & \textit{1.8531/4} \\
\midrule
GT                 & 0.6339 & 0.2968 & 0.9576 & 0.1517 & 2.0400 \\
VGGT               & 0.3990 & 0.4071 & 0.9276 & 0.1543 & 1.8880 \\
VGGT-$\Omega$      & 0.4340 & 0.4010 & 0.9325 & 0.1580 & 1.9255 \\
4DGS               & 0.1964 & 0.4451 & 0.9372 & 0.1352 & 1.7139 \\
4C4D               & 0.1971 & 0.4356 & 0.9551 & 0.1105 & 1.6983 \\
\midrule
\textit{CogVideoX} & & & & & \textit{2.4253/4} \\
\midrule
GT                 & 0.5443 & 0.4492 & 0.9685 & 0.4864 & 2.4484 \\
VGGT               & 0.4480 & 0.4520 & 0.9615 & 0.5512 & 2.4127 \\
VGGT-$\Omega$      & 0.4788 & 0.4830 & 0.9627 & 0.5566 & 2.4811 \\
4DGS               & 0.3730 & 0.5155 & 0.9649 & 0.5684 & 2.4218 \\
4C4D               & 0.2893 & 0.5256 & 0.9720 & 0.5757 & 2.3626 \\
\midrule
\textit{Cosmos} & & & & & \textit{1.8549/4} \\
\midrule
GT                 & 0.5052 & 0.3309 & 0.9234 & 0.1976 & 1.9571 \\
VGGT               & 0.2851 & 0.4316 & 0.9079 & 0.1958 & 1.8204 \\
VGGT-$\Omega$      & 0.3093 & 0.4469 & 0.9095 & 0.1956 & 1.8613 \\
4DGS               & 0.2429 & 0.4279 & 0.9181 & 0.2036 & 1.7925 \\
4C4D               & 0.3080 & 0.4263 & 0.9238 & 0.1851 & 1.8432 \\
\midrule
\textit{RoboDreamer} & & & & & \textit{1.7854/4} \\
\midrule
GT                 & 0.2031 & 0.4126 & 0.9635 & 0.2261 & 1.8053 \\
VGGT               & 0.1382 & 0.4758 & 0.9572 & 0.2326 & 1.8038 \\
VGGT-$\Omega$      & 0.1460 & 0.4788 & 0.9584 & 0.2339 & 1.8171 \\
4DGS               & 0.0868 & 0.4623 & 0.9559 & 0.2161 & 1.7211 \\
4C4D               & 0.0999 & 0.4939 & 0.9637 & 0.2222 & 1.7797 \\
\bottomrule
\end{tabular}
\label{tab:comparison11}
\end{table}

\begin{table}[t]
\centering
\caption{Reconstruction-based geometry score comparison of different prompts.}
\small
\begin{tabular}{lccccc}
\toprule
\textbf{Method} & \textbf{Fidelity} & \textbf{Stability} & \textbf{Reprojection Geometric} & \textbf{Reprojection Photometric} & \textbf{RBG} \\
\midrule
\textit{Wan} & & & & & \textit{2.0835/4} \\
\midrule
GT+Instruction      & 0.5378 & 0.3883 & 0.9722 & 0.2288 & 2.1271 \\
GT+Previous Prompt  & 0.6071 & 0.3305 & 0.9646 & 0.1811 & 2.0833 \\
GT+Current Prompt   & 0.6339 & 0.2968 & 0.9576 & 0.1517 & 2.0400 \\
\midrule
\textit{Cosmos} & & & & & \textit{1.9700/4} \\
\midrule
GT+Instruction      & 0.4980 & 0.3535 & 0.9232 & 0.2034 & 1.9781 \\
GT+Previous Prompt  & 0.5038 & 0.3461 & 0.9223 & 0.2027 & 1.9749 \\
GT+Current Prompt   & 0.5052 & 0.3309 & 0.9234 & 0.1976 & 1.9571 \\
\bottomrule
\end{tabular}
\label{tab:comparison12}
\end{table}

\paragraph{Impact of Reconstruction Quality.}
Reconstruction quality has a clear but component-dependent effect on reconstruction-based geometry. 
On RoboTwin, the aggregate RBG follows $\text{GT} > \text{VGGT-}\Omega > \text{VGGT} > \text{4DGS} > \text{4C4D}$, with scores of $2.0659$, $1.9015$, $1.8942$, $1.7597$, and $1.7234$, respectively. 
Relative to GT, VGGT-$\Omega$ and VGGT retain $92.0\%$ and $91.7\%$ of RBG, whereas 4DGS and 4C4D retain $85.2\%$ and $83.4\%$. 
The degradation is dominated by the Fidelity component: 4DGS and 4C4D retain only $16.5\%$ and $18.0\%$ of the GT Fidelity score. 
In contrast, both 4D representations obtain higher Stability and Reprojection Geometric scores than GT. 
This opposing behavior demonstrates that the individual components measure complementary properties: a reconstruction can be internally stable and geometrically self-consistent under reprojection while still recovering substantially less of the underlying scene structure.
This compensation also reduces the apparent variation in the aggregate. 
Across RoboTwin substrates, the component-wise ranges sum to $0.6433$, whereas RBG itself spans only $0.3425$. 
Therefore, RBG should be interpreted jointly with its component profile rather than as a single measure of geometric fidelity. 
In particular, the relatively strong Stability and Reprojection scores of the 4D representations should not be taken as evidence that they reconstruct more accurate geometry than the feed-forward methods, given their pronounced reduction in Fidelity.

\paragraph{Comparison of Different World Models.}
The world-model comparison reveals substantial differences in reconstruction-based geometric characteristics. 
CogVideoX obtains the highest mean RBG at $2.4253$, exceeding the RoboTwin reference average ($1.8689$) by $29.8\%$. 
Wan and Cosmos achieve similar mean scores of $1.8531$ and $1.8549$, corresponding to $99.2\%$ and $99.3\%$ of RoboTwin, respectively, while RoboDreamer reaches $1.7854$ ($95.5\%$). 
However, the unusually high RBG of CogVideoX should not be interpreted directly as superior physical geometry. 
Its advantage is driven in large part by Reprojection Photometric consistency, whose mean score reaches $0.5477$, substantially above RoboTwin ($0.2024$), Wan ($0.1419$), Cosmos ($0.1955$), and RoboDreamer ($0.2262$). 
This result highlights that reconstruction-based self-consistency and geometric correctness are related but distinct properties.
The component profiles further differentiate the models. 
Wan achieves relatively high Fidelity but reduced Stability and Reprojection Photometric consistency, whereas RoboDreamer exhibits the lowest mean Fidelity ($0.1348$) despite maintaining strong Reprojection Geometric consistency ($0.9597$). 
Cosmos occupies a more balanced regime, with moderate Fidelity and relatively stable reprojection behavior. 
Reconstruction sensitivity also varies across models: the RBG range across substrates is $0.3425$ for RoboTwin and $0.3417$ for Wan, but contracts to $0.1646$ for Cosmos, $0.1185$ for CogVideoX, and $0.0960$ for RoboDreamer. 
The smaller ranges indicate that, for these models, the generated geometry is increasingly determined by model-specific generation behavior rather than by differences among reconstruction substrates.

\paragraph{Prompt Specification.}
Prompt specification induces a clear trade-off among the RBG components. 
For Wan, aggregate RBG decreases monotonically from $2.1271$ with the instruction-only prompt to $2.0833$ with the previous prompt and $2.0400$ with the proposed current prompt, corresponding to a $4.1\%$ reduction. 
Importantly, this decrease does not reflect uniform geometric degradation. 
Fidelity increases substantially from $0.5378$ to $0.6339$ ($+17.9\%$), while Stability decreases from $0.3883$ to $0.2968$, Reprojection Geometric from $0.9722$ to $0.9576$, and Reprojection Photometric from $0.2288$ to $0.1517$. 
Thus, richer prompt conditioning produces a geometry representation with higher Fidelity but lower reconstruction stability and reprojection consistency.
Cosmos exhibits the same aggregate tendency but with substantially weaker sensitivity. RBG decreases only from $1.9781$ to $1.9571$ ($-1.1\%$), while Fidelity increases slightly from $0.4980$ to $0.5052$. 
Stability decreases by $6.4\%$, Reprojection Photometric by $2.9\%$, and Reprojection Geometric remains essentially unchanged. 
These results indicate that Cosmos is comparatively robust to prompt formulation at the reconstruction-based geometry level, whereas Wan exhibits a stronger prompt-dependent trade-off between geometric richness and internal reconstruction consistency. 
Consequently, the lower RBG obtained with the current prompt should not be interpreted in isolation as poorer geometry; rather, the component-wise results show that prompt specification changes the balance between recovered geometric content and reconstruction self-consistency. 
Direct geometric correctness is therefore further assessed using the complementary geometry metrics in the subsequent sub-dimensions.

\subsubsection{Depth Consistency}

\begin{table}[t]
\centering
\caption{Depth consistency score comparison of different world models.}
\small
\begin{tabular}{lccc}
\toprule
\textbf{Method} & \textbf{Depth Accuracy} & \textbf{Perspectivity} & \textbf{DC} \\
\midrule
\textit{RoboTwin} & & & \textit{1.7086/2} \\
\midrule
GT                 & 1.0000 & 0.8965 & 1.8965 \\
VGGT               & 1.0000 & 0.8560 & 1.8560 \\
VGGT-$\Omega$      & 1.0000 & 0.8725 & 1.8725 \\
4DGS               & 0.7982 & 0.6180 & 1.4162 \\
4C4D               & 0.7256 & 0.7760 & 1.5016 \\
\midrule
\textit{Wan} & & & \textit{1.6329/2} \\
\midrule
GT                 & 0.7116 & 0.9780 & 1.6896 \\
VGGT               & 0.6417 & 0.9300 & 1.5717 \\
VGGT-$\Omega$      & 0.6414 & 0.9500 & 1.5914 \\
4DGS               & 0.8686 & 0.8090 & 1.6776 \\
4C4D               & 0.7809 & 0.8535 & 1.6344 \\
\midrule
\textit{CogVideoX} & & & \textit{1.6029/2} \\
\midrule
GT                 & 0.7365 & 0.8305 & 1.5670 \\
VGGT               & 0.6996 & 0.8345 & 1.5341 \\
VGGT-$\Omega$      & 0.6801 & 0.8255 & 1.5056 \\
4DGS               & 0.9300 & 0.7950 & 1.7250 \\
4C4D               & 0.8700 & 0.8130 & 1.6830 \\
\midrule
\textit{Cosmos} & & & \textit{1.7835/2} \\
\midrule
GT                 & 0.9147 & 0.9560 & 1.8707 \\
VGGT               & 0.8780 & 0.8885 & 1.7665 \\
VGGT-$\Omega$      & 0.8713 & 0.9080 & 1.7793 \\
4DGS               & 0.8981 & 0.8225 & 1.7206 \\
4C4D               & 0.8790 & 0.9015 & 1.7805 \\
\midrule
\textit{RoboDreamer} & & & \textit{1.4457/2} \\
\midrule
GT                 & 0.7213 & 0.7455 & 1.4668 \\
VGGT               & 0.6987 & 0.7440 & 1.4427 \\
VGGT-$\Omega$      & 0.7003 & 0.7440 & 1.4443 \\
4DGS               & 0.7112 & 0.7365 & 1.4477 \\
4C4D               & 0.6941 & 0.7330 & 1.4271 \\
\bottomrule
\end{tabular}
\label{tab:comparison13}
\end{table}

\begin{table}[t]
\centering
\caption{Depth consistency score comparison of different prompts.}
\small
\begin{tabular}{lccc}
\toprule
\textbf{Method} & \textbf{Depth Accuracy} & \textbf{Perspectivity} & \textbf{DC} \\
\midrule
\textit{Wan} & & & \textit{1.5715/2} \\
\midrule
GT+Instruction      & 0.6615 & 0.8005 & 1.4620 \\
GT+Previous Prompt  & 0.6909 & 0.8720 & 1.5629 \\
GT+Current Prompt   & 0.7116 & 0.9780 & 1.6896 \\
\midrule
\textit{Cosmos} & & & \textit{1.7879/2} \\
\midrule
GT+Instruction      & 0.9081 & 0.8285 & 1.7366 \\
GT+Previous Prompt  & 0.9114 & 0.8450 & 1.7564 \\
GT+Current Prompt   & 0.9147 & 0.9560 & 1.8707 \\
\bottomrule
\end{tabular}
\label{tab:comparison14}
\end{table}

\paragraph{Impact of Reconstruction Quality.}
Depth consistency is highly sensitive to reconstruction quality. 
On RoboTwin, GT achieves a DC of $1.8965$, while VGGT-$\Omega$ and VGGT retain $98.7\%$ and $97.9\%$ of the GT score, respectively. 
In contrast, 4C4D and 4DGS retain only $79.2\%$ and $74.7\%$, revealing a substantially larger degradation for the 4D reconstruction methods. 
The overall substrate range reaches $0.4803$, indicating that depth-related evaluation is more sensitive to reconstruction choice than reconstruction-based geometry, whose aggregate range is smaller.
The two components expose complementary failure modes. 
Depth Accuracy remains perfect for GT, VGGT, and VGGT-$\Omega$, but decreases to $0.7982$ for 4DGS and $0.7256$ for 4C4D, suggesting considerable distortion of the recovered depth structure. 
Perspectivity exhibits a different ordering: 4DGS obtains the lowest score ($0.6180$), whereas 4C4D reaches $0.7760$. 
Thus, 4C4D is more strongly penalized by reference-based depth accuracy, while 4DGS is more strongly penalized by the perspectivity measure. 
This disagreement illustrates why both components are necessary: numerical depth agreement and perceptual 3D plausibility capture related but distinct aspects of geometric consistency.

\paragraph{Comparison of Different World Models.}
Among the generated models, Cosmos achieves the strongest mean DC at $1.7835$, exceeding the RoboTwin average ($1.7086$) by $4.4\%$. 
Wan and CogVideoX follow with $1.6329$ and $1.6029$, corresponding to $95.6\%$ and $93.8\%$ of RoboTwin, while RoboDreamer obtains the lowest score at $1.4457$ ($84.6\%$). 
The high Cosmos score results from a comparatively balanced profile, with mean Depth Accuracy of $0.8882$ and Perspectivity of $0.8953$. 
In contrast, Wan achieves relatively strong perspectivity ($0.9041$ on average) but substantially lower Depth Accuracy ($0.7288$), demonstrating that geometrically plausible-looking rollouts do not necessarily reproduce the reference depth accurately.
The interaction between reconstruction and generation is strongly model dependent. 
Cosmos largely preserves the expected advantage of stronger substrates, achieving its highest DC with GT ($1.8707$), while RoboDreamer is comparatively insensitive to reconstruction, with only a $0.0397$ range across all substrates. 
Wan and CogVideoX, however, exhibit pronounced non-monotonicity. 
Wan obtains nearly comparable scores with GT ($1.6896$) and 4DGS ($1.6776$), whereas CogVideoX performs better on both 4DGS ($1.7250$) and 4C4D ($1.6830$) than GT ($1.5670$), with its maximum attained on 4DGS. 
These inversions are driven primarily by Depth Accuracy, which increases to $0.8686$ for Wan and $0.9300$ for CogVideoX under 4DGS conditioning. 
Therefore, a higher DC obtained from a degraded reconstruction substrate should not be interpreted as evidence that poorer reconstruction improves world modeling. 
Rather, it indicates that the generative prior can substantially reshape the predicted geometry and, in some cases, produce depth estimates that align more closely with the evaluation criterion than the conditioning reconstruction itself.
The magnitude of substrate dependence further supports this interpretation. 
The DC range contracts from $0.4803$ for RoboTwin to $0.1179$ for Wan and $0.1501$ for Cosmos, and to only $0.0397$ for RoboDreamer, although CogVideoX retains a larger range of $0.2194$. 
Overall, these results show that reconstruction fidelity and generated depth quality are not monotonically coupled; once generation begins, model-specific geometric priors can become a dominant factor.

\paragraph{Prompt Specification.}
Prompt specification produces a strong and consistent improvement in depth consistency for both Wan and Cosmos. 
For Wan, DC increases monotonically from $1.4620$ with the instruction-only prompt to $1.5629$ with the previous prompt and $1.6896$ with the proposed current prompt. 
The total gain of $0.2276$ corresponds to a $15.6\%$ relative improvement. 
Both components contribute positively: Depth Accuracy increases from $0.6615$ to $0.7116$, while Perspectivity rises more substantially from $0.8005$ to $0.9780$. 
Thus, richer prompt specification improves both reference-based depth agreement and perceived perspective consistency.
Cosmos exhibits the same monotonic trend, increasing from $1.7366$ to $1.7564$ and $1.8707$, corresponding to a $7.7\%$ improvement over the instruction baseline. 
Depth Accuracy changes only modestly ($0.9081 !\to! 0.9147$), whereas Perspectivity increases markedly from $0.8285$ to $0.9560$. 
Notably, most of the Cosmos improvement occurs between the previous and current prompts, indicating that the proposed prompt formulation provides additional geometric cues beyond those captured by a generic detailed description. 
Across both models, the dominant gain arises from Perspectivity rather than Depth Accuracy. 
Accordingly, the prompt results support improved geometric plausibility and, to a lesser extent, closer depth agreement with the reference, but should not alone be interpreted as evidence of fully correct 3D dynamics.

\subsubsection{Camera Trajectory Geometry}

\begin{table}[t]
\centering
\caption{Camera trajectory geometry score comparison of different world models.}
\small
\begin{tabular}{lc}
\toprule
\textbf{Method} & \textbf{Camera Control (CTG)} \\
\midrule
\textit{RoboTwin} & \textit{0.8687/1} \\
\midrule
GT                 & 0.8724 \\
VGGT               & 0.9050 \\
VGGT-$\Omega$      & 0.9021 \\
4DGS               & 0.9476 \\
4C4D               & 0.7166 \\
\midrule
\textit{Wan} & \textit{0.9054/1} \\
\midrule
GT                 & 0.8871 \\
VGGT               & 0.9023 \\
VGGT-$\Omega$      & 0.9247 \\
4DGS               & 0.9458 \\
4C4D               & 0.8669 \\
\midrule
\textit{CogVideoX} & \textit{0.9377/1} \\
\midrule
GT                 & 0.9464 \\
VGGT               & 0.9165 \\
VGGT-$\Omega$      & 0.9309 \\
4DGS               & 0.9451 \\
4C4D               & 0.9496 \\
\midrule
\textit{Cosmos} & \textit{0.9063/1} \\
\midrule
GT                 & 0.9112 \\
VGGT               & 0.8927 \\
VGGT-$\Omega$      & 0.9112 \\
4DGS               & 0.9250 \\
4C4D               & 0.8914 \\
\midrule
\textit{RoboDreamer} & \textit{0.9807/1} \\
\midrule
GT                 & 0.9771 \\
VGGT               & 0.9836 \\
VGGT-$\Omega$      & 0.9838 \\
4DGS               & 0.9830 \\
4C4D               & 0.9760 \\
\bottomrule
\end{tabular}
\label{tab:comparison15}
\end{table}

\begin{table}[t]
\centering
\caption{Camera trajectory geometry score comparison of different prompts.}
\small
\begin{tabular}{lc}
\toprule
\textbf{Method} & \textbf{Camera Control (CTG)} \\
\midrule
\textit{Wan} & \textit{0.8820/1} \\
\midrule
GT+Instruction   & 0.8759 \\
GT+Previous Prompt      & 0.8831 \\
GT+Current Prompt               & 0.8871 \\
\midrule
\textit{Cosmos} & \textit{0.9053/1} \\
\midrule
GT+Instruction   & 0.8946 \\
GT+Previous Prompt      & 0.9102 \\
GT+Current Prompt               & 0.9112 \\
\bottomrule
\end{tabular}
\label{tab:comparison16}
\end{table}

\paragraph{Impact of Reconstruction Quality.}
Camera trajectory geometry exhibits a markedly non-monotonic response to reconstruction quality. 
On RoboTwin, 4DGS achieves the highest CTG score of $0.9476$, followed by VGGT ($0.9050$), VGGT-$\Omega$ ($0.9021$), GT ($0.8724$), and 4C4D ($0.7166$). 
Thus, unlike Depth Consistency and Reconstruction-Based Geometry, CTG does not systematically favor higher-fidelity reconstruction substrates. 
The most prominent result is the large separation between the two 4D representations: 4DGS exceeds GT by $8.6\%$, whereas 4C4D retains only $82.1\%$ of the GT score. 
The total substrate range is $0.2310$, but decreases to only $0.0752$ when 4C4D is excluded, indicating that approximately $67\%$ of the observed variation is attributable to the particularly low 4C4D result.
This behavior suggests that CTG is especially sensitive to temporal properties that affect camera-trajectory estimation or consistency, rather than to reconstruction fidelity alone. 
In particular, a reconstruction can exhibit relatively weak appearance or depth fidelity while still providing temporally coherent cues for camera motion estimation. 
Conversely, temporal instability can substantially degrade CTG despite comparatively strong frame-level reconstruction quality. 
CTG therefore provides a complementary geometric signal, but should not be interpreted independently as a direct measure of overall 3D reconstruction accuracy.

\paragraph{Comparison of Different World Models.}
All evaluated world models achieve higher mean CTG than the RoboTwin reference average ($0.8687$). 
RoboDreamer obtains the highest score at $0.9807$, followed by CogVideoX ($0.9377$), Cosmos ($0.9063$), and Wan ($0.9054$), corresponding to improvements of approximately $12.9\%$, $7.9\%$, $4.3\%$, and $4.2\%$ over RoboTwin, respectively. 
The consistently high scores indicate that generated rollouts generally exhibit camera-motion patterns that are highly recoverable or internally consistent under this metric. 
However, this should not be interpreted as evidence that the generated camera trajectories are necessarily more accurate than those of the reference data. 
In particular, RoboDreamer achieves the strongest CTG despite substantially weaker performance in several other geometric and distributional dimensions, demonstrating that camera-trajectory consistency captures a narrower property than overall world-model fidelity.
Generation also substantially reduces sensitivity to the reconstruction substrate. 
The CTG range decreases from $0.2310$ for RoboTwin to $0.0789$ for Wan, $0.0331$ for CogVideoX, $0.0336$ for Cosmos, and only $0.0078$ for RoboDreamer. 
The strongest example occurs for 4C4D: its RoboTwin CTG is only $0.7166$, whereas the corresponding generated scores increase to $0.8669$, $0.9496$, $0.8914$, and $0.9760$ for Wan, CogVideoX, Cosmos, and RoboDreamer, respectively. 
This pronounced contraction suggests that model-specific temporal priors can regularize camera-motion characteristics and substantially attenuate trajectory irregularities inherited from degraded reconstruction substrates. 
Consequently, CTG is informative about temporal camera consistency, but its ranking should be interpreted jointly with the other geometry metrics rather than as an isolated measure of conditioning quality.

\paragraph{Prompt Specification.}
Prompt specification produces a consistent but relatively modest improvement in camera trajectory geometry. 
For Wan, CTG increases monotonically from $0.8759$ with the instruction-only prompt to $0.8831$ with the previous prompt and $0.8871$ with the proposed current prompt. 
The total increase of $0.0112$ corresponds to a $1.3\%$ relative improvement, with approximately $64\%$ of the gain occurring between the instruction and previous prompt. 
Cosmos exhibits the same ordering, increasing from $0.8946$ to $0.9102$ and $0.9112$, for an overall improvement of $0.0166$ ($1.9\%$). 
In this case, approximately $94\%$ of the gain is obtained when replacing the instruction-only prompt with the more informative previous prompt, while the current prompt provides only a marginal additional improvement.
These results indicate that richer language conditioning can improve camera-trajectory consistency for both models, although the effect is substantially smaller than that observed for several other geometric properties such as Depth Consistency. 
The proposed current prompt nevertheless achieves the highest CTG for both Wan and Cosmos, supporting the general benefit of more informative conditioning. 
Given the limited magnitude of the improvement and the non-monotonic relationship between CTG and reconstruction fidelity, however, these results should be interpreted as evidence of improved camera-motion consistency rather than direct evidence of more accurate 3D geometry.

\subsubsection{Roundtrip Consistency}

\begin{table}[t]
\centering
\caption{Roundtrip consistency score comparison of different world models.}
\small
\begin{tabular}{lc}
\toprule
\textbf{Method} & \textbf{Spatial Consistency (RC)} \\
\midrule
\textit{RoboTwin} & \textit{0.7242/1} \\
\midrule
GT                 & 0.7160 \\
VGGT               & 0.7130 \\
VGGT-$\Omega$      & 0.7121 \\
4DGS               & 0.7455 \\
4C4D               & 0.7344 \\
\midrule
\textit{Wan} & \textit{0.7084/1} \\
\midrule
GT                 & 0.7108 \\
VGGT               & 0.7036 \\
VGGT-$\Omega$      & 0.7056 \\
4DGS               & 0.7138 \\
4C4D               & 0.7084 \\
\midrule
\textit{CogVideoX} & \textit{0.7464/1} \\
\midrule
GT                 & 0.7619 \\
VGGT               & 0.7448 \\
VGGT-$\Omega$      & 0.7482 \\
4DGS               & 0.7298 \\
4C4D               & 0.7474 \\
\midrule
\textit{Cosmos} & \textit{0.7163/1} \\
\midrule
GT                 & 0.7132 \\
VGGT               & 0.7095 \\
VGGT-$\Omega$      & 0.7089 \\
4DGS               & 0.7286 \\
4C4D               & 0.7214 \\
\midrule
\textit{RoboDreamer} & \textit{0.7760/1} \\
\midrule
GT                 & 0.7841 \\
VGGT               & 0.7663 \\
VGGT-$\Omega$      & 0.7769 \\
4DGS               & 0.7784 \\
4C4D               & 0.7745 \\
\bottomrule
\end{tabular}
\label{tab:comparison17}
\end{table}

\begin{table}[t]
\centering
\caption{Roundtrip consistency score comparison of different prompts.}
\small
\begin{tabular}{lc}
\toprule
\textbf{Method} & \textbf{Spatial Consistency (RC)} \\
\midrule
\textit{Wan} & \textit{0.7277/1} \\
\midrule
GT+Instruction   & 0.7468 \\
GT+Previous Prompt      & 0.7256 \\
GT+Current Prompt               & 0.7108 \\
\midrule
\textit{Cosmos} & \textit{0.7136/1} \\
\midrule
GT+Instruction   & 0.7117 \\
GT+Previous Prompt      & 0.7159 \\
GT+Current Prompt               & 0.7132 \\
\bottomrule
\end{tabular}
\label{tab:comparison18}
\end{table}

\paragraph{Impact of Reconstruction Quality.}
Roundtrip consistency is relatively insensitive to reconstruction quality and, importantly, does not follow the fidelity-based substrate ordering observed for Depth Consistency or Reconstruction-Based Geometry. 
On RoboTwin, 4DGS achieves the highest RC of $0.7455$, followed by 4C4D ($0.7344$), GT ($0.7160$), VGGT ($0.7130$), and VGGT-$\Omega$ ($0.7121$). 
Thus, VGGT and VGGT-$\Omega$ retain $99.6\%$ and $99.5\%$ of the GT score, whereas 4DGS and 4C4D exceed GT by $4.1\%$ and $2.6\%$, respectively. 
Moreover, the overall substrate range is only $0.0334$, substantially smaller than those observed for Depth Consistency and Reconstruction-Based Geometry. 
These results indicate that RC primarily measures internal spatial consistency under the roundtrip transformation rather than absolute geometric fidelity. 
A representation may therefore achieve strong roundtrip consistency despite exhibiting substantial reconstruction or depth errors, making RC complementary to, rather than interchangeable with, reference-based geometry metrics.

\paragraph{Comparison of Different World Models.}
The world-model comparison further demonstrates that high RC does not necessarily correspond to stronger overall world-model fidelity. 
RoboDreamer achieves the highest mean RC at $0.7760$, followed by CogVideoX ($0.7464$), RoboTwin ($0.7242$), Cosmos ($0.7163$), and Wan ($0.7084$). 
Relative to RoboTwin, RoboDreamer and CogVideoX increase RC by $7.2\%$ and $3.1\%$, respectively, whereas Cosmos and Wan are lower by only $1.1\%$ and $2.2\%$. 
Given the substantially different rankings observed in Depth Consistency, Distribution Similarity, and other reference-sensitive dimensions, these results suggest that RC captures a narrower notion of spatial self-consistency: a rollout can remain internally coherent under a roundtrip transformation without necessarily reproducing the correct scene geometry or dynamics.
The dependence on reconstruction substrate is also generally weak after generation. 
The RC range contracts from $0.0334$ for RoboTwin to $0.0102$ for Wan, $0.0197$ for Cosmos, and $0.0178$ for RoboDreamer, although CogVideoX retains a comparable range of $0.0321$. 
For Wan specifically, generation reduces RC relative to the corresponding RoboTwin substrate in all five cases, with the largest reductions occurring for 4DGS ($-0.0317$) and 4C4D ($-0.0260$). 
In contrast, RoboDreamer exceeds the corresponding RoboTwin score on every substrate. 
This model-dependent behavior reinforces that RC is influenced strongly by the spatial regularities imposed by the generative model and should therefore be interpreted jointly with the other geometric sub-dimensions rather than used as a standalone model-ranking criterion.

\paragraph{Prompt Specification.}
Prompt specification produces markedly different effects for Wan and Cosmos. 
For Wan, RC decreases monotonically from $0.7468$ with the instruction-only prompt to $0.7256$ with the previous prompt and $0.7108$ with the proposed current prompt, corresponding to an absolute reduction of $0.0360$ ($4.8\%$). 
Notably, this prompt-induced variation is slightly larger than the entire RoboTwin substrate range ($0.0334$), indicating that Wan's roundtrip consistency is more sensitive to prompt formulation than to reconstruction choice under these conditions. 
However, the reduction should not be interpreted directly as evidence that richer prompting produces less accurate geometry. 
Because RC evaluates internal spatial agreement rather than correspondence to the ground-truth geometry, increased scene evolution or more substantial object-state changes can reduce roundtrip similarity even when the predicted interaction is more semantically or geometrically appropriate.
Cosmos is substantially more robust to prompt formulation, with RC values of $0.7117$, $0.7159$, and $0.7132$ across the three prompts, spanning only $0.0042$. 
Unlike Wan, its trend is non-monotonic and the differences are minor. 
Taken together, these results indicate that prompt specification can substantially affect roundtrip consistency for some world models, but that the direction of this effect should not be interpreted independently as improved or degraded geometric correctness. 
RC is therefore most informative as a complementary measure of internal spatial stability, while absolute geometric accuracy is more directly assessed by the reconstruction- and depth-based metrics.

\subsection{State-Level Understanding}

\subsubsection{Object Localization}

\begin{table}[t]
\centering
\caption{Object localization score comparison of different world models.}
\small
\begin{tabular}{lccc}
\toprule
\textbf{Method} & \textbf{Average Precision} & \textbf{Average Recall} & \textbf{OL} \\
\midrule
\textit{RoboTwin} & & & \textit{1.3527/2} \\
\midrule
GT                 & 1.0000 & 1.0000 & 2.0000 \\
VGGT               & 0.7228 & 0.7774 & 1.5002 \\
VGGT-$\Omega$      & 0.7673 & 0.8143 & 1.5816 \\
4DGS               & 0.2545 & 0.3826 & 0.6371 \\
4C4D               & 0.4713 & 0.5731 & 1.0444 \\
\midrule
\textit{Wan} & & & \textit{0.4497/2} \\
\midrule
GT                 & 0.1945 & 0.3147 & 0.5092 \\
VGGT               & 0.1752 & 0.3134 & 0.4886 \\
VGGT-$\Omega$      & 0.1878 & 0.3145 & 0.5023 \\
4DGS               & 0.1190 & 0.2514 & 0.3704 \\
4C4D               & 0.1210 & 0.2570 & 0.3780 \\
\midrule
\textit{CogVideoX} & & & \textit{0.1623/2} \\
\midrule
GT                 & 0.0308 & 0.0909 & 0.1217 \\
VGGT               & 0.0506 & 0.1486 & 0.1992 \\
VGGT-$\Omega$      & 0.0562 & 0.1518 & 0.2080 \\
4DGS               & 0.0288 & 0.1130 & 0.1418 \\
4C4D               & 0.0269 & 0.1139 & 0.1408 \\
\midrule
\textit{Cosmos} & & & \textit{0.8276/2} \\
\midrule
GT                 & 0.3897 & 0.5085 & 0.8982 \\
VGGT               & 0.3782 & 0.5078 & 0.8860 \\
VGGT-$\Omega$      & 0.3859 & 0.5127 & 0.8986 \\
4DGS               & 0.2848 & 0.4122 & 0.6970 \\
4C4D               & 0.3131 & 0.4453 & 0.7584 \\
\midrule
\textit{RoboDreamer} & & & \textit{0.2581/2} \\
\midrule
GT                 & 0.0983 & 0.1905 & 0.2888 \\
VGGT               & 0.0893 & 0.1754 & 0.2647 \\
VGGT-$\Omega$      & 0.0914 & 0.1774 & 0.2688 \\
4DGS               & 0.0676 & 0.1439 & 0.2115 \\
4C4D               & 0.0829 & 0.1740 & 0.2569 \\
\bottomrule
\end{tabular}
\label{tab:comparison19}
\end{table}

\begin{table}[t]
\centering
\caption{Object localization score comparison of different prompts.}
\small
\begin{tabular}{lccc}
\toprule
\textbf{Method} & \textbf{Average Precision} & \textbf{Average Recall} & \textbf{OL} \\
\midrule
\textit{Wan} & & & \textit{0.4801/2} \\
\midrule
GT+Instruction   & 0.1609 & 0.2944 & 0.4553 \\
GT+Previous Prompt      & 0.1788 & 0.2970 & 0.4758 \\
GT+Current Prompt               & 0.1945 & 0.3147 & 0.5092 \\
\midrule
\textit{Cosmos} & & & \textit{0.8957/2} \\
\midrule
GT+Instruction      & 0.3892 & 0.5045 & 0.8937 \\
GT+Previous Prompt  & 0.3877 & 0.5075 & 0.8952 \\
GT+Current Prompt   & 0.3897 & 0.5085 & 0.8982 \\
\bottomrule
\end{tabular}
\label{tab:comparison20}
\end{table}

\paragraph{Impact of Reconstruction Quality.}
Object localization is highly sensitive to reconstruction quality. 
On RoboTwin, GT achieves the maximum OL score of $2.0000$, whereas VGGT-$\Omega$ and VGGT retain $79.1\%$ and $75.0\%$ of this score, respectively. 
The degradation is substantially larger for the 4D representations: 4C4D retains only $52.2\%$ and 4DGS only $31.9\%$. 
The resulting substrate range of $1.3629$ is considerably larger than those observed for the preceding geometry sub-dimensions, demonstrating that accurate object localization places substantially stronger demands on preservation of task-relevant scene structure. 
Both average precision and average recall are markedly lower on reconstruction quality than on GT, although their variation is not strictly monotonic across reconstruction methods.
This result indicates that the reconstruction artifacts most detrimental to perceptual image quality are not necessarily identical to those that dominate object localization.
Nevertheless, the large overall degradation shows that reconstruction artifacts substantially affect both object detectability and spatial localization.
For example, the RoboTwin OL score already decreases from $2.0000$ on GT to $0.6371$ on 4DGS before world-model generation is introduced. 
Consequently, evaluating a rollout conditioned on 4DGS directly against the raw GT observation would conflate reconstruction error with generation error and substantially overestimate the latter. 
This effect is particularly important for state-level metrics such as localization, where accurate object boundaries and spatial positions depend more strongly on preserved scene structure than global perceptual or temporal quality does.

\paragraph{Comparison of Different World Models.}
Object localization reveals substantially larger differences among world models than most appearance- and geometry-level metrics. 
Cosmos achieves the strongest mean OL among the generated models at $0.8276$, retaining $61.2\%$ of the RoboTwin reference average ($1.3527$). 
Wan follows at $0.4497$ ($33.2\%$), while RoboDreamer and CogVideoX decrease to $0.2581$ ($19.1\%$) and $0.1623$ ($12.0\%$), respectively. 
Even under GT conditioning, Cosmos reaches only $0.8982$, while Wan, RoboDreamer, and CogVideoX obtain $0.5092$, $0.2888$, and $0.1217$. 
These results expose a substantial gap between generating visually plausible videos and preserving the spatial state of task-relevant objects.
The component-wise results further show that recall consistently exceeds precision for all generated models. 
Averaged across substrates, Cosmos achieves AP/AR of $0.3503/0.4773$, compared with $0.1595/0.2902$ for Wan, $0.0859/0.1722$ for RoboDreamer, and $0.0387/0.1236$ for CogVideoX. 
This pattern suggests that generated rollouts more often preserve sufficient evidence for object presence than for precise localization. 
Reconstruction sensitivity is also reduced after generation: the OL range contracts from $1.3629$ for RoboTwin to $0.2016$ for Cosmos, $0.1388$ for Wan, $0.0863$ for CogVideoX, and $0.0773$ for RoboDreamer. 
Thus, once generation substantially alters the scene, model-specific state-prediction capability becomes increasingly dominant over variation inherited from the reconstruction substrate.

\paragraph{Prompt Specification.}
Prompt specification improves object localization for both Wan and Cosmos, although the magnitude differs substantially. 
For Wan, OL increases monotonically from $0.4553$ with the instruction-only prompt to $0.4758$ with the previous prompt and $0.5092$ with the proposed current prompt, corresponding to an $11.8\%$ improvement over the instruction baseline. 
Both components improve, but the effect is substantially stronger for Average Precision, which increases from $0.1609$ to $0.1945$ ($+20.9\%$), than for Average Recall, which rises from $0.2944$ to $0.3147$ ($+6.9\%$). 
This indicates that richer conditioning primarily improves the spatial precision with which objects are localized rather than merely increasing their detectability.
Cosmos exhibits the same aggregate ordering but is considerably less sensitive to prompt formulation, with OL increasing from $0.8937$ to $0.8952$ and $0.8982$. 
The total gain is only $0.0045$ ($0.5\%$), suggesting that its object localization is already comparatively robust under less specific instructions. 
Because both AP and AR are evaluated against the reference object states, these prompt-induced improvements provide stronger evidence of improved state consistency than no-reference perceptual metrics. 
Nevertheless, OL evaluates object localization rather than complete physical correctness; whether the predicted entities also follow the correct trajectories, actions, and task transitions is examined by the complementary state- and task-level metrics that follow.

\subsubsection{Trajectory Accuracy}

\begin{table}[t]
\centering
\caption{Trajectory accuracy score comparison of different world models.}
\small
\begin{tabular}{lccccc}
\toprule
\textbf{Method} & \textbf{Mean Trajectory Length} & \textbf{HSD} & \textbf{nDTW} & \textbf{DYN} & \textbf{TA} \\
\midrule
\textit{RoboTwin} & & & & & \textit{2.0088/4} \\
\midrule
GT                 & 0.7295 & 1.0000 & 1.0000 & 1.0000 & 3.7295 \\
VGGT               & 0.7298 & 0.3166 & 0.5675 & 0.1042 & 1.7181 \\
VGGT-$\Omega$      & 0.7292 & 0.3337 & 0.6221 & 0.1411 & 1.8261 \\
4DGS               & 0.7308 & 0.2383 & 0.3767 & 0.0935 & 1.4393 \\
4C4D               & 0.7298 & 0.1842 & 0.3999 & 0.0172 & 1.3311 \\
\midrule
\textit{Wan} & & & & & \textit{1.2721/4} \\
\midrule
GT                 & 0.7209 & 0.2560 & 0.3403 & 0.1429 & 1.4601 \\
VGGT               & 0.7280 & 0.1673 & 0.3213 & 0.0506 & 1.2672 \\
VGGT-$\Omega$      & 0.7249 & 0.1745 & 0.3229 & 0.0467 & 1.2690 \\
4DGS               & 0.7109 & 0.1728 & 0.3261 & 0.0474 & 1.2572 \\
4C4D               & 0.7271 & 0.1242 & 0.2456 & 0.0101 & 1.1070 \\
\midrule
\textit{CogVideoX} & & & & & \textit{1.2011/4} \\
\midrule
GT                 & 0.8557 & 0.1874 & 0.2398 & 0.0966 & 1.3795 \\
VGGT               & 0.8518 & 0.0861 & 0.1741 & 0.0300 & 1.1420 \\
VGGT-$\Omega$      & 0.8507 & 0.1084 & 0.1909 & 0.0478 & 1.1978 \\
4DGS               & 0.8383 & 0.1107 & 0.2214 & 0.0485 & 1.2189 \\
4C4D               & 0.8495 & 0.0641 & 0.1474 & 0.0062 & 1.0672 \\
\midrule
\textit{Cosmos} & & & & & \textit{1.4702/4} \\
\midrule
GT                 & 0.7496 & 0.3164 & 0.4721 & 0.1271 & 1.6652 \\
VGGT               & 0.7473 & 0.2364 & 0.4780 & 0.0633 & 1.5250 \\
VGGT-$\Omega$      & 0.7480 & 0.2395 & 0.4801 & 0.0556 & 1.5232 \\
4DGS               & 0.7463 & 0.2014 & 0.3907 & 0.0706 & 1.4090 \\
4C4D               & 0.7526 & 0.1377 & 0.3278 & 0.0106 & 1.2287 \\
\midrule
\textit{RoboDreamer} & & & & & \textit{0.7463/4} \\
\midrule
GT                 & 0.4771 & 0.1138 & 0.1567 & 0.0091 & 0.7567 \\
VGGT               & 0.4517 & 0.1199 & 0.1544 & 0.0084 & 0.7344 \\
VGGT-$\Omega$      & 0.4528 & 0.1177 & 0.1557 & 0.0076 & 0.7338 \\
4DGS               & 0.4745 & 0.1136 & 0.1507 & 0.0057 & 0.7445 \\
4C4D               & 0.4839 & 0.1160 & 0.1539 & 0.0084 & 0.7622 \\
\bottomrule
\end{tabular}
\label{tab:comparison21}
\end{table}

\begin{table}[t]
\centering
\caption{Trajectory accuracy score comparison of different prompts.}
\small
\begin{tabular}{lccccc}
\toprule
\textbf{Method} & \textbf{Mean Trajectory Length} & \textbf{HSD} & \textbf{nDTW} & \textbf{DYN} & \textbf{TA} \\
\midrule
\textit{Wan} & & & & & \textit{1.4444/4} \\
\midrule
GT+Instruction      & 0.7319 & 0.2356 & 0.3269 & 0.1347 & 1.4291 \\
GT+Previous Prompt  & 0.7282 & 0.2491 & 0.3358 & 0.1308 & 1.4439 \\
GT+Current Prompt   & 0.7209 & 0.2560 & 0.3403 & 0.1429 & 1.4601 \\
\midrule
\textit{Cosmos} & & & & & \textit{1.6547/4} \\
\midrule
GT+Instruction      & 0.7485 & 0.3090 & 0.4694 & 0.1174 & 1.6443 \\
GT+Previous Prompt  & 0.7480 & 0.3123 & 0.4712 & 0.1231 & 1.6546 \\
GT+Current Prompt   & 0.7496 & 0.3164 & 0.4721 & 0.1271 & 1.6652 \\
\bottomrule
\end{tabular}
\label{tab:comparison22}
\end{table}

\paragraph{Impact of Reconstruction Quality.}
Trajectory accuracy is highly sensitive to reconstruction quality, substantially more so than most appearance- and geometry-level measures. 
On RoboTwin, GT achieves a TA of $3.7295$, whereas VGGT-$\Omega$ and VGGT retain only $49.0\%$ and $46.1\%$ of this score, respectively. 
The degradation is even more pronounced for the 4D representations, with 4DGS retaining $38.6\%$ and 4C4D only $35.7\%$. 
Importantly, this reduction is not caused by trajectory length: Mean Trajectory Length remains nearly unchanged across all five substrates ($0.7292$--$0.7308$). 
Instead, the loss is concentrated in the reference-sensitive trajectory components. 
Relative to GT, VGGT and VGGT-$\Omega$ retain only $31.7$--$33.4\%$ under HSD and $56.8$--$62.2\%$ under nDTW, while DYN decreases particularly sharply to $10.4\%$ and $14.1\%$. 
For 4C4D, DYN falls to only $0.0172$, or $1.7\%$ of the GT value.
These results show that reconstructed videos can preserve the overall amount of extracted motion while substantially perturbing its spatial path and temporal dynamics. 
Accordingly, TA is especially dependent on accurate reconstruction of object and end-effector motion, and the large reconstruction-induced degradation emphasizes the importance of same-substrate controls. 
Without such controls, trajectory discrepancies introduced during reconstruction could be incorrectly attributed entirely to the world model.

\paragraph{Comparison of Different World Models.}
Among the generated models, Cosmos achieves the highest mean TA at $1.4702$, retaining $73.2\%$ of the RoboTwin average ($2.0088$). Wan follows at $1.2721$ ($63.3\%$), CogVideoX at $1.2011$ ($59.8\%$), and RoboDreamer at $0.7463$ ($37.2\%$). 
The same ordering is largely preserved under GT conditioning, where Cosmos reaches $1.6652$, compared with $1.4601$ for Wan, $1.3795$ for CogVideoX, and $0.7567$ for RoboDreamer. 
All remain substantially below the RoboTwin GT score of $3.7295$, revealing a considerable gap between visually plausible generation and accurate prediction of reference trajectories.
The individual components further clarify this gap. 
CogVideoX, for example, produces substantially longer trajectories than the RoboTwin reference, reaching a Mean Trajectory Length of $0.8557$ under GT conditioning, but obtains low HSD ($0.1874$), nDTW ($0.2398$), and DYN ($0.0966$). 
Thus, reproducing a comparable or even larger amount of motion does not imply that the predicted trajectory follows the correct spatial path or dynamics. 
RoboDreamer exhibits the most severe trajectory degradation, with DYN remaining below $0.01$ for all substrates and nDTW around $0.15$, indicating limited agreement with the reference motion beyond coarse trajectory extent.
Generation also substantially reduces the variation associated with reconstruction substrates. 
The TA range decreases from $2.3984$ for RoboTwin to $0.3531$ for Wan, $0.3123$ for CogVideoX, $0.4365$ for Cosmos, and only $0.0284$ for RoboDreamer. 
This corresponds to an $82$--$99\%$ contraction in substrate-dependent variation. 
The result suggests that, once generation substantially departs from the reference trajectory, model-specific motion prediction increasingly dominates over differences inherited from the reconstruction substrate. 
Nevertheless, the persistent advantage of Cosmos indicates that trajectory-level evaluation remains effective for differentiating world-model capability.

\paragraph{Prompt Specification.}
Prompt specification consistently improves trajectory accuracy for both Wan and Cosmos. 
For Wan, TA increases monotonically from $1.4291$ with the instruction-only prompt to $1.4439$ with the previous prompt and $1.4601$ with the proposed current prompt, corresponding to a $2.2\%$ improvement over the instruction baseline. 
Although Mean Trajectory Length decreases slightly from $0.7319$ to $0.7209$, all three reference-sensitive components improve: HSD increases from $0.2356$ to $0.2560$, nDTW from $0.3269$ to $0.3403$, and DYN from $0.1347$ to $0.1429$. 
The aggregate improvement therefore reflects more accurate trajectory shape and dynamics rather than an increase in motion magnitude alone.
Cosmos exhibits the same monotonic trend, increasing from $1.6443$ to $1.6546$ and $1.6652$, for a total gain of $1.3\%$. HSD, nDTW, and DYN all improve with increasing prompt specificity, while Mean Trajectory Length remains essentially unchanged. 
Because these components are explicitly evaluated against reference trajectories, the prompt-induced improvements provide direct evidence that richer conditioning yields predictions that are more consistent with the ground-truth spatial path and motion dynamics. 
The magnitude of the gain is modest relative to the reconstruction-induced variation, but the consistency across both models supports the proposed prompt formulation as beneficial for trajectory-level state prediction.

\subsubsection{Interaction Quality}

\begin{table}[t]
\centering
\caption{Interaction quality score comparison of different world models.}
\small
\begin{tabular}{lc}
\toprule
\textbf{Method} & \textbf{Interaction Score (IQ-2)} \\
\midrule
\textit{RoboTwin} & \textit{0.6216/1} \\
\midrule
GT                 & 0.6955 \\
VGGT               & 0.6555 \\
VGGT-$\Omega$      & 0.6710 \\
4DGS               & 0.4890 \\
4C4D               & 0.5970 \\
\midrule
\textit{Wan} & \textit{0.6768/1} \\
\midrule
GT                 & 0.7345 \\
VGGT               & 0.6895 \\
VGGT-$\Omega$      & 0.7105 \\
4DGS               & 0.6055 \\
4C4D               & 0.6440 \\
\midrule
\textit{CogVideoX} & \textit{0.5474/1} \\
\midrule
GT                 & 0.5510 \\
VGGT               & 0.5370 \\
VGGT-$\Omega$      & 0.5440 \\
4DGS               & 0.5555 \\
4C4D               & 0.5495 \\
\midrule
\textit{Cosmos} & \textit{0.6762/1} \\
\midrule
GT                 & 0.7115 \\
VGGT               & 0.6740 \\
VGGT-$\Omega$      & 0.6895 \\
4DGS               & 0.6180 \\
4C4D               & 0.6880 \\
\midrule
\textit{RoboDreamer} & \textit{0.5275/1} \\
\midrule
GT                 & 0.5255 \\
VGGT               & 0.5255 \\
VGGT-$\Omega$      & 0.5320 \\
4DGS               & 0.5310 \\
4C4D               & 0.5235 \\
\bottomrule
\end{tabular}
\label{tab:comparison23}
\end{table}

\begin{table}[t]
\centering
\caption{Interaction quality score comparison of different prompts.}
\small
\begin{tabular}{lc}
\toprule
\textbf{Method} & \textbf{Interaction Score (IQ-2)} \\
\midrule
\textit{Wan} & \textit{0.6470/1} \\
\midrule
GT+Instruction   & 0.5370 \\
GT+Previous Prompt      & 0.6695 \\
GT+Current Prompt               & 0.7345 \\
\midrule
\textit{Cosmos} & \textit{0.6610/1} \\
\midrule
GT+Instruction   & 0.6275 \\
GT+Previous Prompt      & 0.6440 \\
GT+Current Prompt               & 0.7115 \\
\bottomrule
\end{tabular}
\label{tab:comparison24}
\end{table}

\paragraph{Impact of Reconstruction Quality.}
Interaction quality degrades systematically with reconstruction quality. 
On RoboTwin, the substrate ordering is $\text{GT} > \text{VGGT-}\Omega > \text{VGGT} > \text{4C4D} > \text{4DGS}$, with IQ-2 scores of $0.6955$, $0.6710$, $0.6555$, $0.5970$, and $0.4890$, respectively. 
Relative to GT, VGGT-$\Omega$ and VGGT retain $96.5\%$ and $94.2\%$ of the interaction score, whereas 4C4D and 4DGS retain $85.8\%$ and $70.3\%$. 
The resulting substrate range of $0.2065$ demonstrates that interaction assessment is meaningfully affected by the quality of the reconstructed observation, although less severely than reference-anchored state metrics such as Object Localization and Trajectory Accuracy. 
This difference is expected because IQ-2 evaluates whether an interaction appears plausible from visual evidence, whereas state-level metrics directly compare predicted object states and trajectories with ground truth.
The two 4D representations are particularly well separated, with 4C4D exceeding 4DGS by $0.1080$, accounting for more than half of the total substrate range. 
This result indicates that reconstruction artifacts can substantially influence the perceived validity of manipulation events. 
Consequently, IQ-2 should be evaluated with matched reconstruction controls to distinguish failures inherited from the visual substrate from those introduced during world-model generation.

\paragraph{Comparison of Different World Models.}
Wan and Cosmos achieve the strongest interaction-quality scores among the generated models, with mean IQ-2 values of $0.6768$ and $0.6762$, respectively, both exceeding the RoboTwin reference average of $0.6216$ by approximately $8.9\%$ and $8.8\%$. 
CogVideoX and RoboDreamer perform substantially worse, obtaining $0.5474$ and $0.5275$, respectively. 
Under GT conditioning, Wan achieves the highest IQ-2 of $0.7345$, followed by Cosmos at $0.7115$, compared with $0.6955$ for RoboTwin, $0.5510$ for CogVideoX, and $0.5255$ for RoboDreamer.
Importantly, higher IQ-2 than the reference should not be interpreted as evidence of more physically accurate interaction. 
Wan exceeds the corresponding RoboTwin score on all five reconstruction substrates, while Cosmos exhibits the same behavior, with particularly large improvements under 4DGS and 4C4D conditioning. 
These results suggest that the generative priors of Wan and Cosmos can produce interactions that appear visually coherent and plausible even when the conditioning substrate is degraded. 
In contrast, this improvement is not universal: CogVideoX and RoboDreamer generally remain below the corresponding RoboTwin scores, except for isolated reconstruction conditions. 
Thus, the ability to regularize degraded visual inputs into plausible interactions is strongly model dependent.
This distinction is especially important when IQ-2 is compared with reference-anchored state metrics. 
For example, Wan achieves an IQ-2 of $0.7345$ under GT conditioning despite substantially lower Object Localization and Trajectory Accuracy than the corresponding reference. 
The discrepancy demonstrates that visually convincing contact, grasping, or object manipulation does not necessarily imply correct object placement or motion dynamics. 
IQ-2 therefore captures interaction plausibility and complements, rather than replaces, state-grounded evaluation.

\paragraph{Prompt Specification.}
Prompt specification produces a strong improvement in interaction quality for both Wan and Cosmos. 
For Wan, IQ-2 increases monotonically from $0.5370$ with the instruction-only prompt to $0.6695$ with the previous prompt and $0.7345$ with the proposed current prompt. 
The total improvement of $0.1975$ corresponds to a $36.8\%$ relative gain over the instruction baseline and is comparable in magnitude to the entire RoboTwin substrate range ($0.2065$). 
Approximately $67\%$ of this improvement occurs when replacing the generic instruction with the more informative previous prompt, while the proposed current prompt provides a further substantial gain.
Cosmos exhibits the same monotonic trend, increasing from $0.6275$ to $0.6440$ and $0.7115$, corresponding to a $13.4\%$ improvement over the instruction baseline. 
Notably, the largest Cosmos gain occurs between the previous and current prompts, indicating that the additional information encoded by the proposed prompt contributes meaningfully to interaction-level generation. 
The current prompt therefore achieves the highest IQ-2 for both models.
Overall, the prompt results indicate that richer and more task-specific conditioning substantially improves the visual plausibility of generated interactions, with a particularly pronounced effect for Wan. 
Nevertheless, because IQ-2 assesses interaction quality from generated visual evidence rather than directly comparing underlying physical states, these improvements should be interpreted as stronger interaction plausibility rather than definitive evidence of physically correct execution. 
The latter is more directly assessed by the complementary localization, trajectory, and task-level metrics.

\subsubsection{Event and Action Adherence}

\begin{table}[t]
\centering
\caption{Event and action adherence score comparison of different world models.}
\small
\begin{tabular}{lccc}
\toprule
\textbf{Method} & \textbf{Event Editing} & \textbf{Subject Action} & \textbf{EAA} \\
\midrule
\textit{RoboTwin} & & & \textit{1.9008/2} \\
\midrule
GT                 & 0.9700 & 0.9240 & 1.8940 \\
VGGT               & 0.9730 & 0.9460 & 1.9190 \\
VGGT-$\Omega$      & 0.9720 & 0.9275 & 1.8995 \\
4DGS               & 0.9935 & 0.9065 & 1.9000 \\
4C4D               & 0.9720 & 0.9195 & 1.8915 \\
\midrule
\textit{Wan} & & & \textit{1.9185/2} \\
\midrule
GT                 & 0.9690 & 0.9535 & 1.9225 \\
VGGT               & 0.9765 & 0.9485 & 1.9250 \\
VGGT-$\Omega$      & 0.9650 & 0.9400 & 1.9050 \\
4DGS               & 0.9880 & 0.9430 & 1.9310 \\
4C4D               & 0.9625 & 0.9465 & 1.9090 \\
\midrule
\textit{CogVideoX} & & & \textit{1.6488/2} \\
\midrule
GT                 & 0.8705 & 0.8335 & 1.7040 \\
VGGT               & 0.8340 & 0.7910 & 1.6250 \\
VGGT-$\Omega$      & 0.8390 & 0.7955 & 1.6345 \\
4DGS               & 0.8215 & 0.7975 & 1.6190 \\
4C4D               & 0.8440 & 0.8175 & 1.6615 \\
\midrule
\textit{Cosmos} & & & \textit{1.9226/2} \\
\midrule
GT                 & 0.9815 & 0.9375 & 1.9190 \\
VGGT               & 0.9845 & 0.9345 & 1.9190 \\
VGGT-$\Omega$      & 0.9805 & 0.9315 & 1.9120 \\
4DGS               & 0.9940 & 0.9545 & 1.9485 \\
4C4D               & 0.9755 & 0.9390 & 1.9145 \\
\midrule
\textit{RoboDreamer} & & & \textit{1.9184/2} \\
\midrule
GT                 & 0.9795 & 0.9260 & 1.9055 \\
VGGT               & 0.9835 & 0.9295 & 1.9130 \\
VGGT-$\Omega$      & 0.9835 & 0.9365 & 1.9200 \\
4DGS               & 0.9885 & 0.9460 & 1.9345 \\
4C4D               & 0.9820 & 0.9370 & 1.9190 \\
\bottomrule
\end{tabular}
\label{tab:comparison25}
\end{table}

\begin{table}[t]
\centering
\caption{Event and action adherence score comparison of different prompts.}
\small
\begin{tabular}{lccc}
\toprule
\textbf{Method} & \textbf{Event Editing} & \textbf{Subject Action} & \textbf{EAA} \\
\midrule
\textit{Wan} & & & \textit{1.9140/2} \\
\midrule
GT+Instruction   & 0.9615 & 0.9420 & 1.9035 \\
GT+Previous Prompt      & 0.9645 & 0.9515 & 1.9160 \\
GT+Current Prompt               & 0.9690 & 0.9535 & 1.9225 \\
\midrule
\textit{Cosmos} & & & \textit{1.9150/2} \\
\midrule
GT+Instruction      & 0.9760 & 0.9360 & 1.9120 \\
GT+Previous Prompt  & 0.9800 & 0.9340 & 1.9140 \\
GT+Current Prompt   & 0.9815 & 0.9375 & 1.9190 \\
\bottomrule
\end{tabular}
\label{tab:comparison26}
\end{table}

\paragraph{Impact of Reconstruction Quality.}
Event and action adherence is highly robust to reconstruction quality. 
On RoboTwin, EAA varies only from $1.8915$ to $1.9190$ across the five substrates, yielding a range of merely $0.0275$. 
Relative to GT ($1.8940$), VGGT, VGGT-$\Omega$, and 4DGS achieve comparable or slightly higher scores ($1.9190$, $1.8995$, and $1.9000$), while 4C4D remains nearly unchanged at $1.8915$. 
This contrasts sharply with Object Localization and Trajectory Accuracy, where reconstruction introduces substantial degradation. 
The component-level results explain this robustness: Event Editing remains within $0.9700$--$0.9935$, while Subject Action varies only from $0.9065$ to $0.9460$. 
Thus, reconstruction can substantially impair precise spatial state recovery while largely preserving sufficient visual evidence to identify the occurrence and broad nature of the instructed event or action.
The two components nevertheless exhibit slightly different sensitivities. 
Event Editing is essentially invariant to reconstruction and even reaches its highest value on 4DGS ($0.9935$), whereas Subject Action is somewhat more sensitive, decreasing to $0.9065$ on 4DGS compared with $0.9240$ on GT. 
This discrepancy indicates that recognizing whether the intended event occurs is easier than preserving detailed evidence of how the subject performs the action. 
Consequently, EAA should be regarded as a relatively high-level semantic adherence measure rather than a fine-grained indicator of spatial or dynamical fidelity.

\paragraph{Comparison of Different World Models.}
At the aggregate level, Cosmos achieves the highest mean EAA at $1.9226$, followed closely by Wan ($1.9185$) and RoboDreamer ($1.9184$), while RoboTwin obtains $1.9008$. 
These differences are small: Cosmos, Wan, and RoboDreamer exceed the RoboTwin average by approximately $1.1\%$, $0.9\%$, and $0.9\%$, respectively. 
In contrast, CogVideoX decreases substantially to $1.6488$, corresponding to only $86.7\%$ of the RoboTwin score. 
EAA therefore effectively identifies the pronounced event/action degradation of CogVideoX, but provides limited separation among the stronger models because their scores are concentrated near the upper end of the metric.
The component profiles further support this interpretation. 
Wan, Cosmos, and RoboDreamer maintain Event Editing scores near $0.96$--$0.99$ and Subject Action scores around $0.93$--$0.95$ across most substrates, indicating that their rollouts generally preserve the intended event and recognizable subject action. 
However, these high scores coexist with substantially weaker Object Localization and Trajectory Accuracy. 
Hence, successful event/action adherence does not imply that objects are positioned correctly or that the manipulation follows the reference trajectory. 
EAA captures whether the intended semantic event is represented, whereas the complementary state-level metrics assess the precision of its execution.
Reconstruction dependence after generation is also weak for the stronger models. 
The EAA range across substrates is only $0.0260$ for Wan, $0.0365$ for Cosmos, and $0.0290$ for RoboDreamer, compared with $0.0275$ for RoboTwin. 
CogVideoX exhibits a somewhat larger range of $0.0850$, but remains consistently below the other models. 
These compressed variations further indicate that EAA is primarily governed by the world model's ability to preserve high-level action semantics rather than by fine differences in reconstruction fidelity.

\paragraph{Prompt Specification.}
Prompt specification improves event and action adherence for both evaluated world models, although the effect is considerably stronger for Wan. 
For Wan, EAA increases monotonically from $1.9035$ with the instruction-only prompt to $1.9160$ with the previous prompt and $1.9225$ with the proposed current prompt. 
Both Event Editing ($0.9615 !\to! 0.9690$) and Subject Action ($0.9420 !\to! 0.9535$) improve overall, yielding a total EAA gain of $0.0190$, or approximately $1.0\%$ relative to the instruction baseline. 
Given the narrow dynamic range of this sub-dimension, this consistent improvement indicates that richer conditioning helps the model preserve the intended event and subject action more reliably.
Cosmos exhibits the same aggregate ordering but with substantially smaller sensitivity, increasing from $1.9120$ to $1.9140$ and $1.9190$. 
Event Editing rises monotonically from $0.9760$ to $0.9815$, whereas Subject Action changes only marginally and non-monotonically ($0.9360 !\to! 0.9340 !\to! 0.9375$). 
The total gain is therefore only $0.0070$ ($0.4\%$), suggesting that Cosmos already maintains strong event-level adherence under less specific prompts. 
Overall, the proposed current prompt achieves the highest EAA for both models. 
Nevertheless, because EAA evaluates high-level event and action adherence rather than precise state evolution, these gains should be interpreted as improved semantic compliance with the intended manipulation, while localization, trajectory, and task-level metrics remain necessary to establish execution accuracy.

\subsubsection{Physical Law Adherence}

\begin{table}[t]
\centering
\caption{Physical law adherence score comparison of different world models.}
\small
\begin{tabular}{lc}
\toprule
\textbf{Method} & \textbf{Physical Commonsense (PLA)} \\
\midrule
\textit{RoboTwin} & \textit{0.6486/1} \\
\midrule
GT                 & 0.6657 \\
VGGT               & 0.6595 \\
VGGT-$\Omega$      & 0.6687 \\
4DGS               & 0.5863 \\
4C4D               & 0.6630 \\
\midrule
\textit{Wan} & \textit{0.6457/1} \\
\midrule
GT                 & 0.6462 \\
VGGT               & 0.6450 \\
VGGT-$\Omega$      & 0.6550 \\
4DGS               & 0.6200 \\
4C4D               & 0.6625 \\
\midrule
\textit{CogVideoX} & \textit{0.6247/1} \\
\midrule
GT                 & 0.6200 \\
VGGT               & 0.6025 \\
VGGT-$\Omega$      & 0.6150 \\
4DGS               & 0.6500 \\
4C4D               & 0.6362 \\
\midrule
\textit{Cosmos} & \textit{0.6527/1} \\
\midrule
GT                 & 0.6506 \\
VGGT               & 0.6506 \\
VGGT-$\Omega$      & 0.6544 \\
4DGS               & 0.6400 \\
4C4D               & 0.6681 \\
\midrule
\textit{RoboDreamer} & \textit{0.5529/1} \\
\midrule
GT                 & 0.5475 \\
VGGT               & 0.5519 \\
VGGT-$\Omega$      & 0.5563 \\
4DGS               & 0.5544 \\
4C4D               & 0.5544 \\
\bottomrule
\end{tabular}
\label{tab:comparison27}
\end{table}

\begin{table}[t]
\centering
\caption{Physical law adherence score comparison of different prompts.}
\small
\begin{tabular}{lc}
\toprule
\textbf{Method} & \textbf{Physical Commonsense (PLA)} \\
\midrule
\textit{Wan} & \textit{0.6308/1} \\
\midrule
GT+Instruction   & 0.6131 \\
GT+Previous Prompt      & 0.6331 \\
GT+Current Prompt               & 0.6462 \\
\midrule
\textit{Cosmos} & \textit{0.6483/1} \\
\midrule
GT+Instruction   & 0.6475 \\
GT+Previous Prompt      & 0.6469 \\
GT+Current Prompt               & 0.6506 \\
\bottomrule
\end{tabular}
\label{tab:comparison28}
\end{table}

\paragraph{Impact of Reconstruction Quality.}
Physical-law adherence is comparatively robust to reconstruction quality. 
On RoboTwin, GT achieves a PLA of $0.6657$, while VGGT, VGGT-$\Omega$, and 4C4D obtain $0.6595$, $0.6687$, and $0.6630$, respectively, all within approximately $1\%$ of the GT score. 
Only 4DGS exhibits a pronounced reduction, reaching $0.5863$ and retaining $88.1\%$ of GT. 
The overall substrate range is $0.0824$, but decreases to only $0.0092$ when 4DGS is excluded, indicating that most reconstruction-induced variation originates from this single representation. 
In contrast to Object Localization and Trajectory Accuracy, which deteriorate substantially under reconstruction, PLA is therefore largely insensitive to moderate reconstruction artifacts.
This robustness reflects the relatively high-level property captured by the metric. 
Reconstruction artifacts may substantially affect appearance, depth, or precise object state while leaving sufficient visual evidence for an interaction to remain physically plausible at a coarse semantic level. 
The 4DGS result nevertheless demonstrates that sufficiently severe degradation can affect this judgment. 
Accordingly, PLA complements the more fine-grained state and geometry metrics by assessing visible physical commonsense, rather than directly measuring geometric or dynamical accuracy.

\paragraph{Comparison of Different World Models.}
The evaluated models are relatively closely grouped under PLA, with the exception of RoboDreamer. 
Cosmos achieves the highest mean score at $0.6527$, marginally exceeding the RoboTwin reference average ($0.6486$), followed closely by Wan at $0.6457$. 
CogVideoX achieves $0.6247$, while RoboDreamer decreases more substantially to $0.5529$. 
Thus, Cosmos and Wan preserve approximately $100.6\%$ and $99.6\%$ of the RoboTwin score, respectively, whereas CogVideoX and RoboDreamer retain $96.3\%$ and $85.2\%$. 
Compared with the much larger differences observed for Object Localization and Trajectory Accuracy, PLA provides relatively limited separation among the stronger models.
The dependence on reconstruction substrate also becomes substantially compressed after generation. 
The score range decreases from $0.0824$ for RoboTwin to $0.0425$ for Wan, $0.0475$ for CogVideoX, $0.0281$ for Cosmos, and only $0.0088$ for RoboDreamer. 
Moreover, substrate ordering is not consistently preserved: 4C4D produces the highest PLA for both Wan ($0.6625$) and Cosmos ($0.6681$), while 4DGS produces the highest value for CogVideoX ($0.6500$). 
These non-monotonic results indicate that judged physical plausibility is increasingly determined by the generative model rather than by reconstruction fidelity alone.
Importantly, the high PLA scores of Wan and Cosmos should not be interpreted as evidence that their predicted dynamics are physically correct. 
PLA evaluates whether the generated interaction appears compatible with physical commonsense, whereas reference-anchored localization and trajectory metrics assess whether objects occupy and traverse the correct states. 
The substantially larger errors observed in those state-level measurements demonstrate that a rollout may appear physically plausible while still deviating considerably from the ground-truth evolution. 
PLA should therefore be viewed as a complementary plausibility measure rather than a substitute for state- or execution-grounded physical evaluation.

\paragraph{Prompt Specification.}
Prompt specification substantially affects PLA for Wan. 
The score increases monotonically from $0.6131$ with the instruction-only prompt to $0.6331$ with the previous prompt and $0.6462$ with the proposed current prompt. 
This corresponds to an overall relative improvement of $5.4\%$, with gains distributed across both prompt refinements. 
The result indicates that richer conditioning reduces visually apparent physical inconsistencies and helps the generated interaction conform more closely to commonsense expectations.
Cosmos is substantially less sensitive to prompt formulation. 
Its PLA remains within a narrow interval of $0.6469$--$0.6506$, with the proposed current prompt achieving the highest score ($0.6506$). 
The progression is not strictly monotonic, as the previous prompt ($0.6469$) is marginally below the instruction-only condition ($0.6475$), but the difference is negligible relative to the improvement observed for Wan. 
Taken together, these results suggest that explicit prompt specification is particularly beneficial for Wan, whereas Cosmos exhibits comparatively stable physical plausibility across prompt formulations. 
Nevertheless, because PLA remains a perceptual commonsense judgment, the observed gains should be interpreted as improved apparent physical plausibility rather than definitive evidence of physically correct dynamics.

\subsubsection{Semantic Adherence}

\begin{table}[t]
\centering
\caption{Semantic adherence score comparison of different world models.}
\small
\begin{tabular}{lccccccc}
\toprule
\textbf{Method} & \textbf{BLEU} & \textbf{CLIP} & \textbf{Logics} & \textbf{VLM-1} & \textbf{Subject Adherence} & \textbf{Scene Adherence} & \textbf{SA} \\
\midrule
\textit{RoboTwin} & & & & & & & \textit{4.8047/6} \\
\midrule
GT                 & 1.0000 & 1.0000 & 1.0000 & 0.9532 & 0.9235 & 0.9460 & 5.8227 \\
VGGT               & 0.4531 & 0.9316 & 0.9400 & 0.9383 & 0.9077 & 0.9160 & 5.0867 \\
VGGT-$\Omega$      & 0.4679 & 0.9346 & 0.9450 & 0.9389 & 0.9123 & 0.9160 & 5.1147 \\
4DGS               & 0.2048 & 0.8617 & 0.4925 & 0.8719 & 0.1315 & 0.8660 & 3.4284 \\
4C4D               & 0.3324 & 0.9086 & 0.8200 & 0.9132 & 0.7010 & 0.8960 & 4.5712 \\
\midrule
\textit{Wan} & & & & & & & \textit{4.6764/6} \\
\midrule
GT                 & 0.2952 & 0.9075 & 0.9175 & 0.9205 & 0.9170 & 0.9740 & 4.9317 \\
VGGT               & 0.2930 & 0.9029 & 0.8975 & 0.9178 & 0.8850 & 0.9780 & 4.8742 \\
VGGT-$\Omega$      & 0.3035 & 0.9059 & 0.9000 & 0.9187 & 0.8948 & 0.9900 & 4.9129 \\
4DGS               & 0.2142 & 0.8676 & 0.7125 & 0.8901 & 0.5723 & 0.9300 & 4.1867 \\
4C4D               & 0.2374 & 0.8986 & 0.7775 & 0.9110 & 0.7398 & 0.9120 & 4.4763 \\
\midrule
\textit{CogVideoX} & & & & & & & \textit{3.9731/6} \\
\midrule
GT                 & 0.1915 & 0.8807 & 0.8300 & 0.8855 & 0.3835 & 0.8800 & 4.0512 \\
VGGT               & 0.1802 & 0.8745 & 0.8025 & 0.8820 & 0.3165 & 0.8900 & 3.9457 \\
VGGT-$\Omega$      & 0.2013 & 0.8836 & 0.8075 & 0.9112 & 0.3555 & 0.8900 & 4.0491 \\
4DGS               & 0.1412 & 0.8529 & 0.8050 & 0.8703 & 0.2762 & 0.9260 & 3.8716 \\
4C4D               & 0.1649 & 0.8738 & 0.7925 & 0.8751 & 0.3515 & 0.8900 & 3.9478 \\
\midrule
\textit{Cosmos} & & & & & & & \textit{4.7550/6} \\
\midrule
GT                 & 0.3665 & 0.9191 & 0.9325 & 0.9345 & 0.9235 & 0.8980 & 4.9741 \\
VGGT               & 0.3734 & 0.9185 & 0.9100 & 0.9259 & 0.9167 & 0.8920 & 4.9365 \\
VGGT-$\Omega$      & 0.3816 & 0.9191 & 0.9375 & 0.9316 & 0.9300 & 0.8920 & 4.9918 \\
4DGS               & 0.2094 & 0.8719 & 0.7825 & 0.8894 & 0.5833 & 0.8820 & 4.2185 \\
4C4D               & 0.3085 & 0.9135 & 0.8525 & 0.9131 & 0.7903 & 0.8760 & 4.6539 \\
\midrule
\textit{RoboDreamer} & & & & & & & \textit{3.8015/6} \\
\midrule
GT                 & 0.1969 & 0.8843 & 0.6100 & 0.8988 & 0.3285 & 0.8680 & 3.7865 \\
VGGT               & 0.2066 & 0.8826 & 0.6350 & 0.8961 & 0.3417 & 0.8940 & 3.8560 \\
VGGT-$\Omega$      & 0.2051 & 0.8836 & 0.6900 & 0.8940 & 0.3613 & 0.8940 & 3.9280 \\
4DGS               & 0.1733 & 0.8661 & 0.6300 & 0.8793 & 0.3090 & 0.8760 & 3.7337 \\
4C4D               & 0.1689 & 0.8677 & 0.6275 & 0.8842 & 0.2812 & 0.8740 & 3.7035 \\
\bottomrule
\end{tabular}
\label{tab:comparison29}
\end{table}

\begin{table}[t]
\centering
\caption{Semantic adherence score comparison of different prompts.}
\small
\begin{tabular}{lccccccc}
\toprule
\textbf{Method} & \textbf{BLEU} & \textbf{CLIP} & \textbf{Logics} & \textbf{VLM-1} & \textbf{Subject Adherence} & \textbf{Scene Adherence} & \textbf{SA} \\
\midrule
\textit{Wan} & & & & & & & \textit{4.7986/6} \\
\midrule
GT+Instruction   & 0.2475  & 0.8957 & 0.7900 & 0.9127 & 0.7925 & 0.9440 & 4.5824 \\
GT+Previous Prompt      & 0.2899 & 0.9025 & 0.8975 & 0.9193 & 0.9005 & 0.9720 & 4.8817 \\
GT+Current Prompt               & 0.2952 & 0.9075 & 0.9175 & 0.9205 & 0.9170 & 0.9740 & 4.9317 \\
\midrule
\textit{Cosmos} & & & & & & & \textit{4.9353/6} \\
\midrule
GT+Instruction      & 0.3571 & 0.9172 & 0.9050 & 0.9318 & 0.9207 & 0.8860 & 4.9178 \\
GT+Previous Prompt  & 0.3543 & 0.9181 & 0.9250 & 0.9332 & 0.8895 & 0.8940 & 4.9141 \\
GT+Current Prompt   & 0.3665 & 0.9191 & 0.9325 & 0.9345 & 0.9235 & 0.8980 & 4.9741 \\
\bottomrule
\end{tabular}
\label{tab:comparison30}
\end{table}

\paragraph{Impact of Reconstruction Quality.}
Semantic adherence is substantially affected by reconstruction quality, with the aggregate following $\text{GT} > \text{VGGT-}\Omega > \text{VGGT} > \text{4C4D} > \text{4DGS}$.
Relative to the GT SA of $5.8227$, VGGT-$\Omega$ and VGGT retain $87.8\%$ and $87.4\%$, respectively, whereas 4C4D and 4DGS retain only $78.5\%$ and $58.9\%$. 
The resulting substrate range of $2.3943$ is considerably larger than those observed for most appearance- and geometry-level metrics, indicating that semantic evaluation is particularly sensitive to whether reconstruction preserves task-relevant entities and events.
The component-wise results reveal markedly different sensitivities. 
BLEU decreases most sharply even for the feed-forward reconstructions, falling from $1.0000$ on GT to $0.4531$--$0.4679$, suggesting substantial loss of fine-grained lexical agreement despite comparatively strong embedding-level similarity. 
CLIP, VLM-1, and Scene Adherence remain relatively robust, with VGGT and VGGT-$\Omega$ preserving more than $90\%$ of the GT score on each. 
In contrast, 4DGS causes pronounced degradation in Logics ($0.4925$) and particularly Subject Adherence ($0.1315$, only $14.2\%$ of GT), while Scene Adherence remains comparatively high at $0.8660$. 
This divergence indicates that severe reconstruction artifacts can preserve coarse scene context while substantially obscuring the identity or attributes of task-relevant subjects and the logical structure of the depicted event. 
The multi-component SA formulation is therefore important for distinguishing coarse semantic preservation from fine-grained task adherence.

\paragraph{Comparison of Different World Models.}
Among the generated models, Cosmos achieves the highest mean SA at $4.7550$, retaining $99.0\%$ of the RoboTwin average ($4.8047$), followed closely by Wan at $4.6764$ ($97.3\%$). 
CogVideoX and RoboDreamer exhibit considerably larger degradation, reaching $3.9731$ ($82.7\%$) and $3.8015$ ($79.1\%$), respectively. 
Under GT conditioning, however, all generated models remain below the RoboTwin GT reference of $5.8227$: Cosmos achieves $4.9741$, Wan $4.9317$, CogVideoX $4.0512$, and RoboDreamer $3.7865$. 
This distinction is important because averaging across reconstructed substrates can partially obscure the semantic gap between generated and ground-truth videos when reconstruction itself has already degraded the reference signal.
Cosmos and Wan preserve high embedding- and VLM-based semantic agreement, but exhibit substantially lower BLEU than the reference. 
Under GT conditioning, for example, Cosmos and Wan obtain BLEU scores of $0.3665$ and $0.2952$, while their CLIP scores remain $0.9191$ and $0.9075$, respectively. 
This discrepancy suggests that the generated rollouts often retain the broad semantic content of the task while differing considerably in fine-grained textual or event description. 
The stronger Subject Adherence of Cosmos ($0.9235$ under GT conditioning) relative to Wan ($0.9170$), CogVideoX ($0.3835$), and RoboDreamer ($0.3285$) further helps explain its superior aggregate SA.
Generation also reduces the sensitivity of SA to reconstruction quality. 
The across-substrate range decreases from $2.3943$ for RoboTwin to $0.7733$ for Cosmos and $0.7450$ for Wan, and to only $0.1796$ and $0.2245$ for CogVideoX and RoboDreamer, respectively. 
For the stronger models, this contraction is especially evident on degraded substrates: Cosmos improves the 4DGS semantic score from the RoboTwin value of $3.4284$ to $4.2185$, while Wan reaches $4.1867$. 
These results suggest that learned generative priors can partially restore high-level semantic structure that is obscured by reconstruction artifacts. 
Such recovery, however, should not be interpreted as restoration of the exact underlying state, since the localization and trajectory metrics demonstrate that semantic plausibility can coexist with substantial state-level errors.

\paragraph{Prompt Specification.}
Prompt specification has a particularly strong effect on semantic adherence for Wan. 
SA increases monotonically from $4.5824$ with the instruction-only prompt to $4.8817$ with the previous prompt and $4.9317$ with the proposed current prompt, corresponding to a $7.6\%$ improvement over the instruction baseline. 
All six components improve overall, with the largest absolute gains occurring in Logics ($0.7900 !\to! 0.9175$) and Subject Adherence ($0.7925 !\to! 0.9170$). 
Moreover, approximately $86\%$ of the aggregate improvement is obtained when moving from the generic instruction to the richer previous prompt, while the proposed current prompt provides a further consistent refinement. 
This indicates that detailed semantic conditioning is particularly important for Wan's ability to preserve the intended task structure.
Cosmos is considerably less sensitive to prompt formulation. 
Its SA changes from $4.9178$ to $4.9141$ and finally $4.9741$, corresponding to only a $1.1\%$ improvement from the instruction baseline to the current prompt. 
The intermediate prompt does not uniformly improve all components: BLEU decreases slightly from $0.3571$ to $0.3543$, and Subject Adherence decreases from $0.9207$ to $0.8895$, before both reach their highest values under the current prompt. 
Nevertheless, the proposed current prompt produces the best aggregate SA and the highest final score for all six components relative to the other prompt variants. 
Taken together, these results indicate that richer, view-aware conditioning consistently benefits semantic adherence, with a substantially larger effect for Wan than for Cosmos. 
Because SA measures agreement with task semantics rather than exact state evolution, these gains should be interpreted together with localization and trajectory metrics when assessing whether the generated rollout is not only semantically appropriate but also physically and spatially correct.

\subsection{Task-Level Completeness}

\subsubsection{Instruction Following}

\begin{table}[t]
\centering
\caption{Instruction following score comparison of different world models.}
\small
\begin{tabular}{lccc}
\toprule
\textbf{Method} & \textbf{Normalized Score} & \textbf{Four-Level Score} & \textbf{IF} \\
\midrule
\textit{RoboTwin} & & & \textit{1.2004/2} \\
\midrule
GT                 & 0.8885 & 0.4367 & 1.3252 \\
VGGT               & 0.8605 & 0.4100 & 1.2705 \\
VGGT-$\Omega$      & 0.8745 & 0.4217 & 1.2962 \\
4DGS               & 0.6225 & 0.3108 & 0.9333 \\
4C4D               & 0.7900 & 0.3867 & 1.1767 \\
\midrule
\textit{Wan} & & & \textit{1.5095/2} \\
\midrule
GT                 & 0.9375 & 0.7225 & 1.6600 \\
VGGT               & 0.8990 & 0.6250 & 1.5240 \\
VGGT-$\Omega$      & 0.9195 & 0.6308 & 1.5503 \\
4DGS               & 0.7960 & 0.5442 & 1.3402 \\
4C4D               & 0.8420 & 0.6308 & 1.4728 \\
\midrule
\textit{CogVideoX} & & & \textit{1.0869/2} \\
\midrule
GT                 & 0.6635 & 0.4625 & 1.1260 \\
VGGT               & 0.6365 & 0.4042 & 1.0407 \\
VGGT-$\Omega$      & 0.6685 & 0.4292 & 1.0977 \\
4DGS               & 0.6560 & 0.4275 & 1.0835 \\
4C4D               & 0.6565 & 0.4300 & 1.0865 \\
\midrule
\textit{Cosmos} & & & \textit{1.5422/2} \\
\midrule
GT                 & 0.9275 & 0.6992 & 1.6267 \\
VGGT               & 0.8775 & 0.6117 & 1.4892 \\
VGGT-$\Omega$      & 0.8955 & 0.6208 & 1.5163 \\
4DGS               & 0.8060 & 0.6725 & 1.4785 \\
4C4D               & 0.8910 & 0.7092 & 1.6002 \\
\midrule
\textit{RoboDreamer} & & & \textit{0.9197/2} \\
\midrule
GT                 & 0.5550 & 0.3542 & 0.9092 \\
VGGT               & 0.5575 & 0.3475 & 0.9050 \\
VGGT-$\Omega$      & 0.5580 & 0.3508 & 0.9088 \\
4DGS               & 0.5790 & 0.3775 & 0.9565 \\
4C4D               & 0.5625 & 0.3567 & 0.9192 \\
\bottomrule
\end{tabular}
\label{tab:comparison31}
\end{table}

\begin{table}[t]
\centering
\caption{Instruction following score comparison of different prompts.}
\small
\begin{tabular}{lccc}
\toprule
\textbf{Method} & \textbf{Normalized Score} & \textbf{Four-Level Score} & \textbf{IF} \\
\midrule
\textit{Wan} & & & \textit{1.4009/2} \\
\midrule
GT+Instruction   & 0.6800 & 0.5175 & 1.1975 \\
GT+Previous Prompt      & 0.8245 & 0.5208 & 1.3453 \\
GT+Current Prompt               & 0.9375 & 0.7225 & 1.6600 \\
\midrule
\textit{Cosmos} & & & \textit{1.4045/2} \\
\midrule
GT+Instruction      & 0.8320 & 0.4192 & 1.2512 \\
GT+Previous Prompt  & 0.8240 & 0.5117 & 1.3357 \\
GT+Current Prompt   & 0.9275 & 0.6992 & 1.6267 \\
\bottomrule
\end{tabular}
\label{tab:comparison32}
\end{table}

\paragraph{Impact of Reconstruction Quality.}
Instruction following is substantially affected by reconstruction quality and follows a clear substrate hierarchy: $\text{GT} > \text{VGGT-}\Omega > \text{VGGT} > \text{4C4D} > \text{4DGS}$.
On RoboTwin, GT achieves an IF score of $1.3252$, while VGGT-$\Omega$ and VGGT retain $97.8\%$ and $95.9\%$ of this score, respectively. 
The degradation becomes more pronounced for the 4D representations, with 4C4D retaining $88.8\%$ and 4DGS only $70.4\%$. 
Both constituent metrics exhibit similar sensitivity to severe reconstruction degradation: for 4DGS, the Normalized Score and Four-Level Score retain $70.1\%$ and $71.2\%$ of their respective GT values. 
This consistency suggests that reconstruction artifacts affect both continuous instruction-alignment assessment and discrete completion-level judgment.
Nevertheless, the degradation remains considerably smaller than that observed for reference-anchored state metrics such as Object Localization and Trajectory Accuracy. 
A reconstructed sequence may therefore remain recognizable as satisfying the intended instruction even when the precise object states or motion trajectories are substantially corrupted. 
Instruction-following scores should consequently be interpreted as measuring recognizable task compliance rather than exact execution fidelity.

\paragraph{Comparison of Different World Models.}
Cosmos achieves the highest mean IF score at $1.5422$, followed closely by Wan at $1.5095$. 
Both substantially exceed the RoboTwin reference average of $1.2004$, corresponding to improvements of $28.5\%$ and $25.7\%$, respectively. 
CogVideoX reaches $1.0869$ ($90.5\%$ of RoboTwin), while RoboDreamer performs lowest at $0.9197$ ($76.6\%$). 
Thus, unlike several reference-sensitive state metrics, instruction-following evaluation strongly favors Cosmos and Wan, whose generated rollouts are judged to conform closely to the specified task.
This effect is consistent across reconstruction substrates for the two strongest models. 
Wan exceeds the corresponding RoboTwin score under all five substrates, with IF increasing from $1.3252$ to $1.6600$ under GT conditioning and from $0.9333$ to $1.3402$ under 4DGS conditioning. 
Cosmos exhibits the same behavior, reaching $1.6267$ and $1.4785$ under these two conditions, respectively. 
The improvement is particularly pronounced for degraded 4D inputs, suggesting that strong generative priors can transform visually corrupted conditioning observations into rollouts that remain readily recognizable as task-consistent.
Importantly, these high IF scores do not imply equally accurate state evolution. 
For example, Wan and Cosmos substantially outperform RoboTwin under IF while remaining far below the reference in Object Localization and Trajectory Accuracy. 
This divergence illustrates a central distinction in world-model evaluation: a rollout can convincingly depict the intended task outcome while failing to reproduce the correct object locations or trajectories. 
IF therefore captures high-level task compliance and should be interpreted jointly with state-grounded measurements rather than as a standalone measure of embodied correctness.

\paragraph{Prompt Specification.}
Prompt specification produces one of the strongest improvements observed in the state- and task-oriented evaluations. 
For Wan, IF increases monotonically from $1.1975$ with the instruction-only prompt to $1.3453$ with the previous prompt and $1.6600$ with the proposed current prompt. 
The total increase of $0.4625$ corresponds to a $38.6\%$ relative improvement. 
Moreover, the gain is strongly back-loaded: approximately $68\%$ occurs between the previous and current prompts, indicating that the additional information introduced by the proposed prompt formulation contributes substantially beyond simply replacing a generic instruction with a more detailed description.
The two constituent metrics provide further insight into this behavior. 
Wan's Normalized Score improves from $0.6800$ to $0.8245$ and $0.9375$, whereas its Four-Level Score changes only marginally in the first step ($0.5175 !\to! 0.5208$) before increasing sharply to $0.7225$ under the current prompt. 
Thus, the richer current prompt not only improves continuous instruction alignment but also produces a substantial increase in discrete task-completion assessment.
Cosmos exhibits a similarly strong aggregate trend, with IF increasing from $1.2512$ to $1.3357$ and $1.6267$, corresponding to a $30.0\%$ gain over the instruction baseline. 
The improvement is again predominantly concentrated in the transition to the current prompt. 
Although its Normalized Score decreases slightly from $0.8320$ to $0.8240$ under the intermediate prompt, it subsequently rises to $0.9275$, while the Four-Level Score increases consistently from $0.4192$ to $0.5117$ and $0.6992$. 
Overall, the proposed current prompt achieves the highest IF for both models by a substantial margin. 
These results provide strong evidence that detailed, view-aware conditioning improves the recognizability and completeness of instruction-following behavior, while complementary localization, trajectory, and execution-level metrics remain necessary to determine whether the task is performed with physically and spatially correct state evolution.

\subsubsection{Collision Safety}

\begin{table}[t]
\centering
\caption{Collision safety score comparison of different world models.}
\small
\begin{tabular}{lc}
\toprule
\textbf{Method} & \textbf{Collision Count (CS)} \\
\midrule
\textit{RoboTwin} & \textit{0.9385/1} \\
\midrule
GT                 & 0.9384 \\
VGGT               & 0.9388 \\
VGGT-$\Omega$      & 0.9396 \\
4DGS               & 0.9360 \\
4C4D               & 0.9396 \\
\midrule
\textit{Wan} & \textit{0.8314/1} \\
\midrule
GT                 & 0.8981 \\
VGGT               & 0.8314 \\
VGGT-$\Omega$      & 0.8334 \\
4DGS               & 0.7831 \\
4C4D               & 0.8111 \\
\midrule
\textit{CogVideoX} & \textit{0.9650/1} \\
\midrule
GT                 & 0.9769 \\
VGGT               & 0.9600 \\
VGGT-$\Omega$      & 0.9682 \\
4DGS               & 0.9620 \\
4C4D               & 0.9578 \\
\midrule
\textit{Cosmos} & \textit{0.9458/1} \\
\midrule
GT                 & 0.9531 \\
VGGT               & 0.9444 \\
VGGT-$\Omega$      & 0.9460 \\
4DGS               & 0.9408 \\
4C4D               & 0.9446 \\
\midrule
\textit{RoboDreamer} & \textit{0.8930/1} \\
\midrule
GT                 & 0.9019 \\
VGGT               & 0.8735 \\
VGGT-$\Omega$      & 0.8799 \\
4DGS               & 0.8956 \\
4C4D               & 0.9140 \\
\bottomrule
\end{tabular}
\label{tab:comparison33}
\end{table}

\begin{table}[t]
\centering
\caption{Collision safety score comparison of different prompts.}
\small
\begin{tabular}{lc}
\toprule
\textbf{Method} & \textbf{Collision Count (CS)} \\
\midrule
\textit{Wan} & \textit{0.8762/1} \\
\midrule
GT+Instruction      & 0.8333 \\
GT+Previous Prompt  & 0.8973 \\
GT+Current Prompt   & 0.8981 \\
\midrule
\textit{Cosmos} & \textit{0.9491/1} \\
\midrule
GT+Instruction      & 0.9435 \\
GT+Previous Prompt  & 0.9506 \\
GT+Current Prompt   & 0.9531 \\
\bottomrule
\end{tabular}
\label{tab:comparison34}
\end{table}

\paragraph{Impact of Reconstruction Quality.}
Collision safety is almost invariant to reconstruction quality in the RoboTwin reference. 
The SS values range only from $0.9360$ to $0.9396$ across the five substrates, with GT, VGGT, VGGT-$\Omega$, 4DGS, and 4C4D differing by at most $0.0036$. 
In particular, even the substantially degraded 4D reconstructions retain more than $99.7\%$ of the GT score. 
This behavior contrasts sharply with Object Localization and Trajectory Accuracy, where reconstruction quality produces large performance gaps. 
The result indicates that reconstruction artifacts alone do not materially alter the collision-safety baseline; instead, their downstream safety impact emerges primarily through how a world model responds to the reconstructed conditioning signal.
This distinction is important for interpreting generated rollouts. 
Although the RoboTwin substrates are nearly indistinguishable in SS, the corresponding generated scores can vary substantially. 
Wan, for example, decreases from $0.8981$ under GT conditioning to $0.7831$ under 4DGS, producing a range of $0.1150$ despite a RoboTwin substrate range of only $0.0036$. 
Thus, reconstruction quality can strongly affect executable safety indirectly through generation even when reconstruction itself has negligible influence on the collision metric. 
The effect is model dependent: the corresponding ranges are only $0.0191$ for CogVideoX and $0.0123$ for Cosmos, compared with $0.0405$ for RoboDreamer. 
Wan is therefore considerably more sensitive to degradation of the conditioning substrate from a safety perspective.

\paragraph{Comparison of Different World Models.}
Collision safety reveals a model ranking that differs markedly from several state- and task-level metrics. 
CogVideoX achieves the highest mean CS at $0.9650$, exceeding the RoboTwin reference average ($0.9385$) by $2.8\%$. 
Cosmos follows at $0.9458$ ($+0.8\%$), while RoboDreamer and Wan achieve $0.8930$ and $0.8314$, corresponding to $95.2\%$ and $88.6\%$ of the RoboTwin score, respectively. 
CogVideoX exceeds the corresponding RoboTwin score under all five reconstruction conditions, while Cosmos also remains slightly above RoboTwin across all substrates. 
In contrast, Wan and RoboDreamer generally exhibit lower collision safety than their corresponding reference conditions.
Importantly, a high collision-safety score should not be interpreted independently as evidence of superior embodied prediction. 
CogVideoX achieves the strongest CS despite substantially weaker Object Localization, Trajectory Accuracy, and Motion Quality. 
Its low Dynamic Degree and Flow Score indicate that comparatively limited generated motion may reduce opportunities for collision, thereby improving safety without necessarily completing the intended interaction accurately. 
Collision safety therefore captures an essential but distinct aspect of embodied performance: a rollout can be conservative and safe yet dynamically or task-wise incorrect. 
Conversely, Wan produces substantially stronger motion and instruction-following behavior but incurs a larger safety penalty, particularly under degraded reconstruction substrates. 
These results motivate evaluating safety jointly with trajectory, interaction, and task-completion metrics rather than optimizing collision avoidance in isolation.

\paragraph{Prompt Specification.}
Prompt specification substantially improves collision safety for Wan. 
CS increases from $0.8333$ with the instruction-only prompt to $0.8973$ with the previous prompt and $0.8981$ with the proposed current prompt, corresponding to a $7.8\%$ improvement over the instruction baseline. 
Nearly all of this gain occurs in the first refinement ($+0.0640$), while the proposed current prompt contributes a smaller additional improvement ($+0.0008$). 
This suggests that providing sufficiently detailed task context is critical for Wan's executable safety, whereas the additional view-aware refinement yields only a marginal safety benefit once the interaction has been adequately specified.
Cosmos is substantially more robust to prompt formulation, with SS increasing from $0.9435$ to $0.9506$ and $0.9531$. 
The overall gain is approximately $1.0\%$, and the proposed current prompt again achieves the highest score. 
Taken together, the results show that richer prompt conditioning improves collision safety for both models, with a substantially stronger effect on Wan. 
Because CS is grounded in collision outcomes rather than visual plausibility alone, these improvements provide complementary evidence that prompt design affects the downstream safety of generated behavior. 
Nevertheless, collision avoidance alone does not establish successful execution, and should therefore be considered jointly with localization, trajectory, instruction-following, and task-completion performance.

\subsubsection{Task Success}

\begin{table}[t]
\centering
\caption{Task success score comparison of different world models.}
\small
\begin{tabular}{lcccc}
\toprule
\textbf{Method} & \textbf{Action Planner} & \textbf{Data Engine} & \textbf{VLM-2} & \textbf{TS} \\
\midrule
\textit{RoboTwin} & & & & \textit{2.4340/3} \\
\midrule
GT                 & 0.9825 & 0.5875 & 1.0000 & 2.5700 \\
VGGT               & 0.9800 & 0.5750 & 0.9825 & 2.5375 \\
VGGT-$\Omega$      & 0.9850 & 0.6000 & 0.9875 & 2.5725 \\
4DGS               & 0.5475 & 0.3750 & 0.9875 & 1.9100 \\
4C4D               & 0.9825 & 0.6125 & 0.9850 & 2.5800 \\
\midrule
\textit{Wan} & & & & \textit{1.5630/3} \\
\midrule
GT                 & 0.5850 & 0.3450 & 0.9550 & 1.8850 \\
VGGT               & 0.4050 & 0.2475 & 0.9225 & 1.5750 \\
VGGT-$\Omega$      & 0.4325 & 0.2625 & 0.9550 & 1.6500 \\
4DGS               & 0.2500 & 0.1575 & 0.8650 & 1.2725 \\
4C4D               & 0.3325 & 0.2050 & 0.8950 & 1.4325 \\
\midrule
\textit{CogVideoX} & & & & \textit{0.4005/3} \\
\midrule
GT                 & 0.0825 & 0.0500 & 0.1675 & 0.3000 \\
VGGT               & 0.1000 & 0.0600 & 0.2450 & 0.4050 \\
VGGT-$\Omega$      & 0.1300 & 0.0775 & 0.3325 & 0.5400 \\
4DGS               & 0.0950 & 0.0575 & 0.2275 & 0.3800 \\
4C4D               & 0.0925 & 0.0550 & 0.2300 & 0.3775 \\
\midrule
\textit{Cosmos} & & & & \textit{2.0910/3} \\
\midrule
GT                 & 0.8350 & 0.4925 & 1.0000 & 2.3275 \\
VGGT               & 0.7525 & 0.4475 & 0.9625 & 2.1625 \\
VGGT-$\Omega$      & 0.7625 & 0.4550 & 0.9650 & 2.1825 \\
4DGS               & 0.4700 & 0.3000 & 0.9300 & 1.7000 \\
4C4D               & 0.7050 & 0.4300 & 0.9475 & 2.0825 \\
\midrule
\textit{RoboDreamer} & & & & \textit{0.1515/3} \\
\midrule
GT                 & 0.0250 & 0.0150 & 0.0525 & 0.0925 \\
VGGT               & 0.0325 & 0.0200 & 0.0825 & 0.1350 \\
VGGT-$\Omega$      & 0.0350 & 0.0200 & 0.0925 & 0.1475 \\
4DGS               & 0.0525 & 0.0325 & 0.1375 & 0.2225 \\
4C4D               & 0.0400 & 0.0250 & 0.0950 & 0.1600 \\
\bottomrule
\end{tabular}
\label{tab:comparison35}
\end{table}

\begin{table}[t]
\centering
\caption{Task success score comparison of different prompts.}
\small
\begin{tabular}{lcccc}
\toprule
\textbf{Method} & \textbf{Action Planner} & \textbf{Data Engine} & \textbf{VLM-2} & \textbf{TS} \\
\midrule
\textit{Wan} & & & & \textit{1.6692/3} \\
\midrule
GT+Instruction        & 0.3525 & 0.2075 & 0.9050 & 1.4650 \\
GT+Previous Prompt    & 0.5150 & 0.3050 & 0.8375 & 1.6575 \\
GT+Current Prompt     & 0.5850 & 0.3450 & 0.9550 & 1.8850 \\
\midrule
\textit{Cosmos} & & & & \textit{2.2025/3} \\
\midrule
GT+Instruction        & 0.7800 & 0.4550 & 0.9050 & 2.1400 \\
GT+Previous Prompt    & 0.8300 & 0.4725 & 0.8375 & 2.1400 \\
GT+Current Prompt     & 0.8350 & 0.4925 & 1.0000 & 2.3275 \\
\bottomrule
\end{tabular}
\label{tab:comparison36}
\end{table}

\paragraph{Impact of Reconstruction Quality.}
Task success is relatively robust to the two feed-forward reconstruction methods and 4C4D, but degrades sharply under 4DGS. 
On RoboTwin, GT achieves a TS of $2.5700$, while VGGT and VGGT-$\Omega$ obtain $2.5375$ and $2.5725$, respectively, and 4C4D reaches $2.5800$. 
These three reconstructed substrates therefore remain within approximately $1.3\%$ of GT. 
In contrast, 4DGS decreases to $1.9100$, retaining only $74.3\%$ of the GT score. 
The resulting substrate range is $0.6700$, but contracts to only $0.0425$ when 4DGS is excluded, indicating that the large reconstruction sensitivity of TS is almost entirely attributable to this representation.
The component-wise results localize this degradation. 
Action Planner decreases from $0.9825$ on GT to $0.5475$ on 4DGS, while Data Engine falls from $0.5875$ to $0.3750$. 
By comparison, VLM-2 remains nearly saturated at $0.9875$. 
Thus, the 4DGS reconstruction has a substantially larger effect on the execution-oriented components than on high-level task-success judgment. 
More broadly, the weak differences among GT, VGGT, VGGT-$\Omega$, and 4C4D indicate that moderate reconstruction artifacts can be tolerated at the task level, whereas sufficiently severe degradation produces an abrupt loss of downstream executability.

\paragraph{Comparison of Different World Models.}
Task success provides strong separation among the evaluated world models. 
Cosmos achieves the highest mean TS among generated models at $2.0910$, retaining $85.9\%$ of the RoboTwin reference average ($2.4340$). 
Wan follows at $1.5630$ ($64.2\%$), while CogVideoX and RoboDreamer decrease sharply to $0.4005$ ($16.5\%$) and $0.1515$ ($6.2\%$), respectively. 
This ranking is substantially more discriminative than several perceptual or VLM-based metrics and highlights large differences in the ability of current world models to generate rollouts that support downstream task execution.
The component profiles reveal an important discrepancy between execution-oriented and judgment-based evaluation. 
Averaged across substrates, Wan achieves Action Planner and Data Engine scores of only $0.4010$ and $0.2435$, whereas its mean VLM-2 score remains high at $0.9185$. 
Cosmos performs substantially better on the execution-oriented components ($0.7050$ and $0.4250$) while also maintaining a high VLM-2 score ($0.9610$). 
CogVideoX and RoboDreamer, in contrast, obtain very low Action Planner and Data Engine performance despite some nonzero VLM-2 assessment. 
These results demonstrate that a rollout can appear to satisfy the task at a semantic level without providing sufficiently accurate state evolution to support successful downstream planning or data generation. 
VLM-based task assessment should therefore complement, rather than replace, execution-oriented evaluation.
Reconstruction sensitivity after generation is likewise model dependent. 
Wan's TS decreases from $1.8850$ under GT conditioning to $1.2725$ under 4DGS, a $32.5\%$ reduction, while Cosmos decreases from $2.3275$ to $1.7000$ ($27.0\%$). 
By contrast, the already weak CogVideoX and RoboDreamer scores exhibit less meaningful correspondence with reconstruction fidelity. 
The persistent degradation on 4DGS for the stronger models indicates that reconstruction errors that appear moderate under perceptual metrics can propagate into substantial downstream task-level failures.

\paragraph{Prompt Specification.}
Prompt specification produces a substantial improvement in task success, particularly for Wan. 
Its TS increases monotonically from $1.4650$ with the instruction-only prompt to $1.6575$ with the previous prompt and $1.8850$ with the proposed current prompt, yielding an overall gain of $0.4200$ ($28.7\%$). 
Both execution-oriented components improve strongly: Action Planner increases from $0.3525$ to $0.5850$ ($+66.0\%$), while Data Engine rises from $0.2075$ to $0.3450$ ($+66.3\%$). 
VLM-2 is less consistent, decreasing from $0.9050$ to $0.8375$ under the intermediate prompt before increasing to $0.9550$ with the current prompt. 
Consequently, the monotonic improvement in aggregate TS is driven primarily by the execution-oriented components, with the proposed current prompt ultimately achieving the strongest result across all three components.
Cosmos exhibits a more stable and higher-performing prompt profile. 
TS remains at $2.1400$ for the instruction and previous prompts because improvements in Action Planner and Data Engine under the previous prompt are offset by a decrease in VLM-2. 
With the proposed current prompt, however, TS increases substantially to $2.3275$, corresponding to an $8.8\%$ improvement over the instruction baseline. 
Importantly, all three components attain their highest values under the current prompt: Action Planner increases from $0.7800$ to $0.8350$, Data Engine from $0.4550$ to $0.4925$, and VLM-2 from $0.9050$ to the maximum score of $1.0000$. 
This simultaneous improvement indicates that the proposed conditioning not only enhances downstream planning and data-generation utility, but also yields a rollout that is fully recognized by the VLM-based evaluator as satisfying the intended task.
Taken together, the prompt results provide strong evidence that richer conditioning improves task-level utility. 
For Wan, the gain is particularly large and is driven primarily by Action Planner and Data Engine, whereas Cosmos starts from a substantially stronger baseline and achieves a smaller but more comprehensive improvement across all three components. 
Unlike purely perceptual metrics, Action Planner and Data Engine directly assess whether information encoded in the generated rollout remains useful for downstream embodied decision-making. 
The consistent gains under the proposed current prompt therefore indicate that improved prompt specification affects not only the visual or semantic quality of generation, but also its practical utility for task execution.

\end{document}

%% file: sec/0_abstract.tex
\begin{abstract}

Video world models increasingly serve as data engines, action planners, and simulators for embodied AI, but conventional embodied world model (EWM) benchmarks lack a unified 3D-grounded protocol for establishing whether generated rollouts preserve the underlying 3D scene state or translate into executable actions. 
We introduce RoboPhys-3D, a 3D-grounded EWM benchmark built on RoboTwin 2.0, covering 50 manipulation tasks across four regimes, with 5,000 episodes and 25,000 multi-view ground-truth videos.
A defining feature of RoboPhys-3D is that generated and ground-truth videos are processed through the same 3D reconstruction pipeline, enabling reconstruction-induced error to be distinguished from generation-induced error. 
The RoboPhys-3D benchmark organizes 50 complementary metrics into 18 sub-dimensions across four levels: pixel-level fidelity, 3D geometry consistency, state-level understanding, and task-level completeness. 
We further introduce Average Full Score, a hierarchical score averaging all 50 metrics for comprehensive evaluation, and RoboPhyscore, a compact task-aligned score averaging the metrics most strongly correlated with task success. 
Among the four representative video world models, Cosmos 3 achieves the highest RoboPhyscore ($0.6330$, 92.7\% of ground truth), while state- and execution-grounded metrics reveal substantial failures that perceptual and vision-language model-based judgments fail to capture. 
RoboPhyscore further exhibits strong agreement with human evaluation (Pearson $r=0.9761$ and Spearman $\rho=0.8962$), demonstrating the importance of grounded, execution-aware evaluation for EWM capability.

\end{abstract}

%% file: sec/1_introduction.tex
\section{Introduction}
\label{sec:introduction}

\begin{table*}[ht]
\caption{Comparison of the RoboPhys-3D benchmark (ours) to other existing embodied world model (EWM) benchmarks. \textbf{(1) Pixel-level fidelity:} IQ-1 = Image Quality, AQ = Aesthetic Quality, DS = Distribution Similarity, MQ = Motion Quality, CC = Content Consistency. \textbf{(2) 3D geometry consistency:} RBG = Reconstruction-Based Geometry, DC = Depth Consistency, CTG = Camera Trajectory Geometry, RC = Roundtrip Consistency. \textbf{(3) State-level understanding:} OL = Object Localization, TA = Trajectory Accuracy, IQ-2 = Interaction Quality, EAA = Event and Action Adherence, PLA = Physical Law Adherence, SA = Semantic Adherence. \textbf{(4) Task-level completeness:} IF = Instruction Following, CS = Collision Safety, TS = Task Success.}
\label{tab1} 
\centering
\small
\resizebox{\linewidth}{!}{%
\begin{tabular}{lcccccccccc}
\toprule
\textbf{Metrics} & \textbf{WBench \cite{ying2026wbench}} & \textbf{ESI-Bench \cite{hong2026esi}} & \textbf{WorldArena \cite{shang2026worldarena}} & \textbf{RoboWM-Bench \cite{jiang2026robowm}} & \textbf{EZSBench \cite{chen2026abot}} & \textbf{Wow,wo,val \cite{fan2026wow}} & \textbf{WorldModelBench \cite{li2026worldmodelbench}} & \textbf{EWMBench \cite{yue2025ewmbench}} & \textbf{World-in-World \cite{zhang2025worldinworld}} & \textbf{Ours}  \\  
\midrule
\textit{Venue \& Year} & arXiv 26' & arXiv 26' & CVPRW 26' & CVPRW 26' & arXiv 26' & arXiv 26' & NeurIPS 25' & arXiv 25' & ICLR 26' & / \\
\midrule
\multicolumn{5}{l}{\textit{Pixel-Level Fidelity}} \\
\midrule
IQ-1 & \checkmark & \texttimes & \checkmark & \texttimes & \checkmark & \checkmark & \checkmark & \texttimes & \checkmark & \checkmark \\ 
AQ & \checkmark & \texttimes & \checkmark & \texttimes & \checkmark & \texttimes & \texttimes & \texttimes & \checkmark & \checkmark \\   
DS & \texttimes & \texttimes & \checkmark & \texttimes & \texttimes & \checkmark & \texttimes & \texttimes & \texttimes & \checkmark \\
MQ & \checkmark & \texttimes & \checkmark & \texttimes & \checkmark & \checkmark & \checkmark & \checkmark & \texttimes & \checkmark \\
CC & \checkmark & \texttimes & \checkmark & \texttimes & \checkmark & \checkmark & \texttimes & \checkmark & \texttimes & \checkmark \\
\midrule
\multicolumn{4}{l}{\textit{3D Geometry Consistency}}\\
\midrule
RBG & \checkmark & \texttimes & \texttimes & \texttimes & \texttimes & \texttimes & \texttimes & \texttimes & \texttimes & \checkmark \\
DC & \checkmark & \texttimes & \checkmark & \texttimes & \texttimes & \texttimes & \texttimes & \texttimes & \texttimes & \checkmark \\
CTG & \checkmark & \texttimes & \texttimes & \texttimes & \texttimes & \texttimes & \texttimes & \texttimes & \texttimes & \checkmark \\
RC & \checkmark & \texttimes & \texttimes & \texttimes & \texttimes & \texttimes & \texttimes & \texttimes & \texttimes & \checkmark \\
\midrule
\multicolumn{6}{l}{\textit{State-Level Understanding}}\\
\midrule
OL & \texttimes & \checkmark & \texttimes & \texttimes & \texttimes & \texttimes & \texttimes & \texttimes & \texttimes & \checkmark \\
TA & \texttimes & \texttimes & \checkmark & \texttimes & \checkmark & \checkmark & \texttimes & \checkmark & \texttimes & \checkmark \\
IQ-2 & \texttimes & \texttimes & \checkmark & \checkmark & \checkmark & \checkmark & \texttimes & \texttimes & \texttimes & \checkmark  \\
EAA & \checkmark & \texttimes & \checkmark & \checkmark & \texttimes & \checkmark & \checkmark & \texttimes & \texttimes & \checkmark \\
PLA & \checkmark & \texttimes & \checkmark & \texttimes & \checkmark & \checkmark & \checkmark & \texttimes & \texttimes & \checkmark \\
SA & \checkmark & \texttimes & \checkmark & \texttimes & \texttimes & \checkmark & \texttimes & \checkmark & \texttimes & \checkmark \\
\midrule
\multicolumn{2}{l}{\textit{Task-Level Completeness}}\\
\midrule
IF & \checkmark & \texttimes & \checkmark & \texttimes & \texttimes & \checkmark & \checkmark & \texttimes & \checkmark & \checkmark \\
CS & \texttimes & \texttimes & \texttimes & \texttimes & \texttimes & \texttimes & \texttimes & \texttimes & \texttimes & \checkmark \\
TS & \texttimes & \checkmark & \checkmark & \checkmark & \texttimes & \checkmark & \texttimes & \texttimes & \checkmark & \checkmark \\
\midrule
\textit{Unified Score} & \checkmark & \texttimes & \checkmark & \texttimes & \checkmark & \checkmark & \checkmark & \checkmark & \texttimes & \checkmark \\
\midrule
\textit{\# of Metrics} & 22 & 1 & 16 & 2 & 14 & 22 & 8 & 6 & 7 & \textbf{50} \\
\bottomrule
\end{tabular}}
\end{table*}

World models are generative models that learn the dynamics of an environment, including its physical and spatial properties, from observation \cite{ha2018recurrent}, and they have shown strong promise for the highly dynamic, safety-critical conditions of embodied tasks \cite{long2025survey}.
In contrast to general world models (GWMs), embodied world models (EWMs) are tied directly to physical action and serve three complementary roles: a data engine that synthesizes interaction trajectories at scale, an action planner that forecasts future states from the current observation, action, and instruction, and a simulator that trains and evaluates control policies \cite{shang2026survey,shang2026worldarena,shang2026worldarena2}. 
Unlike a physics engine, whose dynamics are explicitly specified, an EWM acquire its dynamics implicitly from data, and consequently, there is no guarantee that a generated rollout follows the geometry and physics of the scene it depicts \cite{yu2025survey}. 
Therefore, standardized measurement of the physical and geometric fidelity of generated rollouts represents a critical prerequisite for deploying EWMs in embodied AI.

Existing benchmarks have made important progress, but they have not provided a unified 3D-grounded evaluation protocol for EWMs. 
General video generation benchmarks \cite{huang2024vbench,huang2025vbench++,zheng2025vbench,zhou2026pai,zhang2026physion,motamed2026generative,gu2025phyworldbench,meng2025towards} mainly assess perceptual quality and physical plausibility from the generated frames alone, and do not assess agreement with a ground-truth 3D scene state or the executability of the depicted motion. 
Prior studies \cite{fan2026wow,zhang2025worldinworld} show that a model can score highly on conventional video metrics while producing physically impossible or contextually incorrect predictions, resulting in poor downstream task success. 
GWM benchmarks \cite{xu2026worldmark,ye2026mind,wu2026omni,lu20264dworldbench,duan2025worldscore} emphasize camera-trajectory control and viewpoint synthesis rather than the scene-level geometric consistency on which reasoning and decision-making in robotic manipulation depend. 
Current EWM benchmarks \cite{ying2026wbench,hong2026esi,shang2026worldarena,jiang2026robowm,chen2026abot,fan2026wow,li2026worldmodelbench,yue2025ewmbench,zhang2025worldinworld} broaden the range of evaluated dimensions but remain limited in scope, as summarized in Table \ref{tab1}. 
In addition, their metrics follow inconsistent taxonomies \cite{li2025comprehensive,kong20253d,zhang2025step}, hindering fair cross-benchmark comparison.
A further gap concerns how physical consistency itself is measured.
WBench \cite{ying2026wbench} scores reconstruction-based, depth, camera-trajectory, and roundtrip consistency, but quantifies only the internal 3D self-consistency of a generated clip rather than its agreement with a reference scene.
ESI-Bench \cite{hong2026esi} uses a reconstructed 3D structure as input to a downstream reasoner rather than as an object of measurement.
Consequently, no prior benchmark provides a unified evaluation protocol that compares the physical consistency of generated videos against the corresponding ground-truth motion within the same reconstructed 3D scene \cite{mei2026video}.

\begin{figure*}[htbp!]
  \centering
  \includegraphics[width=\linewidth]{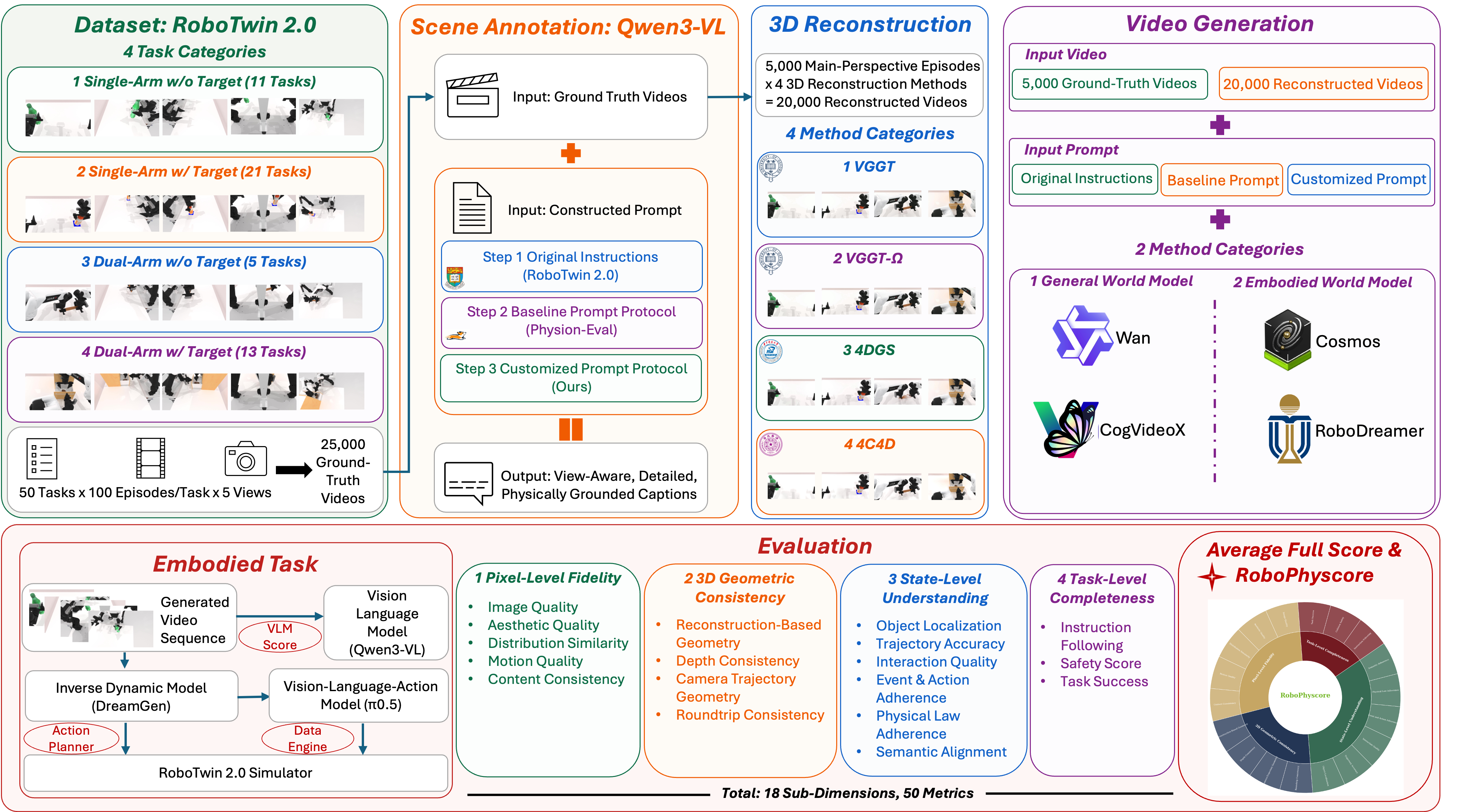}
  \caption{Overview of the proposed RoboPhys-3D benchmark. 
  }
  \label{fig1}
\end{figure*}

To address this gap, we introduce \textbf{RoboPhys-3D} (Figure \ref{fig1}), a 3D-grounded benchmark for EWMs. 
Built on RoboTwin 2.0 \cite{chen2025robotwin}, \textbf{RoboPhys-3D} covers four robotic manipulation regimes: single- and dual-arm tasks, each with and without a target location, comprising 50 tasks with 100 episodes per task rendered from five camera views. 
For each episode, we construct a view-aware natural-language prompt via Qwen3-VL \cite{bai2025qwen3}, and reconstruct the main-perspective video using four 3D reconstruction methods \cite{wang2025vggt,wang2026vggt,wu20244d,zhou20264c4d}.
Each candidate model autoregressively generates the future frames depicting the manipulation, which are reconstructed and decoded by an inverse dynamics model (IDM) \cite{jang2025dreamgen} into executable robot actions replayed in the simulator.
The defining feature of the protocol is that each generated rollout is evaluated against its reconstruction-matched ground truth, rather than against the raw ground truth, thereby isolating generation-induced error from reconstruction-induced error, yielding a fair, unified 3D-level EWM evaluation that no prior benchmark provides. 
Beyond the evaluation protocol, \textbf{RoboPhys-3D} introduces two complementary scores: \textit{Average Full Score (AFS)} hierarchically consolidates all 50 normalized metrics across 18 sub-dimensions and four evaluation levels into a comprehensive diagnostic summary; and \textit{RoboPhyscore} is a compact, task-aligned indicator of EWM's capability that averages the non-task-success metrics exhibiting strong Pearson correlation with task-success measurements.

The main contributions are summarized as follows:
\begin{itemize}
\item \textbf{A 3D-Grounded Embodied World Model Benchmark.} We introduce \textit{RoboPhys-3D}, the first benchmark to evaluate EWMs by comparing generated and ground-truth videos within a shared reconstructed 3D scene.
\item \textbf{A Multi-Level Evaluation Framework.} We establish a unified evaluation taxonomy spanning 50 complementary metrics organized into 18 sub-dimensions and four levels, which jointly characterize visual, geometric, physical, semantic, and execution-level behavior. 
\item \textbf{Two Complementary Scores.} We introduce \textit{AFS} for a comprehensive diagnostic summary, and \textit{RoboPhyscore} for a compact, task-relevant assessment validated against human judgment, enabling systematic analysis of how reconstruction and generation errors propagate to downstream embodied performance.
\end{itemize}

%% file: sec/2_relatedwork.tex
\section{Related Work}
\label{sec:relatedwork}

\subsection{3D \& 4D Reconstruction}

\textbf{Method.}
Recovering 3D structure and camera motion from images remains a fundamental problem in computer vision, classically addressed by structure-from-motion (SfM) \cite{pan2024glomap} and multi-view stereo (MVS) \cite{jiang2026mvsmamba}.
Neural radiance fields (NeRF) \cite{mildenhall2021nerf} achieve high-quality novel-view synthesis by volume-rendering a learned implicit field, while 3D Gaussian Splatting (3DGS) \cite{kerbl20233d} rasterizes explicit Gaussian primitives to improve rendering efficiency.
To eliminate per-scene optimization and the reliance on known camera poses, feed-forward methods \cite{wang2025vggt,wang2026vggt} directly regress geometry and camera parameters from uncalibrated images.
Extending reconstruction to dynamic content, 4D Gaussian Splatting (4DGS) \cite{wu20244d} and subsequent 4D representations \cite{zhou20264c4d} model time-varying geometry for deformable and moving scenes. 

\textbf{Evaluation.}
Reconstruction quality is conventionally assessed by rendering fidelity and geometric fidelity \cite{jia2026high}. 
Rendering fidelity compares rendered novel views against held-out images using PSNR \cite{pratt2007digital}, SSIM \cite{wang2004image}, and LPIPS \cite{zhang2018unreasonable}, while geometric fidelity is evaluated against reference geometry using Chamfer distance \cite{fan2017point}, per-frame depth error \cite{shang2026worldarena}, and camera-pose accuracy \cite{li2025megasam}. 
Beyond assessing reconstruction in isolation, recent work purposes it as a probe of a video world model's geometric coherence \cite{mei2026video}, testing whether the generated frames remain physically consistent.
RoboPhys-3D builds on this premise and extends it from open-domain video to robotic manipulation, where the reconstructed scene is not only measurable but actionable.

\subsection{Video World Model}

\textbf{Method.}
Video generation has evolved rapidly from early U-Net-based diffusion models \cite{ho2022video} to scalable diffusion Transformers (DiT) trained with flow-matching objectives on large-scale data \cite{yang2025cogvideox}, yielding longer, higher-resolution, and temporally coherent outputs. 
Video world models, which predict how a scene evolves under actions or instructions, enable interactive, controllable, and action-conditioned generation \cite{wu2026video}. 
EWMs, such as Cosmos \cite{agarwal2025cosmos}, RoboDreamer \cite{zhou2024robodreamer}, and EnerVerse \cite{huang2026enerverse}, specialize this capability for manipulation, generating future rollouts from an initial frame together with a language or action input, trained or fine-tuned on robot data.
Then, their predictions are intended to support planning, data synthesis, and policy learning.

\textbf{Evaluation.}
General video generation benchmarks \cite{huang2024vbench,zhou2026pai,gu2025phyworldbench,motamed2026generative} decompose visual quality into fine-grained dimensions and probe physical commonsense through curated prompts and video question answering (VQA), but they neither accept action inputs nor evaluate multi-turn interaction.
GWM benchmarks \cite{duan2025worldscore,lu20264dworldbench,wu2026omni} emphasize camera-trajectory control and viewpoint synthesis rather than scene-level geometric agreement with a manipulation environment. 
Recent EWM benchmarks instead transfer evaluation to embodied settings. 
For example, WorldModelBench \cite{li2026worldmodelbench}, EZSBench \cite{chen2026abot}, WBench \cite{ying2026wbench} and EWMBench \cite{yue2025ewmbench} evaluate core dimensions such as perception, prediction, and generalization, but not planning and execution.
WBench \cite{ying2026wbench}, the only prior benchmark to score reconstruction-based, depth, camera-trajectory, and roundtrip consistency, quantifies the internal 3D self-consistency of open-domain interactive video rather than its agreement with a reference manipulation scene.
In contrast, WoW-World-Eval~\cite{fan2026wow}, WorldArena~\cite{shang2026worldarena}, RoboWM-Bench~\cite{jiang2026robowm} and World-in-World~\cite{zhang2025worldinworld} close the loop by decoding generated videos into robot actions, further unifying perception metrics with functional utility.
Besides, ESI-Bench~\cite{hong2026esi} targets embodied spatial reasoning through a perception-action loop.
These efforts substantially advance EWM evaluation, but two limitations persist: their metrics are organized under inconsistent taxonomies that impede fair comparison, and none grounds physical-consistency assessment in the reconstruction of the generated video relative to the corresponding ground-truth motion in the same 3D scene. 
A detailed comparison with existing EWM benchmarks is provided in Table \ref{tab1}.

%% file: sec/3_dataset.tex
\section{RoboPhys-3D Dataset}
\label{sec:dataset}

The RoboPhys-3D dataset is constructed through a four-stage pipeline (Figure \ref{fig1}): (1) collecting multi-view ground-truth videos of robotic manipulation in simulation across four task categories, (2) synthesizing view-aware natural-language prompts and exporting per-episode scene annotations, (3) lifting every main-perspective ground-truth and generated rollout into 3D with four reconstruction methods, and (4) conditioning candidate video world models on an initial frame and the corresponding prompt.
In total, RoboPhys-3D comprises 50 tasks and 145,000 videos for final evaluation (5,000 ground-truth videos, 20,000 reconstructed videos, and 120,000 generated videos). 
Full dataset statistics are given in Tables \ref{tab2} and \ref{tab3}, and configurations of the reconstruction methods, video world models, and IDMs are listed in Tables \ref{tab4}--\ref{tab6} of Appendix \ref{appendix:dataset}, respectively.

\subsection{Problem Formulation}

We treat a candidate video world model $g_\theta$ as a conditional video generator that predicts how a manipulation scene evolves under a control specification \cite{fan2026wow}. 
Let $o_t\in\mathcal{O}$ denote the RGB observation at frame $t$ and $o_{\le t}=\{o_0,\dots,o_t\}$ the observed history. 
Given a conditioning signal $\mathcal{C}$, the model factorizes the next-frame distribution as $o_{t+1}\sim g_\theta\{o_{t+1}\mid o_{\le t},\mathcal{C}\}$, and is unrolled autoregressively over a horizon of $H$ frames to yield a generated rollout $\hat{\mathbf{X}}=\{\hat{o}_1,\dots,\hat{o}_H\}$.
Each RoboPhys-3D instance is defined as:
\begin{equation}
\Omega=\{\mathcal{S}, c, p, \mathbf{A}^{\mathrm{gt}}, \mathbf{X}^{\mathrm{gt}}, \Phi^{\mathrm{gt}}\}, 
\end{equation}
where $\mathcal{S}$ is a 3D manipulation scene instantiated in RoboTwin 2.0 \cite{chen2025robotwin}, $c$ is the initial frame that fixes the embodiment and environment, $p$ is the view-aware language prompt produced by Qwen3-VL \cite{bai2025qwen3}, $\mathbf{A}^{\mathrm{gt}}=\{a_1,\dots,a_H\}$ is the ground-truth action sequence, in which each $a_t\in\mathbb{R}^{d}$ encodes end-effector position, orientation, and gripper state, with $d{=}7$ for single-arm and $d{=}14$ for dual-arm regimes following the action parameterization of Abot-physworld \cite{chen2026abot}, $\mathbf{X}^{\mathrm{gt}}=\{\mathbf{X}^{\mathrm{gt}}_v\}_{v=1}^{V}$ are the multi-view ground-truth videos with $V{=}5$, and $\Phi^{\mathrm{gt}}$ is the dense privileged scene state exported from the simulator.

Given the action conditioning $\mathcal{C}=\{c,p\}$, each model applies its own preprocessing $f_{\mathrm{proc}}$, and all outputs pass through a common normalizer $v_{\mathrm{norm}}$ that matches resolution, frame count, and frame rate before scoring $\hat{\mathbf{X}} = v_{\mathrm{norm}}\{g_\theta\{f_{\mathrm{proc}}\{\mathcal{C}\}\}\}$, following the protocol of EWMBench \cite{yue2025ewmbench}.
A reconstruction operator $\mathcal{R}$ then lifts a video into a shared 3D representation $\hat{G}=\mathcal{R}\{\hat{\mathbf{X}}\}$ and returns its main-perspective rendering.
Then, an IDM \cite{jang2025dreamgen} $\pi_\phi$ decodes the generated rollout into an executable action sequence $\hat{\mathbf{A}}=\pi_\phi\{\hat{\mathbf{X}}\}$, which is compared against $\mathbf{A}^{\mathrm{gt}}$ in action space and decoded in the 3D manipulation scene $\mathcal{S}$ to obtain a task-completion outcome. 
Finally, RoboPhys-3D scores each instance with a family of metric functions.

\subsection{Ground-Truth Data Collection}

We build the RoboPhys-3D dataset on the RoboTwin 2.0 simulation platform \cite{chen2025robotwin}, which provides diverse manipulation environments together with privileged object, robot, and camera states.
We retain all 50 RoboTwin 2.0 tasks and record 100 episodes per task, each captured from five time-synchronized camera views, containing 5,000 episodes and 25,000 ground-truth videos. 
Tasks are organized along two orthogonal axes that stress different physical capabilities: the number of active manipulators (single-arm vs. dual-arm), which controls kinematic and coordination complexity, and whether the task specifies a designated target location or receptacle for the manipulated object (with vs. without target), which controls whether success depends on goal-directed placement or on the manipulation itself.

\subsection{Prompt Construction and Scene Annotation}

The native RoboTwin 2.0 instructions \cite{chen2025robotwin} specify task goals but often omit visually important motion details such as the acting gripper, approach direction, and manipulation trajectory.
Following Physion-Eval \cite{zhang2026physion}, we design prompts that foreground explicit, visually observable descriptions of physical behavior rather than abstract statements of physical law. 
Therefore, we prompt Qwen3-VL \cite{bai2025qwen3} with the instruction and the main-perspective ground-truth video to construct a detailed, view-aware description for each episode. 
The caption further separates static appearance, which is already supplied by the conditioning frame, from the temporal dynamics that the model must synthesize.
A manual review pass over all generated captions discards those containing hallucinated or physically ambiguous content.
Each episode is annotated at three levels of textual conditioning: the original RoboTwin instruction \cite{chen2025robotwin}, the Physion-Eval \cite{zhang2026physion} caption, and our view-aware caption, which enables a controlled study of how increasing language specificity affects video generation.
The full caption-generation prompts are provided in Appendix \ref{appendix:dataset}.

\subsection{3D Reconstruction}

Each main-perspective video is reconstructed by four methods: the pose-free feed-forward estimators VGGT \cite{wang2025vggt} and VGGT-$\Omega$ \cite{wang2026vggt}, which regress geometry and camera parameters directly from frames, and the dynamic 4D representations 4DGS \cite{wu20244d} and 4C4D \cite{zhou20264c4d}, which optimize time-varying geometry. 
Applying the identical pipeline to every video yields, for every episode, four reconstructed representations with the raw video for scoring.
Scoring across independent reconstruction methods prevents the benchmark from rewarding a model whose failure modes happen to be invisible to a single method, and the disagreement among methods indicates which artifacts dominate the evaluation.
Moreover, the residual discrepancy measured on ground truth calibrates the error intrinsic to reconstruction itself, and this score helps distinguish generation-induced error from reconstruction-induced error.
This calibration allows RoboPhys-3D to make 3D-level claims that are not confounded by reconstruction artifacts, the failure mode measured by ESI-Bench \cite{hong2026esi}.

\subsection{Video Generation}

We evaluate four representative open-source video world models from two families: the GWMs Wan 2.2 \cite{wan2025wan} and CogVideoX 1.5 \cite{yang2025cogvideox}; and the EWMs such as Cosmos 3 \cite{agarwal2025cosmos} and RoboDreamer \cite{zhou2024robodreamer}. 
All models are post-trained on the RoboPhys-3D dataset following their official implementations, and all runs use eight A100 80GB GPUs. 
Episodes are split sequence-wise into training, validation, and test sets at a 70\%/15\%/15\% ratio.
Each model predicts the future manipulation frames from the initial observation and the prompt under the matched output settings. 
Each generated video is then reconstructed with the same backends for geometric scoring and decoded by the IDM \cite{jang2025dreamgen} into executable actions that are replayed in the simulator scene.

%% file: sec/4_evaluation.tex
\section{RoboPhys-3D Evaluation}
\label{sec:evaluation}

The RoboPhys-3D evaluation framework consists of four levels (Figure \ref{fig1}): (1) pixel-level fidelity, which assesses the perceptual reliability of generated frames; (2) 3D geometric consistency, which evaluates whether the generated video corresponds to a coherent 3D scene that agrees with the ground-truth scene; (3) state-level understanding, which measures whether the structured state of the scene matches the ground truth; and (4) task-level completeness, which evaluates whether the generated video, once decoded into executable actions, completes the manipulation task. 
Most metrics are computed automatically from open-source components, except for embodied task success, and per-metric implementation details are provided in Appendix \ref{appendix:evaluation}.

\subsection{Four-Level Metric Taxonomy}

\textbf{Pixel-level fidelity} spans five sub-dimensions and 20 metrics. 
\textit{(a) Image quality} combines single-image no-reference quality (MUSIQ \cite{ke2021musiq}), frame-paired fidelity (PSNR \cite{pratt2007digital}, SSIM \cite{wang2004image}, LPIPS \cite{zhang2018unreasonable}), distribution-level distance to the reference set (FVD \cite{unterthiner2018towards}), and reference-free distribution statistics (Inception Score \cite{salimans2016improved}), so that the same frames are judged under four distinct notions of quality. 
\textit{(b) Aesthetic quality} applies four learned human-preference predictors: LAION-AP \cite{schuhmann2022laion}, HPSv3 \cite{ma2025hpsv3},  CLIP-AP \cite{schuhmann2022clip+}, and CLIP-IQA \cite{wang2023exploring}.
\textit{(c) Distribution similarity} measures the maximum mean discrepancy between V-JEPA feature distributions of generated and reference videos \cite{luo2024beyond}, relaxing the Gaussian assumption underlying FVD. 
\textit{(d) Motion quality} quantifies the intensity \cite{huang2024vbench}, magnitude \cite{liu2024evalcrafter}, temporal coherence \cite{duan2025worldscore}, and high-frequency flicker \cite{huang2024vbench} of motion from RAFT optical flow \cite{teed2020raft} and a frame-interpolation prior \cite{zhang2024vfimamba}. 
\textit{(e) Content consistency} measures the temporal stability of subject \cite{ying2026wbench}, background \cite{huang2024vbench}, photometry \cite{duan2025worldscore}, semantics \cite{huang2024vbench}, and rendering style \cite{duan2025worldscore} using SAM 2 \cite{ravi2025sam}, DINOv2 \cite{oquab2023dinov2}, CLIP \cite{radford2021learning}, and ViCLIP \cite{wang2024internvid} features.

\textbf{3D geometric consistency} spans four sub-dimensions and eight metrics. 
\textit{(a) Reconstruction-based geometry} reconstructs the rollout with COLMAP SfM \cite{schonberger2016structure} and 3DGS \cite{kerbl20233d} and reports the recovered primitive count, the keep-ratio of correct matches under increasing frame intervals \cite{li2024sora}, and the geometric and photometric residuals of depth-based reprojection~\cite{ying2026wbench}, since reconstruction succeeds only where the frames are mutually multi-view consistent.
\textit{(b) Depth consistency} pairs a grounded comparison of monocular depth \cite{lin2025depth} against the simulator's depth with a VLM judgment of perspective plausibility \cite{shang2026worldarena}. 
\textit{(c) Camera-trajectory geometry} scores the camera path recovered by DROID-SLAM \cite{teed2021droid}, because fixed-view episodes measures hallucinated drift. 
\textit{(d) Roundtrip consistency} locates the frame whose viewpoint best matches the first \cite{li2025megasam} and scores the perceptual similarity to the first frame \cite{fu2023dreamsim}, capturing the out-of-sight persistence.

\textbf{State-level understanding} spans six sub-dimensions and 16 metrics. 
\textit{(a) Object localization} reports detection precision and recall for the manipulated objects \cite{lin2014microsoft}. 
\textit{(b) Trajectory accuracy} compares object and end-effector paths with the ground-truth paths in shape, timing, and dynamics via HSD \cite{serra1998hausdorff}, nDTW \cite{muller2007dynamic}, DYN \cite{yue2025ewmbench}, and decision-step count \cite{wang2025roboeval}. 
\textit{(c) Interaction quality} \cite{shang2026worldarena} assesses contact, force transmission, interpenetration, and grasp correctness with Qwen3-VL \cite{bai2025qwen3}. 
\textit{(d) Event and action adherence} applies five binary checks to the instructed events and subject's actions \cite{ying2026wbench}.
\textit{(e) Physical law adherence} uses a VLM to judge compliance with six physical laws \cite{bansal2025videophy}. 
\textit{(f) Semantic adherence} scores caption-level agreement with the instruction via BLEU \cite{papineni2002bleu}, key-step agreement with the descriptions via CLIP similarity \cite{schuhmann2022clip+}, and four VLM rubrics covering logical consistency \cite{yue2025ewmbench}, instruction execution \cite{shang2026worldarena}, and subject and scene adherence \cite{ying2026wbench}.

\textbf{Task-level completeness} spans three sub-dimensions and six metrics. 
\textit{(a) Instruction following} reports two VLM judgments of whether the depicted behavior achieves the instructed final state \cite{shang2026worldarena,li2026worldmodelbench}. 
\textit{(b) Collision safety} counts unintended robot-environment and robot-self collisions during execution \cite{wang2025roboeval}. 
\textit{(c) Task success} reports the success rate of directly replaying the decoded actions, the success rate of a $\pi_{0.5}$ policy \cite{intelligence2025pi} trained on the rollouts as synthetic data \cite{hu2025video}, and a VLM judgment of task completion \cite{shang2026worldarena}, so that a world model is assessed as an action planner, as a data engine, and by observed task completion.
An IDM \cite{jang2025dreamgen} decodes each rollout into joint-space actions replayed in the RoboTwin 2.0 simulator \cite{chen2025robotwin} within the same 3D scene. 

\begin{figure*}[t]
  \centering
  \includegraphics[width=\linewidth]{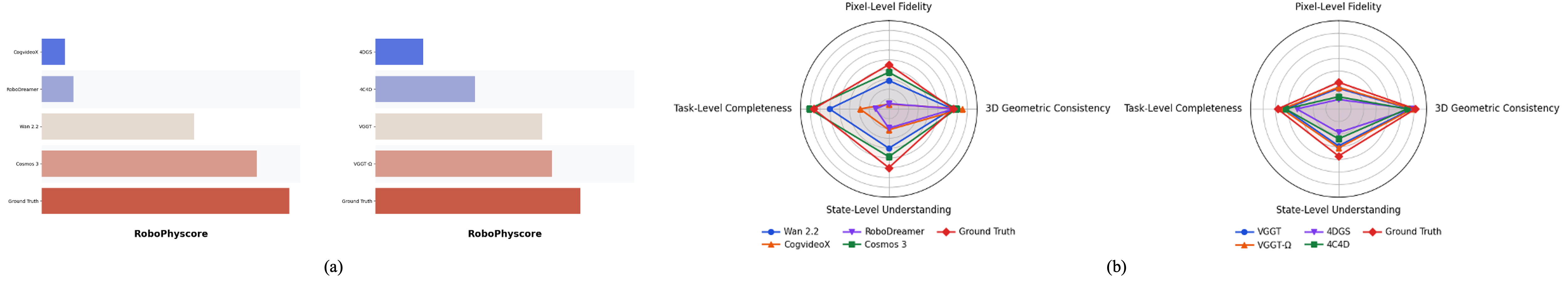}
  \caption{Overall results of Average Full Score (AFS) and RoboPhyscore. (a) RoboPhyscore for different video world models and reconstruction methods. (b) AFS for different video world models and reconstruction methods.}
  \label{fig2}
\end{figure*}

\subsection{Average Full Score}

To provide a comprehensive summary of the benchmark, \textit{AFS} hierarchically aggregates all 50 normalized metrics according to the RoboPhys-3D taxonomy:
\begin{equation}
\mathrm{AFS} = \sum_{\ell\in\mathcal{L}} \underbrace{\frac{1}{|D_\ell|}\sum_{d\in D_\ell} \underbrace{\frac{1}{|M_d|}\sum_{m\in M_d}\tilde{s}_m}_{\displaystyle S_d}}_{\displaystyle S_\ell},
\label{eq:afs}
\end{equation}
where $\mathcal{L}=\{1,\dots,4\}$ indexes the four levels, $D_\ell$ is the set of sub-dimensions belonging to level $\ell$, $M_d$ is the set of metrics belonging to sub-dimension $d$, $S_d$ is the sub-dimension score, and $S_\ell$ is the level score. 
Averaging within $M_d$ makes a sub-dimension's score independent of how many metrics instantiate it, so that image quality with six metrics does not outweigh collision safety with one metric. 
Averaging within $D_\ell$ makes a level's score independent of how finely it is subdivided. 
Averaging over $\mathcal{L}$ fixes each level's contribution at exactly 25\%, so the relative importance of appearance, geometry, state, and execution is an explicit design choice.
Scores are computed within each task regime and macro-averaged with equal weight across the four regimes.
The normalization process is described in Appendix \ref{appendix:evaluation}.

\subsection{RoboPhyscore}

\textit{RoboPhyscore} selects a task-aligned subset of metrics based on the three complementary task success measurements: Action Planner, Data Engine, and VLM-2.
For each non-task-success metric $m$, we compute its Pearson correlation \cite{benesty2009pearson} with these three task-success measurements and retain the metric only when:
\begin{equation}
\min\left\{
r(m,\mathrm{AP}),
r(m,\mathrm{DE}),
r(m,\mathrm{VLM\text{-}2})
\right\} > 0.70 ,
\label{eq:rps_selection}
\end{equation}
where $r(\cdot,\cdot)$ denotes the Pearson correlation coefficient.
This criterion identifies eight metrics that exhibit consistently strong positive associations with all three task-success measurements: JEPA Similarity (DS), Semantics Consistency (CC), Subject Adherence (SA), Normalized Score (IF), Average Recall (OL), Physical Commonsense (PLA), nDTW (TA), and FVD (IQ-1).
The parameter sensitivity analysis of the threshold is reported in Appendix \ref{appendix:evaluation}.
Let $\mathcal{M}_{\mathrm{RPS}}$ denote this set of eight selected metrics.
\textit{RoboPhyscore} is computed as the mean of the normalized selected metrics:
\begin{equation}
\mathrm{RoboPhyscore}
=
\frac{1}{|\mathcal{M}_{\mathrm{RPS}}|}
\sum_{m\in\mathcal{M}_{\mathrm{RPS}}} \tilde{s}_m.
\label{eq:rps}
\end{equation}

Overall, \textit{AFS} summarizes overall benchmark performance, and \textit{RoboPhyscore} provides a compact task-aligned measure of EWM capability.

%% file: sec/5_experiment.tex
\section{Experiments}
\label{sec:experiment}

\begin{table}[t]
\caption{Detailed quantitative results of the RoboPhys-3D benchmark for four representative video world models.}
\centering
\small
\resizebox{\linewidth}{!}{%
\begin{tabular}{lccccc}
\toprule
\textbf{Metrics} & \textbf{Ground Truth} & \multicolumn{2}{c}{\textbf{General World Model}} & \multicolumn{2}{c}{\textbf{Embodied World Model}} \\  
& & \textbf{Wan} & \textbf{CogVideoX} & \textbf{Cosmos} & \textbf{RoboDreamer} \\
\midrule
\textit{Pixel-Level Fidelity (/5)}
& 3.1300 & 2.7249 & 2.1236 & 2.9329 & 2.1359 \\
\midrule
IQ-1 (/6)
& 4.3156 & 3.5108 & 3.0803 & 3.6128 & 3.2018 \\ 
AQ (/4)
& 1.3323 & 1.6047 & 1.3855 & 1.5019 & 1.0521 \\   
DS (/1)
& 0.8142 & 0.3602 & 0.1505 & 0.6525 & 0.0898 \\
MQ (/4)
& 2.5848 & 3.1620 & 2.0245 & 2.8713 & 2.6996 \\
CC (/5)
& 3.0863 & 2.9396 & 3.0361 & 2.9250 & 2.8726 \\
\midrule
\textit{3D Geometry Consistency (/4)}
& 2.9145 & 2.8936 & 3.0919 & 2.9781 & 2.9260 \\
\midrule
RBG (/4)
& 1.8689 & 1.8531 & 2.4253 & 1.8549 & 1.7854 \\
DC (/2)
& 1.7086 & 1.6329 & 1.6029 & 1.7835 & 1.4457 \\
CTG (/1)
& 0.8687 & 0.9054 & 0.9377 & 0.9063 & 0.9807 \\
RC (/1)
& 0.7242 & 0.7084 & 0.7464 & 0.7163 & 0.7760 \\
\midrule
\textit{State-Level Understanding (/6)}
& 4.2000 & 3.6041 & 3.0401 & 3.8641 & 2.9888 \\
\midrule
OL (/2)
& 1.3527 & 0.4497 & 0.1623 & 0.8276 & 0.2581 \\
TA (/4)
& 2.0088 & 1.2721 & 1.2011 & 1.4702 & 0.7463 \\
IQ-2 (/1)
& 0.6216 & 0.6768 & 0.5474 & 0.6762 & 0.5275 \\
EAA (/2)
& 1.9008 & 1.9185 & 1.6488 & 1.9226 & 1.9184 \\
PLA (/1)
& 0.6486 & 0.6457 & 0.6247 & 0.6527 & 0.5529 \\
SA (/6)
& 4.8047 & 4.6764 & 3.9731 & 4.7550 & 3.8015 \\
\midrule
\textit{Task-Level Completeness (/3)}
& 2.3500 & 2.1072 & 1.6419 & 2.4139 & 1.4034 \\
\midrule
IF (/2)
& 1.2004 & 1.5095 & 1.0869 & 1.5422 & 0.9197 \\
CS (/1)
& 0.9385 & 0.8314 & 0.9650 & 0.9458 & 0.8930 \\
TS (/3)
& 2.4340 & 1.5630 & 0.4005 & 2.0910 & 0.1515 \\
\midrule
\textit{Average Full Score (/4)}
& 2.8379 & 2.5714 & 2.2517 & 2.7797 & 2.1246 \\ 
\midrule
\textit{\textbf{RoboPhyscore (/1)}}
& \textbf{0.6829} & \textbf{0.5359} & \textbf{0.3363} & \textbf{0.6330} & \textbf{0.3492} \\ 
\bottomrule
\end{tabular}}
\label{tab:quantitative1}
\end{table}

\begin{table}[t]
\caption{Detailed quantitative results of the RoboPhys-3D benchmark for four representative reconstruction methods.}
\centering
\small
\resizebox{\linewidth}{!}{%
\begin{tabular}{lccccc}
\toprule
\textbf{Metrics} & \textbf{Ground Truth} & \multicolumn{2}{c}{\textbf{Feed-Forward Estimator}} & \multicolumn{2}{c}{\textbf{Dynamic Representation}} \\  
& & \textbf{VGGT} & \textbf{VGGT-$\Omega$} & \textbf{4DGS} & \textbf{4C4D} \\
\midrule
\textit{Pixel-Level Fidelity (/5)}
& 2.7784 & 2.6583 & 2.6806 & 2.4376 & 2.4923 \\ 
\midrule
IQ-1 (/6)
& 3.7084 & 3.4867 & 3.5443 & 3.5562 & 3.4258 \\ 
AQ (/4)
& 1.6312 & 1.4247 & 1.4841 & 1.0956 & 1.2409 \\   
DS (/1)
& 0.4802 & 0.4646 & 0.4664 & 0.3218 & 0.3340 \\ 
MQ (/4)
& 2.6947 & 2.6530 & 2.6409 & 2.6261 & 2.7275 \\ 
CC (/5)
& 2.9932 & 2.9658 & 2.9609 & 2.9636 & 2.9762 \\ 
\midrule
\textit{3D Geometry Consistency (/4)}
& 3.0209 & 2.9555 & 2.9795 & 2.9577 & 2.8903 \\ 
\midrule
RBG (/4)
& 2.0633 & 1.9638 & 1.9973 & 1.8818 & 1.8814 \\
DC (/2)
& 1.6981 & 1.6342 & 1.6386 & 1.5974 & 1.6053 \\
CTG (/1)
& 0.9188 & 0.9200 & 0.9305 & 0.9493 & 0.8801 \\
RC (/1)
& 0.7372 & 0.7274 & 0.7303 & 0.7392 & 0.7372 \\
\midrule
\textit{State-Level Understanding (/6)}
& 3.8210 & 3.5781 & 3.6264 & 3.2604 & 3.4112 \\ 
\midrule
OL (/2)
& 0.7636 & 0.6677 & 0.6919 & 0.4116 & 0.5157 \\
TA (/4)
& 1.7982 & 1.2773 & 1.3100 & 1.2138 & 1.0992 \\
IQ-2 (/1)
& 0.6436 & 0.6163 & 0.6294 & 0.5598 & 0.6004 \\
EAA (/2)
& 1.8690 & 1.8602 & 1.8542 & 1.8666 & 1.8591 \\
PLA (/1)
& 0.6260 & 0.6219 & 0.6299 & 0.6101 & 0.6368 \\
SA (/6)
& 4.7132 & 4.5398 & 4.5993 & 3.8878 & 4.2705 \\
\midrule
\textit{Task-Level Completeness (/3)}
& 2.0767 & 1.9869 & 2.0232 & 1.8484 & 1.9811 \\ 
\midrule
IF (/2)
& 1.3294 & 1.2459 & 1.2739 & 1.1584 & 1.2511 \\
CS (/1)
& 0.9337 & 0.9096 & 0.9134 & 0.9035 & 0.9134 \\
TS (/3)
& 1.4350 & 1.3630 & 1.4185 & 1.0970 & 1.3265 \\
\midrule
\textit{Average Full Score (/4)}
& 2.6400 & 2.5292 & 2.5598 & 2.3865 & 2.4500 \\ 
\midrule
\textit{\textbf{RoboPhyscore (/1)}}
& \textbf{0.5584} & \textbf{0.5288} & \textbf{0.5364} & \textbf{0.4369} & \textbf{0.4768} \\ 
\bottomrule
\end{tabular}}
\label{tab:quantitative2}
\end{table}

\subsection{Setup}

We evaluate four open-source video world models, covering both GWMs (Wan 2.2 \cite{wan2025wan}, and CogVideoX 1.5 \cite{yang2025cogvideox}) and EWMs (Cosmos 3 \cite{agarwal2026cosmos}, and RoboDreamer \cite{zhou2024robodreamer}). 
Every rollout is scored on five visual substrates: the raw video and its reconstructions by the pose-free feed-forward estimators VGGT \cite{wang2025vggt}, and VGGT-$\Omega$ \cite{wang2026vggt} and by the dynamic 4D representations 4DGS \cite{wu20244d}, and 4C4D \cite{zhou20264c4d}.
We further ablate three scene annotations, namely the original RoboTwin 2.0 instruction \cite{chen2025robotwin}, the caption from Physion-Eval \cite{zhang2026physion}, and our view-aware caption, and three IDMs, including DreamGen \cite{jang2025dreamgen}, J-IDM \cite{li2026turning}, and MIDM \cite{feng2025vidar}.
The Ground Truth scores the ground-truth video through the same pipeline and serves as the reference ceiling.
Full quantitative and qualitative results are available in Appendices \ref{appendix:experiment} and \ref{appendix:qualitative}.

\subsection{Quantitative Results}

\subsubsection{Per-Dimension Analysis}

Overall \textit{AFS} and \textit{RoboPhyscore} results are reported in Figure \ref{fig2} and Tables \ref{tab:quantitative1}--\ref{tab:quantitative3}. 
Cosmos achieves the strongest performance among the evaluated world models, with a \textit{RoboPhyscore} of $0.6330$ ($92.7\%$ of Ground Truth), followed by Wan ($0.5359$), RoboDreamer ($0.3492$), and CogVideoX ($0.3363$). 
Cosmos also ranks first among generated models under every reconstruction method, demonstrating consistent robustness to reconstruction quality. 
Reconstruction remains an important source of error: VGGT and VGGT-$\Omega$ retain $94.7\%$ and $96.1\%$ relative to the score obtained on ground-truth video, whereas 4C4D and 4DGS retain only $85.4\%$ and $78.2\%$, respectively.

The four evaluation levels reveal markedly different behaviors. 
For pixel-level fidelity, Cosmos retains $93.7\%$ of Ground Truth, while Wan achieves higher aesthetic quality and motion quality despite substantially lower distribution similarity ($44.2\%$ of Ground Truth), showing that perceptually appealing and smooth videos need not follow the reference evolution. 
3D geometry consistency is comparatively insensitive: video world models can match or exceed Ground Truth, e.g., CogVideoX $3.0919$ vs. Ground Truth $2.9145$, and it is nearly uncorrelated with the other three levels, because internal geometric consistency does not necessarily imply correct scene state. 
This distinction becomes clear at the state-level understanding, where Cosmos remains strongest ($3.8641$), but retains only $61.2\%$ and $73.2\%$ of Ground Truth in object localization and trajectory accuracy, respectively; for Wan, these ratios further decrease to $33.2\%$ and $63.3\%$. 
In contrast, their judged interaction, event, and physical plausibility remain close to or above Ground Truth. 
Therefore, plausible rollouts can still contain substantial spatial and dynamical errors that perceptual and VLM-based evaluation does not detect..
The same discrepancy persists at the task-level completeness. 
Cosmos achieves the highest generated task-level score ($2.4139$), slightly exceeding Ground Truth ($2.3500$), yet its execution-grounded task success retains only $85.9\%$ of Ground Truth. 
Wan similarly exceeds Ground Truth in instruction following but retains only $64.2\%$ in task success. 
Most notably, CogVideoX obtains the highest collision safety ($0.9650$) while achieving only $16.5\%$ of Ground Truth task success, indicating that conservative behavior can appear safe without completing the task. 
These results motivate the multi-level design of RoboPhys-3D: visual or semantic plausibility alone is insufficient to characterize EWM capability.

\begin{table}[t]
\centering
\caption{RoboPhyscore for each video world model and each reconstruction method.}
\small
\resizebox{\linewidth}{!}{%
\begin{tabular}{lccccc}
\toprule
\textbf{Method} & \textbf{Ground Truth} & \textbf{VGGT} & \textbf{VGGT-$\Omega$} & \textbf{4DGS} & \textbf{4C4D} \\
\midrule
Ground Truth & 0.8361 & 0.7444 & 0.7592 & 0.4660 & 0.6088 \\
Wan          & 0.5782 & 0.5559 & 0.5609 & 0.4816 & 0.5029 \\
CogVideoX    & 0.3473 & 0.3269 & 0.3366 & 0.3387 & 0.3322 \\
Cosmos       & 0.6800 & 0.6649 & 0.6705 & 0.5530 & 0.5965 \\
RoboDreamer  & 0.3503 & 0.3519 & 0.3549 & 0.3453 & 0.3434 \\
\bottomrule
\end{tabular}}
\label{tab:quantitative3}
\end{table}

\begin{table}[t]
\centering
\caption{RoboPhyscore for different task categories. Task 1 is single-arm w/o target. Task 2 is single-arm w/ target. Task 3 is dual-arm w/o target. Task 4 is dual-arm w/ target.}
\small
\resizebox{\linewidth}{!}{%
\begin{tabular}{lcccc}
\toprule
\textbf{Method} & \textbf{Task 1} & \textbf{Task 2} & \textbf{Task 3} & \textbf{Task 4} \\
\midrule
Ground Truth & 0.7117 & 0.7033 & 0.6702 & 0.6464 \\ 
Wan          & 0.5596 & 0.5778 & 0.5306 & 0.4756 \\ 
CogVideoX    & 0.3424 & 0.3453 & 0.3290 & 0.3285 \\
Cosmos       & 0.6546 & 0.6648 & 0.6245 & 0.5881 \\ 
RoboDreamer  & 0.3452 & 0.3779 & 0.3450 & 0.3287 \\ 
\bottomrule
\end{tabular}}
\label{tab:quantitative4}
\end{table}

\subsubsection{Per-Task Analysis}

Table \ref{tab:quantitative4} evaluates \textit{RoboPhyscore} across four manipulation regimes. 
Performance generally decreases as coordination complexity increases: Ground Truth declines from $0.7117$ on single-arm tasks without a target to $0.6464$ on dual-arm tasks with a target, and Task 4 yields the lowest score for every evaluated model. 
Cosmos remains the strongest model across all four regimes, retaining $91.0\%$ of Ground Truth in the hardest regime (Task 4), compared with $73.6\%$ for Wan. 
The target constraint alone does not consistently increase difficulty for single-arm tasks, whereas combining target-directed manipulation with dual-arm coordination produces the clearest degradation. 
This result identifies coordinated, goal-directed multi-arm prediction as a major remaining challenge for current video world models.

\begin{table}[t]
\centering
\caption{RoboPhyscore for different scene annotations.}
\small
\resizebox{\linewidth}{!}{%
\begin{tabular}{lccc}
\toprule
\textbf{Method} & \textbf{Instruction} & \textbf{Previous Prompt} & \textbf{Current Prompt} \\
\midrule
Wan    & 0.4962 & 0.5474 & 0.5782 \\
Cosmos & 0.6519 & 0.6541 & 0.6800 \\
\bottomrule
\end{tabular}}
\label{tab:quantitative5}
\end{table}

\begin{table}[t]
\centering
\caption{Performance of different inverse dynamic models.}
\small
\resizebox{\linewidth}{!}{%
\begin{tabular}{lccc}
\toprule
\textbf{Method} & \textbf{MAE$\downarrow$} & \textbf{Task Success (Action Planner)$\uparrow$} & \textbf{Task Success (Data Engine)$\uparrow$} \\
\midrule
J-IDM               & 0.0744 & 0.7575 & 0.4875 \\
MIDM                & 0.0273 & 0.9000 & 0.4750 \\
DreamGen            & 0.0112 & 0.9825 & 0.5875 \\
\bottomrule
\end{tabular}}
\label{tab:quantitative6}
\end{table}

\subsubsection{Ablation Study}

Table \ref{tab:quantitative5} ablates scene annotation while fixing the visual substrate to the ground-truth video. 
Increasing prompt specificity consistently improves \textit{RoboPhyscore} for both models. 
For Wan, the score increases from $0.4962$ with the original RoboTwin instruction to $0.5474$ with the Physion-Eval prompt and $0.5782$ with our view-aware prompt, yielding a $16.5\%$ improvement over the instruction baseline. 
Cosmos exhibits the same trend with a smaller $4.3\%$ gain, indicating greater robustness to language specification. 
These results show that richer, view-aware conditioning improves task-aligned embodied prediction.

Table \ref{tab:quantitative6} further validates our choice of DreamGen as the IDM. 
DreamGen achieves the lowest action prediction error (MAE $=0.0112$) and the highest downstream success for both Action Planner ($0.9825$) and Data Engine ($0.5875$), consistently outperforming J-IDM and MIDM. 
Compared with MIDM, DreamGen reduces MAE by $59.0\%$ while improving Action Planner and Data Engine success by $9.2\%$ and $23.7\%$, respectively. 
Therefore, we adopt DreamGen for all execution-grounded evaluations.

\begin{figure*}[t]
  \centering
  \includegraphics[width=\linewidth]{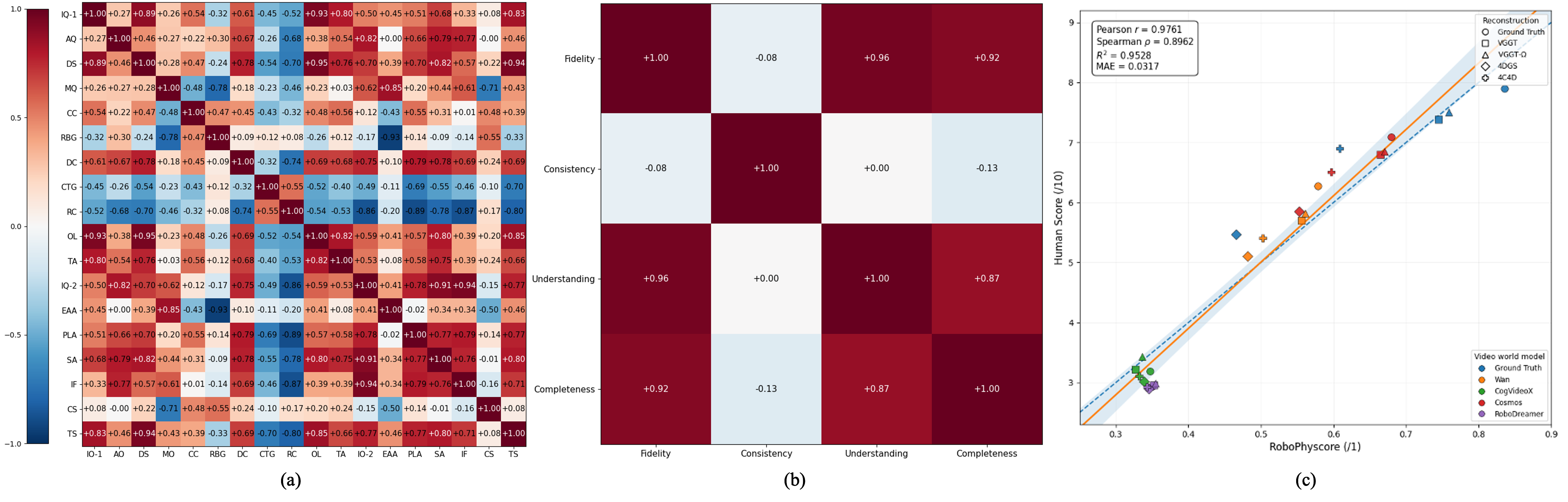}
  \caption{Cross-dimension correlation analysis. (a) Pearson correlation among 18 sub-dimensions. (b) Pearson correlation among four levels. (c) Pearson correlation between RoboPhyscore and human evaluation.}
  \label{fig4}
\end{figure*}

\begin{figure*}[t]
  \centering
  \includegraphics[width=\linewidth]{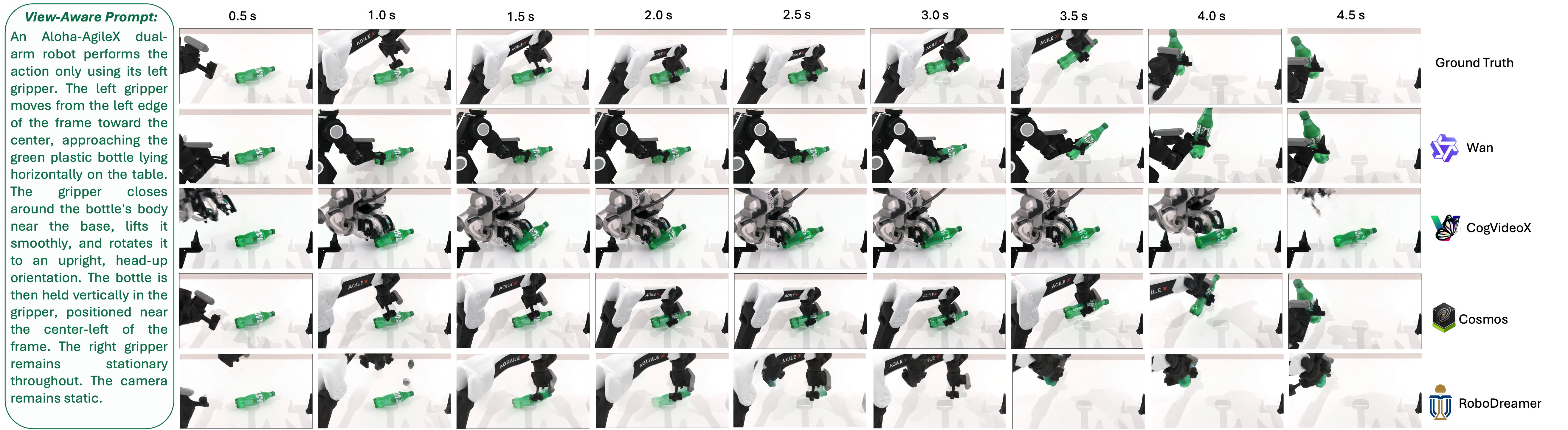}
  \caption{Qualitative examples of \textit{Adjust Bottle} task (different world models + ground truth + view-aware prompt).}
  \label{fig3}
\end{figure*}

\subsubsection{Cross-Dimension Analysis}

Figure \ref{fig4} reports Pearson correlations among the 18 sub-dimensions and among the four evaluation levels, with the agreement between \textit{RoboPhyscore} and human judgments.
Following the human-alignment protocol \cite{huang2024vbench}, 70 annotators assign each sample an overall score on a 10-point scale over the video generation quality considering whether each video depicts a coherent and physically plausible robot interaction that correctly executes the intended task.
\textit{RoboPhyscore} exhibits strong agreement with human evaluation, achieving Pearson $r=0.9761$ and Spearman $\rho=0.8962$. 
\textit{RoboPhyscore} also outperforms alternative unified scores.
More details about human alignment are provided in Appendix \ref{appendix:human}.

\subsection{Qualitative Results}

Figure \ref{fig3} compares representative rollouts for the \textit{Adjust Bottle} under the same view-aware instruction. 
The ground-truth sequence performs the required multi-stage manipulation, but the evaluated world models exhibit distinct failure modes. 
Wan preserves the overall manipulation intent but shows noticeable embodiment inconsistency. 
CogVideoX does not complete the required task, leaving the predicted interaction inconsistent with the instructed final state. 
RoboDreamer suffers from pronounced visual and structural artifacts that compromise both scene coherence and manipulation fidelity. 
In contrast, Cosmos produces the most coherent rollout, preserving the robot embodiment and static viewpoint while successfully completing the task.

%% file: sec/6_conclusion.tex
\section{Conclusion}
\label{sec:conclusion}

We introduce \textbf{RoboPhys-3D}, a 3D-grounded benchmark that evaluates EWMs by comparing generated rollouts with ground-truth manipulation trajectories within a shared reconstructed 3D scene.
Built on RoboTwin 2.0, RoboPhys-3D evaluates 50 tasks in four regimes and scores each rollout with 50 metrics across 18 sub-dimensions and four complementary levels.
\textit{AFS} provides comprehensive diagnostic coverage and \textit{RoboPhyscore} provides a compact, task-aligned indicator whose agreement with human judgment (Pearson $r=0.9761$, Spearman $\rho=0.8962$). 
Across four representative world models, Cosmos 3 achieves the highest \textit{RoboPhyscore} ($0.6330$, $92.7\%$ of ground truth) under every reconstruction method and in every task regime. 
The results reveal a systematic gap between plausible generation and correct embodied prediction: models can achieve strong perceptual, VLM-based, or collision-safety scores while exhibiting substantially weaker object localization, trajectory accuracy, and downstream task success. 
Reconstruction quality alone shifts RoboPhyscore by up to 21.8\%, confirming that reconstruction-induced error must be separated from generation-induced error before 3D-level conclusions are drawn.
This work also has limitations.
RoboPhys-3D currently focuses on a single simulation platform and embodiment. 
Future work will extend the benchmark toward cross-embodiment, cross-dataset, and sim-to-real evaluation, as well as broader comparisons with VLA policies.